\documentclass{article}
\PassOptionsToPackage{numbers, compress}{natbib}

\usepackage[preprint]{mirai}
\usepackage[utf8]{inputenc} % allow utf-8 input
\usepackage[T1]{fontenc}    % use 8-bit T1 fonts
\usepackage{url}            % simple URL typesetting
\usepackage{booktabs}       % professional-quality tables
\usepackage{amsfonts}       % blackboard math symbols
\usepackage{nicefrac}       % compact symbols for 1/2, etc.
\usepackage{microtype}      % microtypography
\usepackage{xcolor}         % colors
\usepackage{tabularray}
\UseTblrLibrary{booktabs}
\usepackage{enumitem}
\setlist{nosep}
\setlist{leftmargin=*,topsep=0pt}
\usepackage{graphicx}

\definecolor{uclablue}{rgb}{0.15, 0.45, 0.68}
\usepackage[
    pagebackref,
    breaklinks,
    citecolor=uclablue,
    colorlinks=true,
]{hyperref}
\definecolor{custom2}{HTML}{F58157}
\definecolor{custom3}{HTML}{E7434C}
\definecolor{custom4}{HTML}{99216A}
\definecolor{custom5}{HTML}{64256E}
\definecolor{custom6}{HTML}{291956}

\usepackage{multirow} 
\usepackage{graphicx}
\usepackage[export]{adjustbox}
\usepackage{float}
\usepackage{epsfig}
\usepackage{graphicx}
\usepackage{amsmath}
\usepackage{amssymb} % http://ctan.org/pkg/amssymb
\usepackage{caption}
\usepackage{xparse} % data cards
\usepackage{wrapfig} % wrapped table
\usepackage{tabularx}
\usepackage{subcaption}
\usepackage{fancyhdr}
\usepackage[hang,flushmargin]{footmisc} % remove indentation in footnote
\usepackage{pifont} % http://ctan.org/pkg/pifont
\usepackage{color}
\usepackage{wrapfig} % wrapped table
\usepackage{boxedminipage}
\usepackage{framed}
\usepackage{soul}
\usepackage{xspace}
\usepackage{booktabs} % professional-quality tables
\usepackage{makecell}
\usepackage{xcolor}
\usepackage{vcell}
\usepackage{colortbl}
\usepackage[nodisplayskipstretch]{setspace}
\usepackage{seqsplit} 
\usepackage{array}
\usepackage{todonotes}
\usepackage{forest}

\usepackage[most]{tcolorbox}

\definecolor{codegreen}{rgb}{0,0.6,0}
\definecolor{codegray}{rgb}{0.5,0.5,0.5}
\definecolor{codepurple}{rgb}{0.58,0,0.82}
\definecolor{backcolour}{rgb}{0.95,0.95,0.92} % to change to a lighter color

\usepackage{enumitem}
\usepackage[compact]{titlesec}
\usepackage{sidecap}
\titlespacing*{\section}{3pt}{*0.5}{*0.5}
\titlespacing*{\subsection}{2pt}{*0.5}{*0.5}
\titlespacing*{\subsubsection}{2pt}{*0.5}{*0.5}
\lstdefinestyle{compressedstyle}{
    language=Python,
    basicstyle=\ttfamily\scriptsize,  % Smaller font for code
    breaklines=true,
    aboveskip=0pt,  % Reduce space above code
    belowskip=0pt,  % Reduce space below code
    lineskip=-2pt,  % Reduce line spacing
}

\newtcblisting{compressedlisting}[2][]{
    listing only,
    arc=1mm,  % Smaller arc for corners
    outer arc=1mm,
    top=0mm,  % Smaller top margin
    bottom=0mm,  % Smaller bottom margin
    left=3mm,  % Smaller left margin
    right=3mm,  % Smaller right margin
    boxsep=2mm,  % Smaller padding inside the box
    titlerule=0mm,  %// No space between title and box content
    titlerule style={opacity=0},  %// Make title rule invisible
    title after break={},  %// No title in break continuation
    colback=white,
    listing options={style=compressedstyle},
    title=#2,
    #1
}

\newcommand{\hlc}[2][yellow]{{%
    \colorlet{foo}{#1}%
    \sethlcolor{foo}\hl{#2}}%
}
\newcommand{\customfootnotetext}[2]{{%
  \renewcommand{\thefootnote}{#1}%
  \footnotetext[0]{#2}}}%

\addtocontents{toc}{\protect\setcounter{tocdepth}{-1}}

\newif\ifshowcomment
\showcommenttrue % To show  comments
\newcommand{\Skip}[1]{}

\newcommand{\dataset}{\textsc{Mirai}\xspace}

\newcommand{\dotieconcat}[2]{% auxiliary macro, don't use it directly
  \text{\raisebox{.8ex}{$\smallfrown$}}%
}
\newcommand{\mypar}[1]{\noindent\textbf{#1}}

\newcommand{\colorcellnull}[1]{\cellcolor{gray!20}---}

\usepackage{amsmath,amsfonts,bm}

\def\eqref#1{equation~\ref{#1}}
\def\1{\bm{1}}

\def\eve{{e}}

\def\evt{{t}}

\def\mE{{\bm{E}}}

\DeclareMathAlphabet{\mathsfit}{\encodingdefault}{\sfdefault}{m}{sl}
\SetMathAlphabet{\mathsfit}{bold}{\encodingdefault}{\sfdefault}{bx}{n}

\def\gC{{\mathcal{C}}}

\def\gR{{\mathcal{R}}}

\newcommand{\KL}{D_{\mathrm{KL}}}

\title{\dataset: Evaluating LLM Agents for Event Forecasting}

\author{Chenchen Ye$^{*1}$, Ziniu Hu$^{*2}$, Yihe Deng$^{*1}$, Zijie Huang$^{1}$,\vspace{0.1cm}\\ \textbf{Mingyu Derek Ma$^{1}$, Yanqiao Zhu$^{1}$, Wei Wang$^{1}$}\vspace{0.2cm}\\
  $^1$University of California, Los Angeles\\
  $^2$California Institute of Technology\vspace{0.2cm}\\
  \url{https://mirai-llm.github.io/} 
}
\begin{document}
\customfootnotetext{}{$^{*}$ Equal Contribution.}
\maketitle

\begin{abstract}
    \label{sec:abstract}

Recent advancements in Large Language Models (LLMs) have empowered LLM agents to autonomously collect world information, over which to conduct reasoning to solve complex problems. Given this capability, increasing interests have been put into employing LLM agents for predicting international events, which can influence decision-making and shape policy development on an international scale.
Despite such a growing interest, there is a lack of a rigorous benchmark of LLM agents' forecasting capability and reliability. 
To address this gap, we introduce \dataset, a novel benchmark designed to systematically evaluate LLM agents as temporal forecasters in the context of international events. 
Our benchmark features an agentic environment with tools for accessing an extensive database of historical, structured events and textual news articles. 
We refine the GDELT\footnote{GDELT: \href{https://www.gdeltproject.org/}{https://www.gdeltproject.org/}} event database with careful cleaning and parsing to curate a series of relational prediction tasks with varying forecasting horizons, assessing LLM agents' abilities from short-term to long-term forecasting.
We further implement APIs to enable LLM agents to utilize different tools via a code-based interface.
In summary, \dataset comprehensively evaluates the agents' capabilities in three dimensions: 1) autonomously source and integrate critical information from large global databases; 2) write codes using domain-specific APIs and libraries for tool-use; and 3) jointly reason over historical knowledge from diverse formats and time to accurately predict future events. 
Through comprehensive benchmarking, we aim to establish a reliable framework for assessing the capabilities of LLM agents in forecasting international events, thereby contributing to the development of more accurate and trustworthy models for international relation analysis.\footnote{Our dataset is available on \href{https://drive.google.com/file/d/1xmSEHZ_wqtBu1AwLpJ8wCDYmT-jRpfrN/view?usp=sharing}{Google Drive}, the code can be found on \href{https://github.com/yecchen/MIRAI}{GitHub}, and an interactive demo is provided through a \href{https://colab.research.google.com/drive/1QyqT35n6NbtPaNtqQ6A7ILG_GMeRgdnO?usp=sharing}{Google Colab notebook}.}

\end{abstract}

\section{Introduction}

Accurate forecasting of international events is crucial~\citep{brown_economic_2018}, as understanding the evolution of geopolitical developments enables stakeholders to make informed decisions, mitigate risks, and seize opportunities in the interconnected world. Traditionally, researchers in international relations rely on domain expertise~\citep{smith_geopolitical_2020, johnson_diplomatic_2019}.
They conduct detailed analyses of the complex interplay among nations, considering alliances, trade agreements, ideological affinities, and historical rivalries to forecast events such as conflicts, collaborations, or alliance shifts~\citep{davis_alliances_2017}.
With the rapid development of deep learning techniques, forecasting through data-driven neural networks becomes an attractive alternative. Despite their success,
current methods rely on single types of information—either structured knowledge graphs~\citep{mahdisoltani_yago3_2015, jin_recurrent_2020, li_temporal_2021} or textual datasets~\citep{zou_forecasting_2022, reddy_smartbook_2023}. Knowledge graphs, although organized, can suffer from incompleteness~\citep{DBLP:conf/wsdm/GalarragaRAS17} or bias~\citep{DBLP:conf/www/0002HZBZWSG024}, while textual analyses may lack the necessary factual grounding for precise predictions. Moreover, these models are unable to ground their reasoning to historical evidence, which compromises the interpretability and validation of their forecasts. These limitations raise concerns about AI forecasters' reliability, particularly for high-stake scenarios~\citep{article}.

Large Language Model (LLM) agents present a promising path for overcoming these challenges~\citep{sumers_cognitive_2024,liu_agentbench_2023,weng_llm_2023,noauthor_autogpt_nodate,wang_voyager_2023}. 
These advanced AI systems exhibit the potential to mimic human experts by utilizing a diverse array of tools to automatically gather and process information from various sources, including text, knowledge graphs, and numerical data~\citep{shen_hugginggpt_2023,lu_chameleon_2023,zhuang_toolqa_2023, li_api-bank_2023}. 
Trained on extensive textual data, LLMs~\citep{achiam2023gpt,anthropic2023claude,deepmind2023gemini,touvron2023llama} are capable of 
grasping the subtleties of international relations, reasoning through complex relationships with linguistic explanations, and planning their tool usage effectively~\citep{yuan_distilling_2023,liu_llmp_2023, valmeekam_planbench_2023, ma_context-aware_2023}. 
Such capability opens up new possibilities for developing transparent and interpretable forecasting models that can be further scrutinized and refined.

Despite the immense potential of LLM agents for event forecasting, there is a lack of standardized benchmarks to assess their forecasting abilities in the realm of intricate international events. 
To address this gap, we introduce \dataset (\underline{M}ulti-\underline{I}nformation Fo\underline{R}ecasting \underline{A}gent \underline{I}nterface), a novel benchmarking environment designed to rigorously assess and advance the capabilities of LLM agents for international event forecasting over time. 
\dataset adapts real-world international event data derived from the Global Database of Events, Language, and Tone (GDELT)~\citep{leetaru_gdelt_nodate} to an event-forecasting task format across different horizons, providing a robust assessment of LLM performance. 
The framework's agentic environment enables LLMs to interact with both relational and textual databases through Application Programming Interfaces (APIs), facilitating autonomous information gathering, processing, and application in a contextually relevant manner.

We evaluate both open and closed sourced LLMs on \dataset with ReAct-Style~\citep{DBLP:conf/iclr/YaoZYDSN023} Agents and ``Single Function'' and ``Code Block'' action types.
Our extensive experiments revealed:
1) Temporal forecasting tasks are challenging for LLM agents, with the highest-performing GPT-4o agent using full suite of APIs achieves a 29.6 F1 score in second-level relation prediction tasks; while tasks involving long-term and fine-grained event forecasting are even more challenging;
2) The ``Code Block'' tool-use strategy, which allows more flexible interactions, demands robust code generation capabilities. Notably, GPT-4o is the only model that effectively utilizes and benefits from this strategy.
These findings emphasize the need for ongoing research into temporal reasoning and the effective use of tools by LLM agents. We expect that \dataset could serve as a standard benchmark for evaluating LLMs in event forecasting. This would support the development of more precise and reliable models for political analysis, enhancing our understanding of global dynamics.
\section{The \dataset Benchmark}
\label{sec:dataset}
In this section, we introduce \dataset benchmark from: the specifics of the data and tasks (Sec.~\ref{sec:data-and-task}), the implemented agents and environments (Sec.~\ref{sec:agent}), and the database construction details (Sec.~\ref{sec:database-construction}). 

\subsection{Data and Tasks}\label{sec:data-and-task}

\begin{figure}[t!]
    \centering
    \vspace{-0.8cm}
    \includegraphics[width=0.95\linewidth]{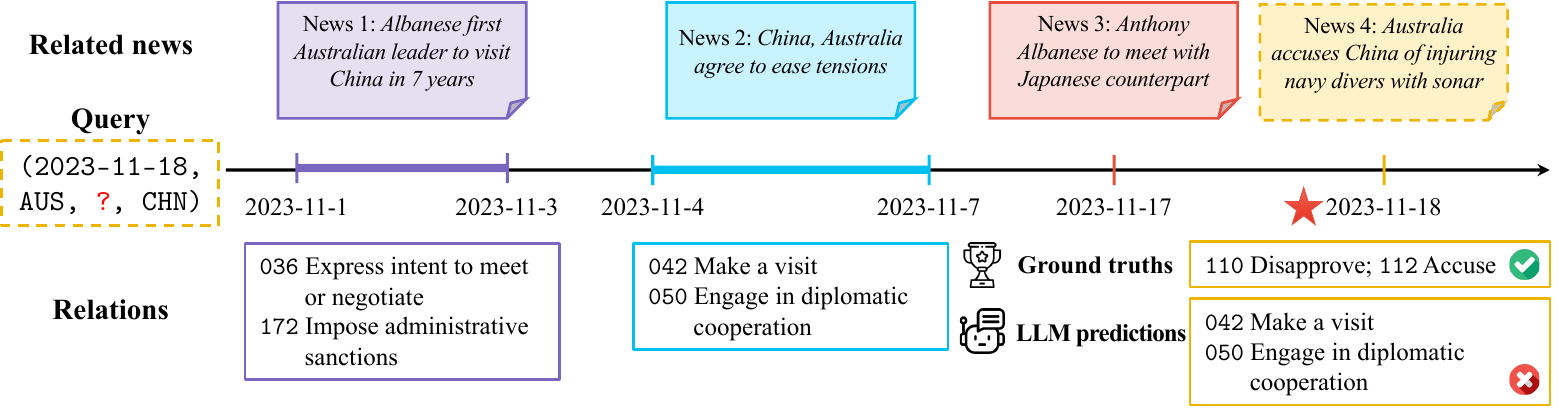}
    \caption{An example of forecasting the relations between Australia and China on Nov.18.2023. The database contains query-related historical relations and news articles, while the agent fails to predict the change of relation and makes a wrong forecast.}
    \label{fig:example}
\end{figure}

\begin{figure}[t!]
    \centering
    \vspace{-0.8cm}
    \includegraphics[width=\textwidth]{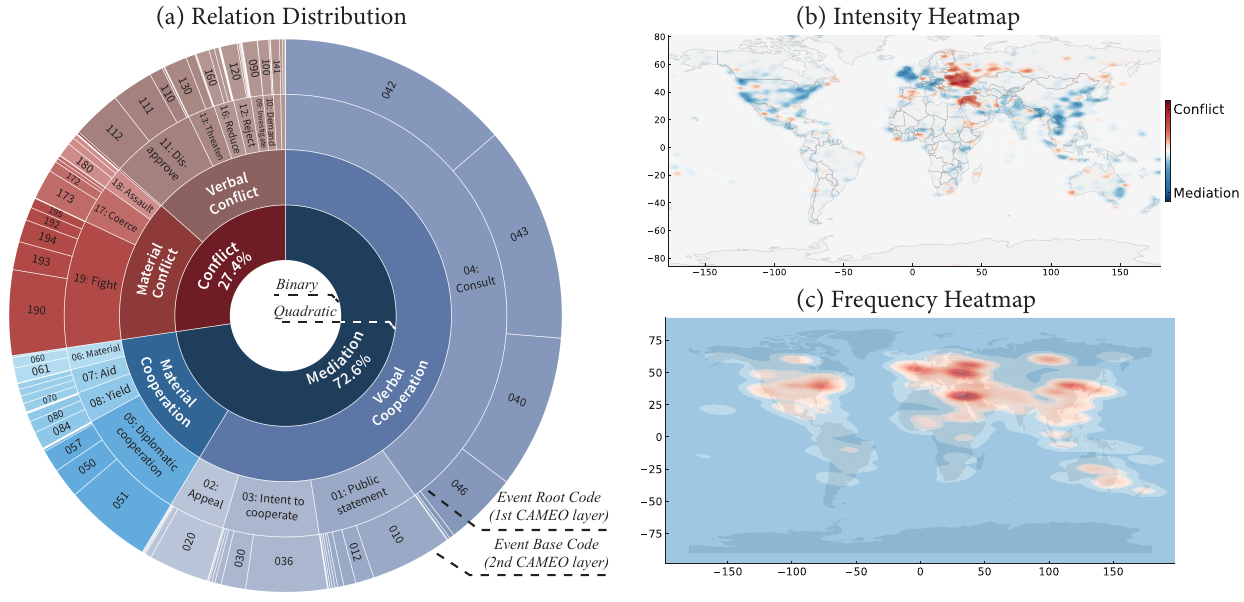}
    \caption{\dataset comprehensively covers global event data. (a) The circular chart shows the relation hierarchy and distribution in \dataset. (b) The heatmap visualizes the intensity of these events globally, distinguishing between areas of conflict (red) and mediation (blue). (c) The heatmap illustrates the frequency of these events, highlighting regions with the most occurrences.}
    \label{fig:data}
\end{figure}

We introduce \dataset, a benchmark crafted for evaluating LLM agents for temporal forecasting in the realm of international events, with tool-use and complex reasoning.
We consider \textit{forecasting} as the process of collecting essential historical data and performing temporal reasoning to anticipate the outcomes of future events.

Formally, we represent an event as $\eve^t = (t, s, r, o)$, where $t$ is the timestamp\footnote{Each timestamp uniquely represents a day, formatted in "YYYY-MM-DD".
}, $s,o \in \gC$ are respectively the \underline{s}ubject and \underline{o}bject countries\footnote{In this paper, the term "country" includes all countries, dependent territories, special geographic areas, and their subdivisions. We use the standardized ISO-3166 Alpha-3 codes for country names, such as "\texttt{AUS}" for Australia and "\texttt{CHN}" for China, to maintain global consistency.} from the country pool $\gC$, and $r \in \gR$ denotes the relation type defined by the Conflict and Mediation Event Observations (CAMEO) ontology~\citep{CAMEO}\footnote{CAMEO is a well-established ontology for categorizing international political events, meticulously developed by experts to enable detailed analysis across multiple levels of granularity.}. 
As shown in Figure~\ref{fig:data}a, we incorporate two hierarchical levels from the CAMEO ontology to facilitate a detailed and comprehensive analysis of geopolitical dynamics.  
The first level includes 20 broad categories, represented by a two-digit code (e.g., ``\texttt{01}: \texttt{Public Statement}'' or ``\texttt{04}: \texttt{Consult}''), which are subdivided into second-level categories identified by a three-digit code that corresponds to its parent category. 
For example, ``\texttt{03}: \texttt{Express intent to cooperate}'' is a first-level category that includes 10 different second-level relations such as ``\texttt{036}: \texttt{Express intent to meet or negotiate}''. 
Subsequently, the quadruple ``(\texttt{2023-11-03}, \texttt{AUS}, \texttt{036}, \texttt{CHN})'' denotes that on 3 November 2023, the Australian leader announces a planned visit to China.  
These relations are also organized along two dimensions, from \hlc[yellow!30]{\texttt{Verbal}} to \hlc[orange!30]{\texttt{Material}} and from \hlc[red!30]{\texttt{Conflict}} to \hlc[cyan!30]{\texttt{Cooperation}}, to form a quadratic categorization in the inner circle of Figure~\ref{fig:data}a.

Furthermore, a collection of events that happened at the same timestamp $\evt$ form a \emph{timestamped event set} $\mE^t = \{\eve_1^t, ..., \eve_M^t\} $, where $M$ is the number of unique events at time $\evt$. 
These events $\{\mE^t\}_{t=1}^T$ can therefore be organized into a temporal graph, with countries as nodes and events' relations as edges.

The task of \textbf{event forecasting} $(t+l, s, ?, o)$ is to predict all the events between a pair of countries $s$ and $o$, happening $l$ days in the future from the current time $t$.
Formally, given historical events up to the current time $\mE^{\leq t}$, our goal is to forecast future relationships $\mE_{s,o}^{t+l}$ between a pair of countries. This includes utilizing events involving third parties, such as interactions with their mutual neighbors.

In the example of Figure~\ref{fig:example}, we demonstrate how to predict which actions Australia will adopt towards China on 18 November 2023 based on all available information up to 17 November 2023. 
This can also be formatted as a structured query ``(\texttt{2023-11-17} + [1 day], \texttt{AUS}, ?, \texttt{CHN})'', where the temporal distance of the forecasted event is one day. From the historical events, we can see that the two countries last long-standing tensions; though news show the leader of Australia visited China and agreed to ease tension, this signal does not change the overall tendency, and their relationship keeps being \texttt{Disapprove} and \texttt{Accuse} on certain focuses. Thus, the agent over-relying short-term news and eventually gives the wrong prediction \texttt{diplomatic cooperation}.

\subsection{Agents and Environments} \label{sec:agent}

\begin{figure}[t!]
    \centering
    \vspace{-0.8cm}
    \includegraphics[width=1\textwidth]{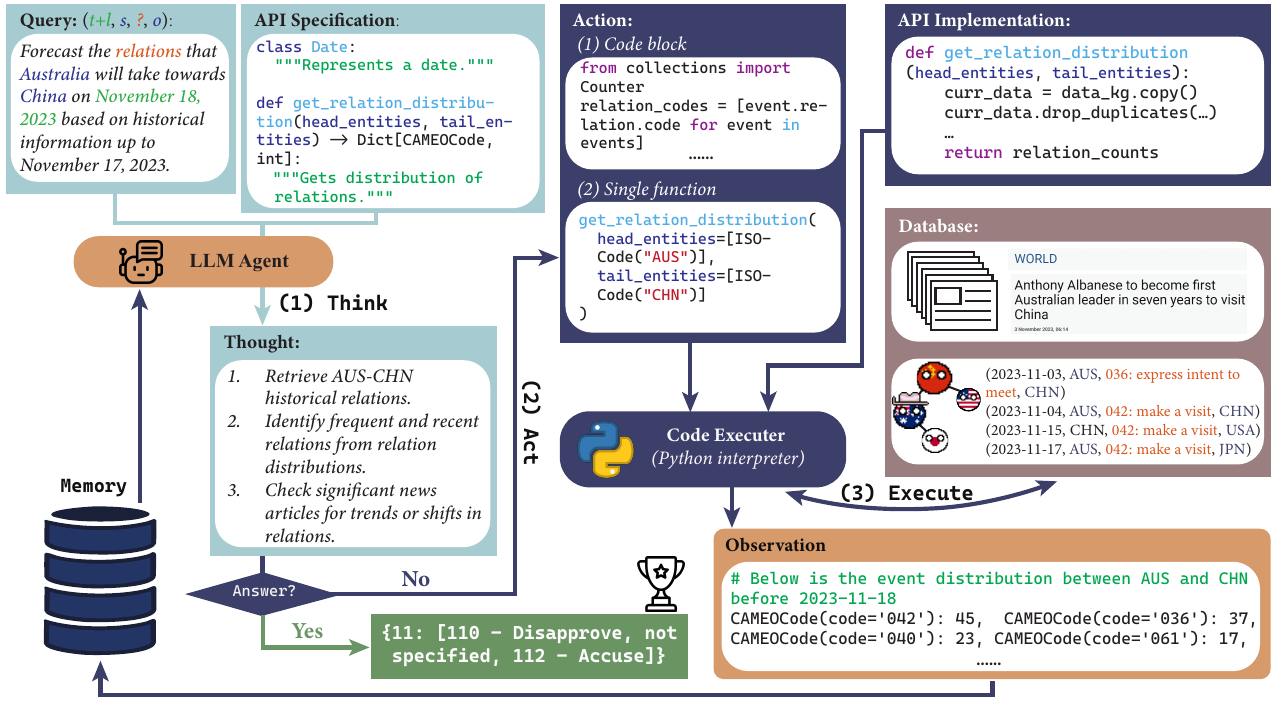}
    \vspace{-0.4cm}
    \caption{Overview of the LLM agent's interaction with the multi-source environment using the ReAct strategy for forecasting a query event. The framework consists of three main steps: (1) Think: The agent analyzes the current status and plans the next action based on the query and the provided API specifications. (2) Act: The agent generates a ``\hlc[uclablue!20]{Single Function}'' call or a ``\hlc[blue!20]{Code Block}'' to retrieve and analyze relevant data from the database. (3) Execute: The Python interpreter runs the generated code with the API implementation and database and produces observations. These steps are iteratively performed until the agent reaches a final forecast for the future relation.}
    \label{fig:draft}
\end{figure}

Similar to human political analysts, LLM agents must leverage a variety of information sources to make reliable predictions. 
To this end, we abstract the environment by incorporating coding APIs that facilitate interactions with various knowledge sources. 
Within this environment, we can develop an LLM agent that employs these APIs through the ReAct strategy~\citep{yao_react_2023}, characterized by the iterative steps of \textit{think}, \textit{act}, and \textit{observe}. 
This structured approach allows the agent to analyze the current situation, retrieve additional data, and observe outcomes to make reliable forecasting.

\noindent\textbf{APIs.} We provide a comprehensive set of API to the LLM agent, enabling the execution of the generated codes to access a rich database of historical events and news articles. 
The API contains the essential \textit{data classes} and a suite of \textit{functions} designed to accommodate various types of information within the database. 
\textit{Data classes} range from unary types such as date, country, and relation, to composite types like date range, event, and news articles.
\textit{Functions}, executable via standard Python syntax, allow the agent to efficiently query the database through code interface. The design of these functions is characterized in two aspects:
\begin{itemize}
    \item \textbf{Information types.} These functions cover various types of information, including country and relation code-name mapping and hierarchy, and count, listing, and distribution of events and news articles. 
    For example, the function \texttt{map\_relation\_description\_to\_cameo} takes a relation description and returns a list of the five most likely relations with their respective CAMEO codes, names, and descriptions, providing precise relation information as needed.
    \item \textbf{Search conditions.} Additionally, the API functions related to events and news articles include various optional parameters that enable tailored searches based on different criteria. 
    For instance, the function \texttt{get\_event} enables searches by specifying conditions like \texttt{date\_range}, \texttt{head\_entities}, \texttt{tail\_entities}, \texttt{relations}, and \texttt{text\_description} to fetch events that meet these conditions from the database. 
\end{itemize}
The API specifications and full implementations are shown in Appendix~\ref{append:api}.

\noindent\textbf{Interaction with environment.}
In \dataset, agents are given the API specifications, which include meticulously documented Python docstrings for all data classes and function definitions, abstracting the underlying implementation details. 
The environment is equipped with a Python code sandbox with full access to APIs and databases. The agent uses the ReAct strategy in an iterative manner, alternating between \textit{think}, \textit{act}, and \textit{observe} to gather information and perform forecasting. The pipeline is illustrated in Figure~\ref{fig:draft}.
% \vspace{-3pt}
\begin{enumerate}
\setlength{\itemsep}{0pt}
\item \textbf{Think.} Specifically, the agent first \textit{thinks} about the current situation. If it is confident to deliver a final forecast based on the current information and analysis, it provides the answer and stops the iteration. Otherwise, it further analyzes the current status and plans subsequent actions. In the first step shown in Figure~\ref{fig:draft}, the agent plans to first retrieve recent events directly involving the two countries and later verify the details by checking related news articles. 

\item \textbf{Act.} Next, the agent \textit{acts} by generating executable codes to interact with the environment. There are two types of action: ``\hlc[uclablue!20]{Single Function}'' or ``\hlc[blue!20]{Code Block}''. A ``\hlc[uclablue!20]{Single Function}'' facilitates straightforward data retrieval, whereas a ``\hlc[blue!20]{Code Block}'' supports a more flexible and larger action space, accommodating complex control flows such as loops and conditional statements. 
Also, the agent is allowed to use functions defined in the API and can additionally utilize safe, well-established Python libraries such as \texttt{numpy}, \texttt{networkx}, and \texttt{scikit-learn}. As shown in Figure~\ref{fig:draft}, the agent can write either a ``\hlc[blue!20]{Code Block}'' with an imported library and an inline loop, or a ``\hlc[uclablue!20]{Single Function}'' call with targeted countries as the argument value. 

\item \textbf{Observe.} The environment receives the code generated by the agent, executing it internally using a Python interpreter with full access to API implementations and the database. The output from this execution is then passed to the agent as the content of its `Observation' step. 
In particular, the agent can \textit{observe} the corresponding output if its generated code executes successfully; otherwise, it can only \textit{observe} a specified error message along with the error type. In the figure, the written code is executed, and results are sent back to the agent memory to proceed next iteration.

\end{enumerate}

These `Thought', `Action', and `Observation' from the previous iterations are stored in the agent's memory base and used as the context of the agent in the next iteration. Additionally, the variables defined in the code generated in any previous action steps are stored and available for future actions. Examples are shown in Appendix~\ref{append:examples}.

\subsection{Database Construction}\label{sec:database-construction}

\textbf{Pre-processing.} We construct a GDELT-based environment to enable agents' interaction with both knowledge graph and textual information. 
Initially, we filtered the dataset for events occurring between Jan. 1, 2023, and Nov. 30, 2023, aligning event dates with their respective news publish dates to avoid information leakage. 
We standardize the \texttt{CountryCode} to ISO-3166 norms and exclude any entries with missing or outdated codes, and standardize Event codes to the second level of the CAMEO ontology to ensure a consistent and detailed representation.
Third-level event codes, which are more noisy and contain more extraction errors were excluded. We also only consider international events. For the news articles, credibility was assessed by counting the daily mentions of each event, with a threshold of at least 50 mentions to filter out unreliable sources. 
News content, including titles and body text, was then downloaded and cleaned following the OBELICS~\citep{DBLP:conf/nips/LaurenconSTBSLW23} protocol, which involved removing low-quality text based on word counts, character and word repetition ratios, and the presence of special characters and flagged words, thereby significantly reducing noise and enhancing the reliability of the textual information stored in our database. 
We list the details for dataset construction in Appendix~\ref{sec:construction}.
We finalized a collection of 991,759 GDELT event records, corresponding to 59,161 unique $(t,s ,r, o)$ events and 296,630 unique news articles. 
We display our curated events over the world map in Figure~\ref{fig:data}b and \ref{fig:data}c, illustrating global coverage and varying intensities of conflict and mediation across regions.

\noindent\textbf{Test query set construction.}
We construct the test data using records from November 2023, 
comprising 152,500 GDELT event records and 7385 unique events. To further ensure the data quality and minimize potential extraction errors inherent in GDELT records, we set a higher threshold of at least 100 daily mentions and 5 downloaded news articles for each event. This stricter filtering results in a collection of 2,136 unique $(t, s, r, o)$ events. 
We construct the test query set based on this collection, which leads to 705 $(t, s, ?, o)$ query and answer, where each answer includes a list of relations occurring between the specified subject and object country at time $t$ in this collection. Note that this stricter filtering is only applied for the test query construction and will not remove data from the database. Additionally, we sample 100 queries from the test query set to form a test query subset, ensuring a balanced representation of dates throughout the month, countries across continents, and relations spanning the first-level CAMEO code types.

\noindent\textbf{Evaluation metrics.}
\label{sec:evalmetric}
We instruct the agent to generate final forecasting answer in a JSON dictionary, where keys are predicted two-digit first-level CAMEO codes and the values are lists of predicted three-digit second-level CAMEO codes inherited from the keys. 
For example, the JSON dictionary ``\{"\texttt{01}": ["\texttt{011}", "\texttt{012}"], "\texttt{02}": ["\texttt{023}"]\}'' indicates the first level relation predictions for "\texttt{01}: \texttt{Make public statement}" and "\texttt{02}: \texttt{Appeal}", along with second level relation predictions "\texttt{011}: \texttt{Decline comment}", "\texttt{012}: \texttt{Make pessimistic comment}", and "\texttt{023}: \texttt{Appeal for material aid}". 
We evaluate these relation predictions by calculating precision, recall, and F1 score between the predicted and ground-truth lists. 
Moreover, we map the predicted and ground truth relations to their respective binary and quadratic classes (as shown in Figure~\ref{fig:data}a). 
Since each relation in the ground-truth list may belong to different classes, we employ the empirical Kullback-Leibler (KL) divergence of predictions to the ground truth: $D_{\text{KL}}(P || Q) = \sum_{i} P(i) \log\left(\frac{P(i)}{Q(i)}\right)$ to measure the discrepancy, where \(P\) and \(Q\) represent the frequencies of ground-truth and predicted relations respectively. A lower KL divergence indicates a better alignment of the model's predictions with the ground-truth list. 

\section{Experiments}
\label{sec:exp}

Our experiment section begins with comparing the forecasting performance of different agent methods with different prediction horizons (Sec.\ref{sec:exp_tool}). Moreover, we evaluate agents with different base language models (Sec.\ref{sec:exp_basemodel}), 
and finally analyze several key aspects (Sec.\ref{sec:analysis}) to understand agents' behavior.

\subsection{Evaluate Forecasting with Different Agent Methods and Tools}\label{sec:exp_tool}

We investigate the effect of different tools (APIs) and agent tool-use strategies. We use \texttt{gpt-3.5-turbo-0125}~\citep{gpt-3.5} as the base model and evaluate on the 705 queries in the test set. For all experiments, we set the model temperature to $0.4$ and run 5 times to calculate the mean and standard deviation. We provide the detailed prompts in Appendix~\ref{append:prompts}.

\begin{table*}[!t]
    \centering
    \vspace{-0.8cm}
    \caption{Experimental results on the test set with different agent tools and the tool-use strategies. The best-performing score is highlighted in \textbf{bold} and the second-best is \underline{underlined}.}
    \resizebox{0.99\textwidth}{!}{%
    \begin{tblr}{
        colspec = {ccccccccccc},
        row{1-2} = {bg=gray!25},
        row{4,7-8} = {bg=gray!10}
    }
    \toprule
        \SetCell[r=2]{c}{Agent} & \SetCell[c=2]{c}{Tool-Use} && \SetCell[r=2]{c}{Binary \\ KL ($\Downarrow$)} & \SetCell[r=2]{c}{Quad \\ KL ($\Downarrow$)} & \SetCell[c=3]{c}{First-level Relation (\%)} &&& \SetCell[c=3]{c}{Second-level Relation (\%)} \\
        \cmidrule[lr]{2-3} \cmidrule[lr]{6-8} \cmidrule[lr]{9-11}
        & Action Type & API & & & Pre. ($\Uparrow$) & Rec. ($\Uparrow$) & F1 ($\Uparrow$) & Pre. ($\Uparrow$)& Rec. ($\Uparrow$) & F1 ($\Uparrow$)\\
        \midrule
       Direct IO & --- & --- & 6.5\textsubscript{±1.7} & 15.9\textsubscript{±1.5}&27.6\textsubscript{±8.1} & 19.7\textsubscript{±5.9} & 18.8\textsubscript{±6.9} & 6.6\textsubscript{±1.5} & 5.1\textsubscript{±0.4} & 3.5\textsubscript{±0.8}
 \\
        ZS-COT & --- & --- &6.9\textsubscript{±0.8} & 10.1\textsubscript{±0.8}&27.6\textsubscript{±4.0} & 36.0\textsubscript{±4.5} & 26.7\textsubscript{±4.1} & 10.2\textsubscript{±1.4} & 17.4\textsubscript{±1.1} & 10.5\textsubscript{±0.7} \\
        
        \SetCell[r=2]{c}{ReAct} & Single Function & \emph{Event-Only} & 33.5\textsubscript{±0.7} & \underline{6.7}\textsubscript{±0.7}&\underline{44.3}\textsubscript{±3.9} & \underline{54.2}\textsubscript{±3.9} & \underline{41.4}\textsubscript{±1.7} & \underline{25.3}\textsubscript{±2.6} & \underline{47.4}\textsubscript{±2.4} & \underline{26.9}\textsubscript{±1.9} \\
        
        & Single Function & \emph{News-Only} & 6.1\textsubscript{±1.0} & 12.8\textsubscript{±0.6}&27.8\textsubscript{±3.1} & 25.9\textsubscript{±2.9} & 21.8\textsubscript{±2.3} & 6.3\textsubscript{±2.2} & 9.0\textsubscript{±2.0} & 5.4\textsubscript{±1.3} \\
        
        \SetCell[r=2]{c}{ReAct} & Single Function & All & \textbf{3.1}\textsubscript{±0.5} & \textbf{5.9}\textsubscript{±1.0}&\textbf{47.6}\textsubscript{±5.8} & \textbf{58.3}\textsubscript{±2.6} & \textbf{44.2}\textsubscript{±4.0} & \textbf{28.7}\textsubscript{±3.9} & \textbf{51.0}\textsubscript{±4.0} & \textbf{29.6}\textsubscript{±3.7}
 \\
        & Code Block & All & \underline{5.1}\textsubscript{±0.9} & 8.9\textsubscript{±0.5}&27.1\textsubscript{±4.0} & 38.6\textsubscript{±2.5} & 25.9\textsubscript{±2.2} & 11.6\textsubscript{±2.4} & 26.3\textsubscript{±2.0} & 12.6\textsubscript{±1.7} \\
    \bottomrule
    \end{tblr}%
    }
    \label{tab:main_strategy}
\end{table*}

For agent implementations, we consider the following two methods without tool-use:

\textbf{Direct IO} represents the standard LLM chatbot approach that provides answers without tool-use or explicit reasoning. This serves as a baseline to reflect the internal world knowledge of the LLMs when forecasting.

\textbf{Zero-Shot Chain-of-Thought (ZS-CoT)}~\citep{wei_chain--thought_2023, DBLP:conf/nips/KojimaGRMI22}: In the ZS-CoT variant, we add an instruction prompt to the LLM to encourage explicit step-by-step thinking before making the final prediction.

For Tool-Use agents, We follow 
\textbf{ReAct}~\citep{DBLP:conf/iclr/YaoZYDSN023} to interact with our provided environments through an iterative process of thinking, acting, and observing. 
We implement two variants of ReAct for tool-use: 1) ``\hlc[uclablue!20]{Single Function}'' and 2) ``\hlc[blue!30]{Code Block}''. Details are discussed in Sec.~\ref{sec:agent}.

ReAct agents can utilize tools. We thus implement different variants by restricting their access to utilize 1) \emph{News-Only} APIs; 2) \emph{Event-Only} APIs; 3) \emph{All} data classes and functions.

The experimental results in Table~\ref{tab:main_strategy} reveal several key insights into agent performance:

\textbf{1) \dataset presents a challenging task for LLM agents.}
The best agent (ReAct with ``\hlc[uclablue!20]{Single Function}'' using all APIs) for second-level relation predictions achieves a precision of $28.7$ and an F1 score of $29.6$. 
These results underscore the complexity and difficulty of the temporal forecasting tasks in \dataset and highlight the substantial room for improvement in LLM agents for event forecasting.

\textbf{2) Predicting fine-grained relations proves more difficult.}
All models exhibit higher KL divergence for quadratic than binary classes, and lower F1 scores for second-level predictions compared to first-level ones. These findings confirm that predicting fine-grained relation types is more challenging. 

\textbf{3) Diverse tool-use is critical for temporal forecasting.}
ZS-CoT and Direct-IO, which rely solely on the internal world knowledge of LLMs for forecasting without tool-use, significantly underperform the ReAct agent with full API access to the database. 
This emphasizes the importance of basing forecasting and reasoning on retrieved historical data and knowledge.
In terms of tool types, ReAct agents using \emph{News-Only} APIs perform much worse than agents with \emph{Event-Only} APIs. 
While news articles provide detailed context for events, they can also introduce noise and lead to issues such as excessively long context, posing additional challenges for LLM agents. Moreover, the agents using both types of information achieve the optimal results.

\subsection{Evaluate Forecasting with Different Base LLMs}\label{sec:exp_basemodel}

\begin{table*}[!t]
    % \vspace{-0.4cm}
    \centering
    \caption{ Experimental results on the test subset using different base LLMs and action types. The best-performing score is highlighted in \textbf{bold} and the second-best is \underline{underlined}.
    }
    \resizebox{\textwidth}{!}{%
\begin{tblr}{colspec = {ccccccccccc},
row{1-2} = {bg=gray!25},
row{5-6,9-10} = {bg=gray!10}}
    \toprule
       \SetCell[r=2]{c}{Model} & \SetCell[r=2]{c}{Training Data\\Cutoff Date} & \SetCell[r=2]{c}{Action Type} & \SetCell[r=2]{c}{Binary \\ KL ($\Downarrow$)} & \SetCell[r=2]{c}{Quad\\KL ($\Downarrow$)} & \SetCell[c=3]{c}{First-level Relation (\%)}  & & & \SetCell[c=3]{c}{Second-level Relation (\%)}\\
        \cmidrule[lr]{6-8} \cmidrule[lr]{9-11}
        & &  &  & & Pre. ($\Uparrow$)& Rec. ($\Uparrow$) & F1 ($\Uparrow$) & Pre. ($\Uparrow$) & Rec. ($\Uparrow$) & F1 ($\Uparrow$)\\\midrule
         \SetCell[r=2]{c}{\texttt{Mistral-7B}}  & \SetCell[r=2]{c}{---} & Single Function & 9.6\textsubscript{±2.4} & 14.1\textsubscript{±1.9}&27.2\textsubscript{±8.2} & 13.1\textsubscript{±3.6} & 14.4\textsubscript{±3.2} & 14.1\textsubscript{±4.4} & 7.0\textsubscript{±2.2} & 6.6\textsubscript{±2.3} \\
        & & Code Block& 9.0\textsubscript{±3.0} & 13.1\textsubscript{±2.3}&25.8\textsubscript{±6.9} & 10.4\textsubscript{±4.8} & 11.1\textsubscript{±3.8} & 10.0\textsubscript{±3.8} & 5.8\textsubscript{±4.1} & 4.6\textsubscript{±1.1}\\
        \SetCell[r=2]{c}{\texttt{gpt-3.5-turbo}}& \SetCell[r=2]{c}{{2021-09}} & Single Function & 3.9\textsubscript{±1.0} & 6.7\textsubscript{±1.7}&\underline{52.6}\textsubscript{±9.3} & 46.8\textsubscript{±8.5} & 38.2\textsubscript{±4.9} & \textbf{36.7}\textsubscript{±9.2} & 41.7\textsubscript{±7.6} & 27.9\textsubscript{±3.6}\\
        &&Code Block&3.6\textsubscript{±0.7} & 8.0\textsubscript{±1.7}&32.8\textsubscript{±4.7} & 41.0\textsubscript{±2.0} & 29.7\textsubscript{±2.9} & 14.4\textsubscript{±5.0} & 33.7\textsubscript{±5.3} & 14.4\textsubscript{±2.6}\\
        \SetCell[r=2]{c}{\texttt{gpt-4-turbo}}& \SetCell[r=2]{c}{{2023-12}} & Single Function & 2.4\textsubscript{±0.6} & 6.1\textsubscript{±0.8}&\textbf{55.0}\textsubscript{±7.5} & 43.9\textsubscript{±7.8} & 39.8\textsubscript{±4.9} & \underline{33.5}\textsubscript{±7.5} & 43.5\textsubscript{±5.4} & 30.0\textsubscript{±1.9}\\
        &&Code Block&2.0\textsubscript{±1.1} & 6.0\textsubscript{±2.0}&46.4\textsubscript{±8.7} & 51.7\textsubscript{±10.0} & 40.0\textsubscript{±8.7} & 32.7\textsubscript{±4.9} & 43.2\textsubscript{±9.5} & 28.7\textsubscript{±5.0}\\
        \SetCell[r=2]{c}{\texttt{gpt-4o}}& \SetCell[r=2]{c}{{2023-10}} & Single Function & \underline{1.8}\textsubscript{±1.0} & \underline{4.4}\textsubscript{±1.8}&48.7\textsubscript{±8.7} & \underline{63.7}\textsubscript{±8.8} & \textbf{45.5}\textsubscript{±4.6} & 31.2\textsubscript{±6.7} & \underline{59.2}\textsubscript{±7.3} & \underline{32.6}\textsubscript{±2.7}\\
        &&Code Block&\textbf{1.6}\textsubscript{±1.1} & \textbf{4.3}\textsubscript{±1.9}&42.3\textsubscript{±9.4} & \textbf{64.5}\textsubscript{±9.5} & \underline{44.1}\textsubscript{±8.1} & 29.1\textsubscript{±7.3} & \textbf{59.6}\textsubscript{±7.6} & \textbf{32.8}\textsubscript{±6.4}
\\
    \bottomrule
    \end{tblr}%
    }
    \label{tab:main_res}
\end{table*}

We then investigate the role of the underlying LLMs in the agent's performance. We evaluate both open-sourced LLM \texttt{Mistral-7B-Instruct-v0.2}~\citep{jiang2023mistral} (run on a single NVIDIA RTX A6000 GPU), as well as close-sourced LLMs including \texttt{gpt-3.5-turbo-0125}~\citep{gpt-3.5}, \texttt{gpt-4-turbo-2024-04-09}~\citep{achiam2023gpt}, and the recently released \texttt{gpt-4o-2024-05-13}~\citep{gpt4o}. 
Comparisons are done on a data-balanced test subset comprising 100 queries, with all models evaluated under the ReAct framework, allowing access to all APIs. 
The action types can be either ``\hlc[uclablue!20]{Single Function}'' or ``\hlc[blue!20]{Code Block}'' with a maximum tool call limit set to 20 steps. 
All models use the same prompt content and structure shown in Appendix~\ref{append:prompts}. 
The experimental results are presented in Table~\ref{tab:main_res}, and a further analysis of the agent's final status is provided in Appendix~\ref{append:final_status}. We observe the following findings:

\textbf{1) Code Block benefits stronger LLMs but hurts weaker models:} 
The ``\hlc[blue!20]{Code Block}'' action type provides greater flexibility than the ``\hlc[uclablue!20]{Single Function}''. However, its benefits are not uniformly achieved across all models. \texttt{Mistral-7B} and \texttt{gpt-3.5-turbo} show reduced performance with Code Block, while stronger models like \texttt{gpt-4-turbo} and \texttt{gpt-4o} gains improvements from it. 
This indicates that the ability to generate coherent and practical long Code Blocks is a distinguishing factor that determines LLMs as reliable forecasting agents.

\textbf{2) GPT-4o consistently outperforms other models:} \texttt{gpt-4o}, the most recent LLM, consistently achieves the highest performance across different metrics. 
Notably, for second-level relation prediction, \texttt{gpt-4o} achieves F1 scores of $32.6$ and $32.8$ using ``\hlc[uclablue!20]{Single Function}'' and ``\hlc[blue!20]{Code Block}'', surpassing all other models including \texttt{gpt-4-turbo-2024-04-09}. This indicates that \dataset is hard enough, and can effectively distinguish different LLMs' advanced reasoning capabilities.

\begin{figure}[t!]
    % \vspace{-0.1cm}
     \centering
     \begin{subfigure}[b]{0.46\textwidth}
         \centering
         \includegraphics[width=\textwidth]{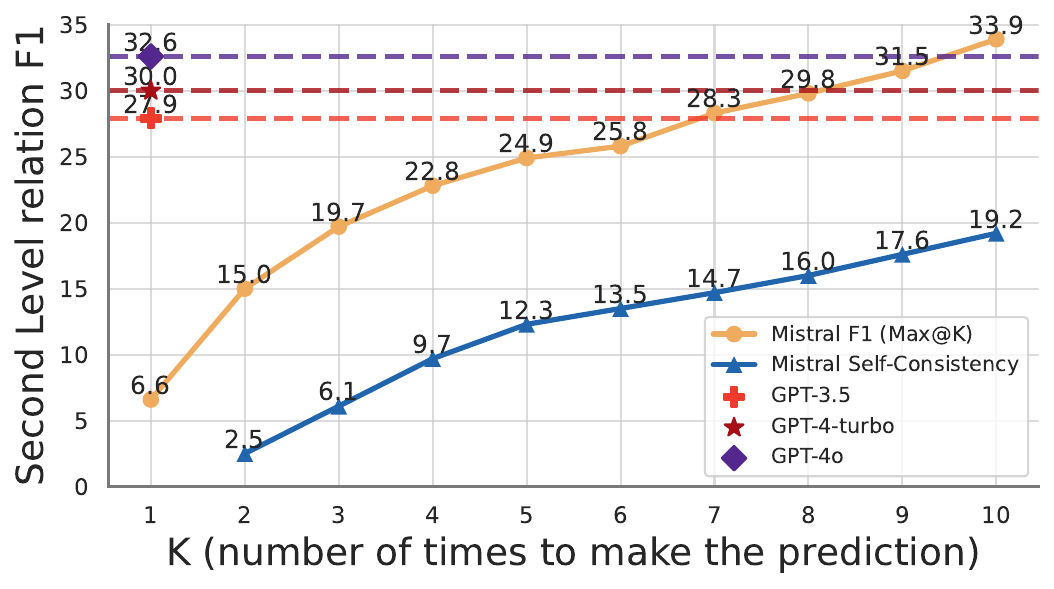}
     \end{subfigure}
     \hfill
     \begin{subfigure}[b]{0.49\textwidth}
         \centering
         \includegraphics[width=\textwidth]{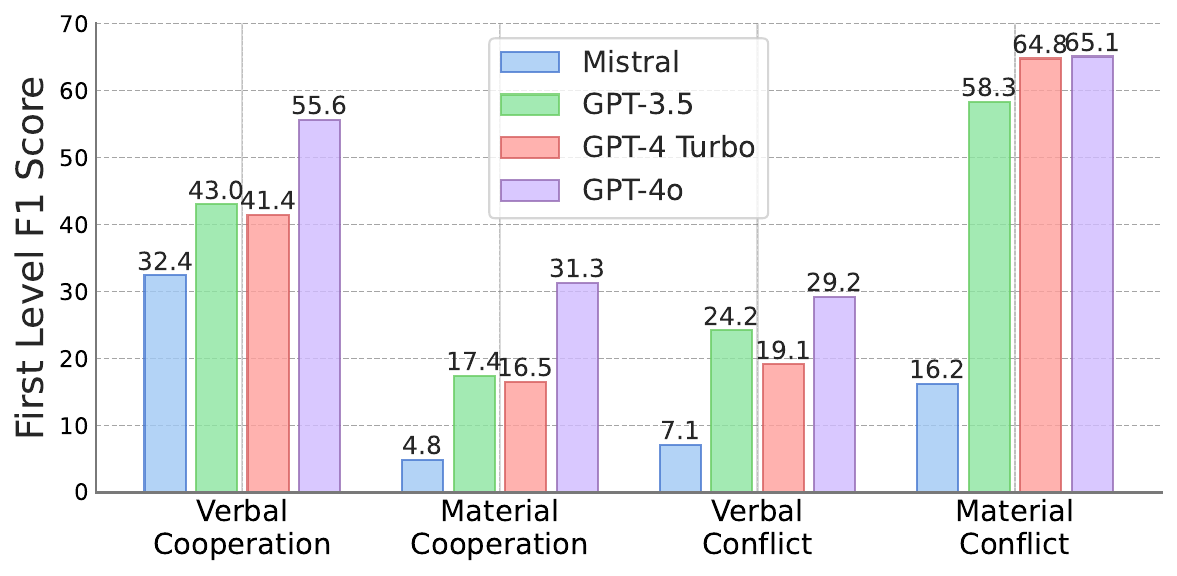}
     \end{subfigure}
     \caption{a) Self-consistency of Mistral-7B model increases with more samples. b) F1 scores of different base LLM agents on relation prediction, categorized based on the quadratic classes.}
     \label{fig:multi_analysis}
\end{figure}

\textbf{Can self-consistency make a small model stronger?}
It is evident that stronger LLMs have better agent performance; however, can we enhance a weaker LLM to achieve comparable forecasting performance using inference-time computation? 
To explore this, we take \texttt{Mistral-7B} as the base LLM with ReAct using the ``\hlc[uclablue!20]{Single Function}'' strategy. 
For each query, we perform multiple sampling at a temperature of $0.4$. 
We then consider a variant of \textit{self-consistency}, which only keeps entries appearing more than twice. 
We also calculate F1 (Max@K), which assesses the F1 score for each instance and keeps the maximum score across all rounds. 
As shown in Figure~\ref{fig:multi_analysis}a, with more samples, the performance of Mistral-7B significantly improves. 
Initially, a single sample achieves an F1 score of $6.6$ for Mistral-7B, which is considerably lower than the scores achieved by larger models. 
However, as more solutions being sampled, the F1 (Max@K) and self-consistency of Mistral-7B improve progressively. By the 10th samples, the Max@K reaches $33.9$, even surpassing \texttt{gpt-4o}. 
This result highlights the potential of the inference-time search methods like self-consistency to push the boundaries of smaller language models in event forecasting.

\begin{figure}[t!]
     \centering
     % \vspace{-0.5cm}
     \begin{subfigure}[b]{0.52\textwidth}
         \centering
         \includegraphics[width=\textwidth]{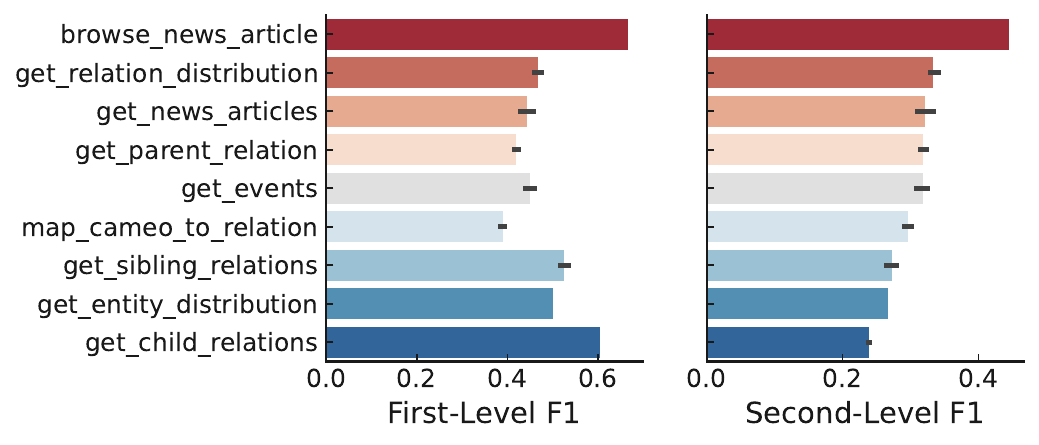}
         \label{fig:single_function}
     \end{subfigure}
     \hfill
     \begin{subfigure}[b]{0.47\textwidth}
         \centering
         \includegraphics[width=\textwidth]{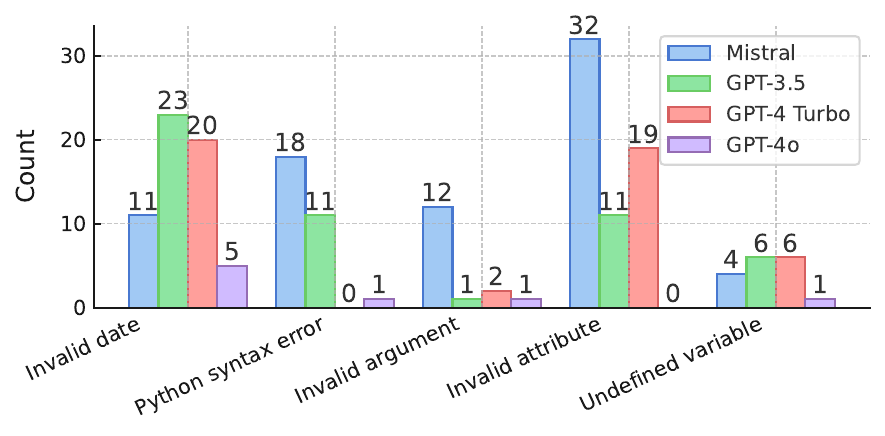}
        \label{fig:error_analysis}
     \end{subfigure}
     \vspace{-0.8cm}
     \caption{a) F1 Accuracy for each API function. b) Code execution error analysis for different LLMs.}
     \label{fig:single_error}
\end{figure}

\textbf{Code execution error analysis.} Our implemented agents interact with tools via code but often encounter execution error. 
We summarize the error types for different LLMs in Figure~\ref{fig:single_error}b. 
We can see that invalid dates and invalid attributes are the two most frequent errors, with even \texttt{gpt-4-turbo} exhibiting a frequency of such errors. 
Notably, \texttt{gpt-4o} makes significantly fewer execution errors. 
This enhanced code generation capability contributes to its superior performance for event forecasting.

\label{sec:exp_rounds}

\subsection{Analyzing Agent Behaviours}\label{sec:analysis}

\begin{figure}[t!]
     \centering
     \begin{subfigure}[b]{0.48\textwidth}
         \centering
         \includegraphics[width=\textwidth]{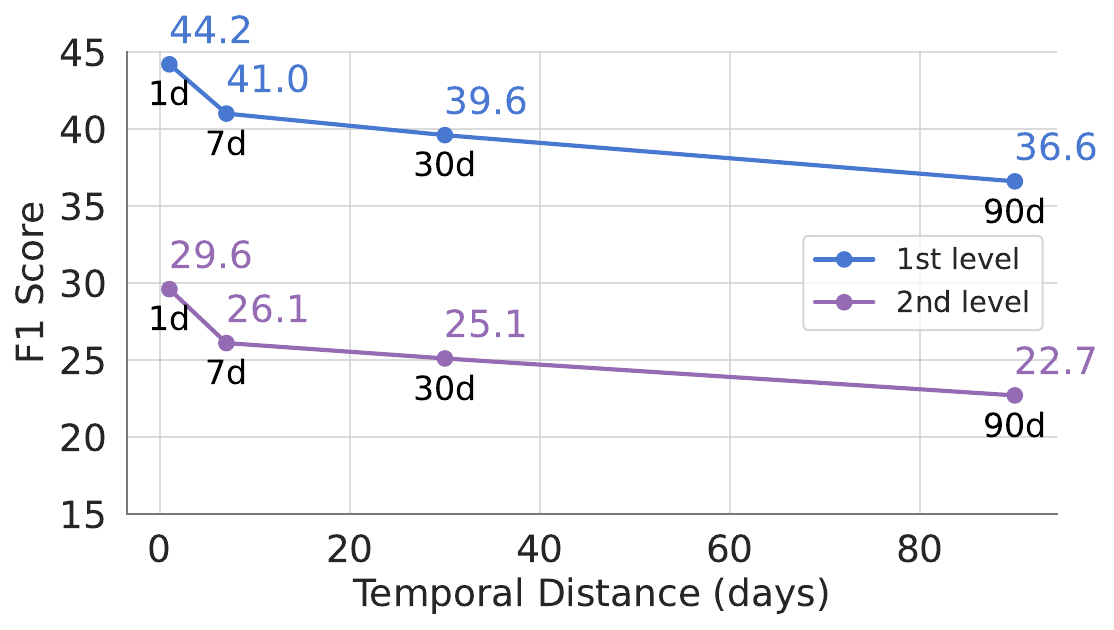}
          % \caption{F1 score of the LLM agent decrease when the temporal distance increases.}
         % \label{fig:temporal_distance_f1}
     \end{subfigure}
     \hfill
     \begin{subfigure}[b]{0.46\textwidth}
         \centering
         \includegraphics[width=\textwidth]{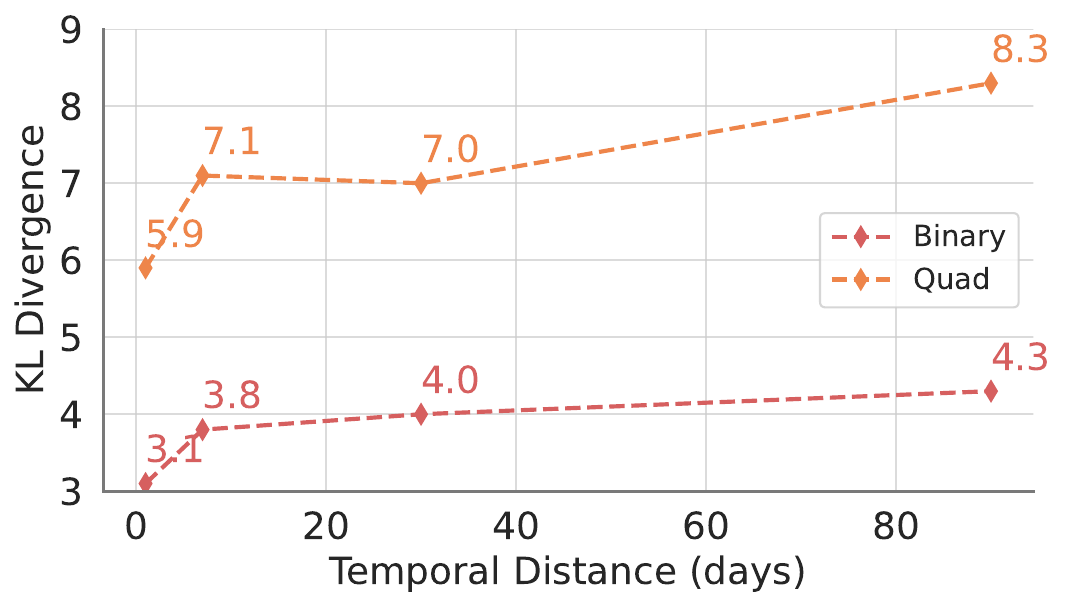}
        % \caption{KL divergence of the LLM agent's predictions increases with larger temporal distances.}
        % \label{fig:temporal_distance_kl}
     \end{subfigure}
     % \vspace{-0.1cm}
     \caption{Evaluation of LLM Agents in different temporal distances of the forecasting event.}
     \label{fig:temporal}
\end{figure}

\begin{figure}[t!]
     \centering
     \begin{subfigure}[b]{0.46\textwidth}
         \centering
         \includegraphics[width=\textwidth]{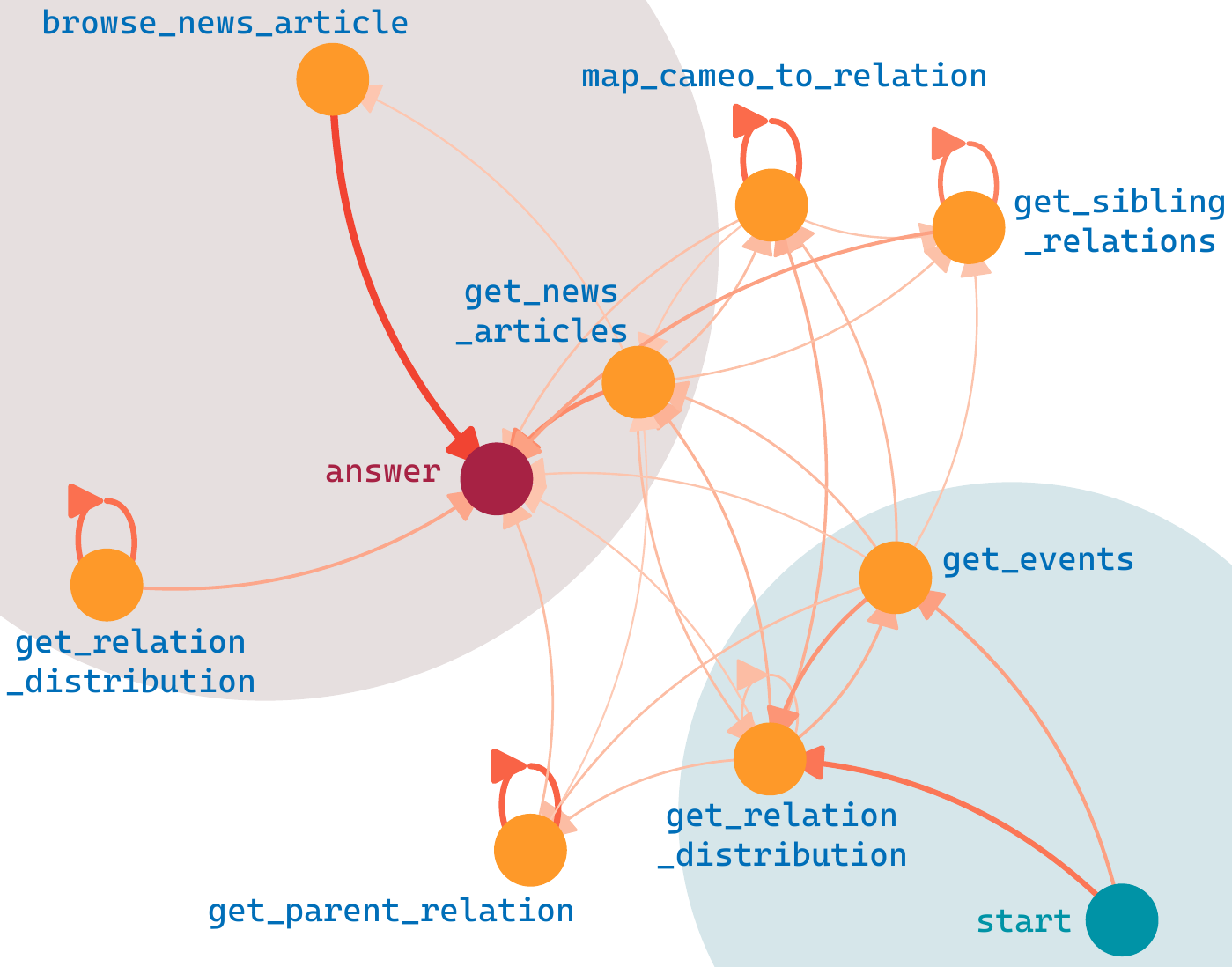}
     \end{subfigure}
     \hfill
     \begin{subfigure}[b]{0.46\textwidth}
         \centering
         \includegraphics[width=\textwidth]{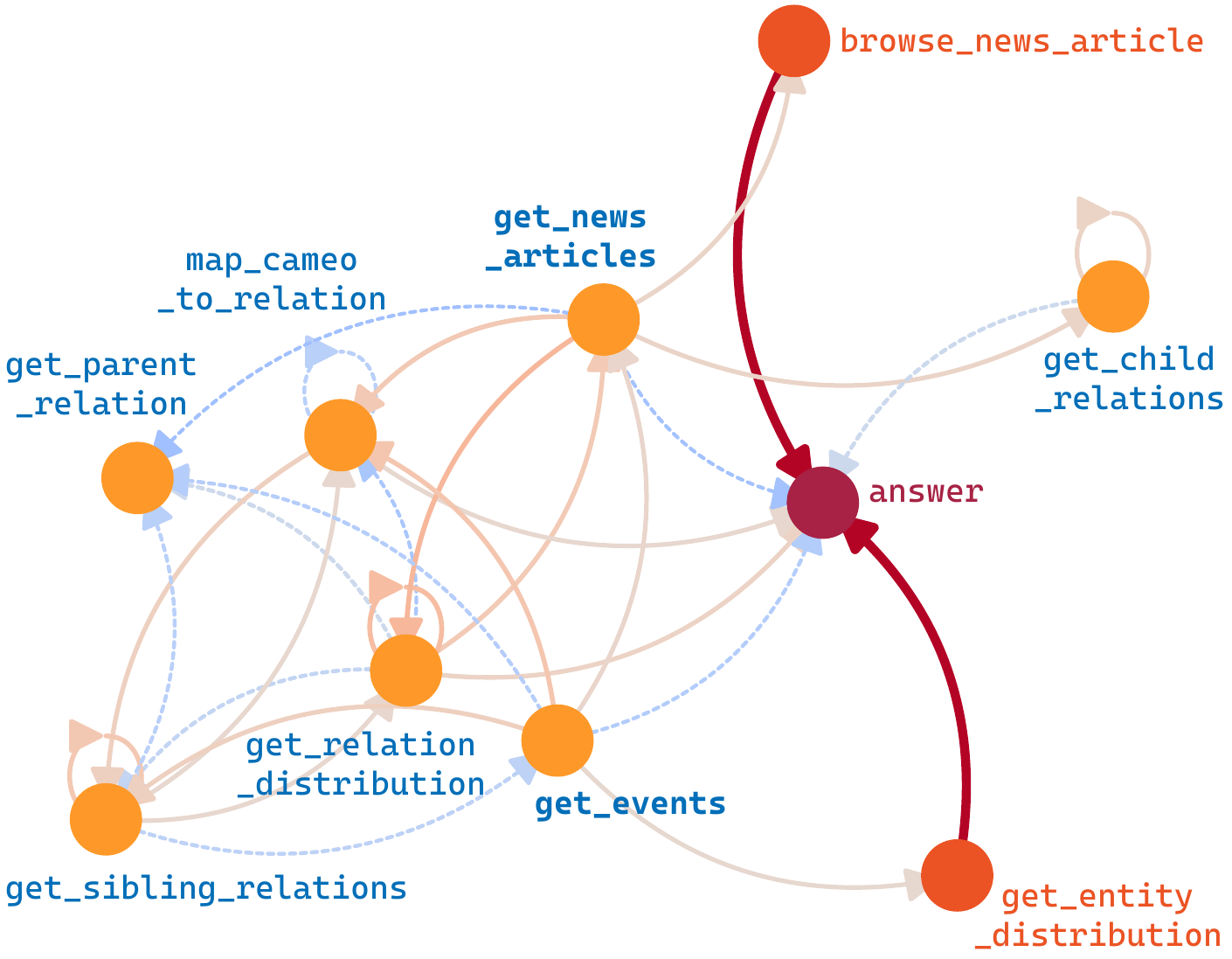}
     \end{subfigure}
     \caption{Action order analysis in LLM agents. a) Tool-Use Transition Graph of called API functions. Edge thickness indicates transition frequency. b) Freq.(correct) - Freq.(incorrect), in which red (blue) edges indicate positive (negative) contributions. }
     \label{fig:transition}
\end{figure}
\textbf{Impact of temporal distance of the forecasting target.}
\label{sec:exp_distance}
Our defined event forecasting task varies by temporal distance $l$, which specifies how far into the future we want to predict. 
We thus conduct an ablation study with $l$ set to 1, 7, 30, and 90 days. 
Specifically, we fix the query event date and limit the accessible data to $l$ days prior to the query event date. 
The experimental results depicted in Figure~\ref{fig:temporal} reveal a clear trend: as the temporal distance increases, the F1 score decreases and KL-divergence increases. 
This indicates that the agent's ability to provide accurate predictions diminishes for events further in the future. 
When the temporal distance is small, such as 1 or 7 days, the agent has access to more recent and relevant information, providing a strong signal (e.g., human experts' analysis) for making accurate predictions.  
Thus, to comprehensively benchmark the forecasting capabilities of LLM agents, we should focus on long-term predictions such as those spanning 30 or 90 days. 
These longer durations require the agents to capture and anticipate potential trend shifts, which may be influenced by a broader range of factors and more complex dependencies.

\textbf{Forecasting accuracy on different relation types.}
We further split the datasets into distinct quadratic relation classes and compute the F1 score for each class, as illustrated in Figure~\ref{fig:multi_analysis}b. 
The results show that all models exhibit significantly higher performance for ``verbal cooperation'' and ``material conflict'', while lower in the other two categories.
Several factors contribute to these observations. 
First, ``verbal cooperation'' events are more prevalent in the dataset, allowing the model to retrieve more such historical events. Second, ``material conflicts'' has a consistent pattern of extended duration within the same set of countries. 
Conversely, events categorized under ``material cooperation'' and ``verbal conflicts'', such as ``\texttt{057}: \texttt{Sign formal agreement}'' and ``\texttt{084}: \texttt{Return or release}'', tend to be more abrupt and unpredictable, demanding subtle trend analysis and contextual knowledge, leading to lower performance in these categories.
These observations highlight the need for LLMs capable of understanding the nuances and complexities of different event types.

\textbf{How tool-use ordering influences forecasting.}
We further investigate the impact of action order on the GPT-4o agent in ``\hlc[uclablue!20]{Single Function}'' mode. 
Figure~\ref{fig:transition} shows the transition graph from the initial query to the correct final answer, with thicker edges indicating more frequent transitions. Typically, the agent begins with \texttt{get\_relation\_distribution} or \texttt{get\_event} to gather an initial set of recent and frequent events for key information, and often concludes with \texttt{browse\_news\_article} and \texttt{get\_news\_articles}, which retrieve news content to make accurate forecasts. 
To further understand whether each function helps forecasting, we subtract the frequency of incorrect predictions from those of correct predictions, as shown in Figure~\ref{fig:transition}b. 
Here, {\color{red}red} edges represent sequences typically leading to accurate outcomes, and {\color{blue}blue} edges indicate error-prone paths. 
Actions like \texttt{browse\_news\_article} and \texttt{get\_entity\_distribution} typically result in correct answers more frequently. Notably, \texttt{get\_news\_articles} has a direct {\color{blue}blue} link to the answer, suggesting that this function leads more often to incorrect answers because it only returns news titles, which are too vague for accurate prediction. However, when followed by \texttt{browse\_news\_article} and then providing the answer, the agent is more likely to produce correct outcomes. 
Similar patterns are observed with \texttt{get\_event}, where adding \texttt{get\_entity\_distribution} turns a negative link to a positive one to the answer. 
 Figure~\ref{fig:single_error}a further demonstrates how each function contributes to the final performance, 
showing that for the first-level prediction, \texttt{get\_\{child/sibling\}\_relation} are more useful.
These results emphasize the importance for strategic action planning in LLM agents for effective temporal forecasting. 

\section{Related Work}

\begin{table*}[t!]
\caption{Comparison of \dataset with other temporal reasoning benchmarks. ``Method'' refers to the methodology of original and recent models evaluated on the benchmark. ICL stands for LLM w/ In-Context Learning, FT stands for Task-Specific Fine-Tuning.} 
\resizebox{\textwidth}{!}{%
\begin{tblr}{
colspec = {ccccccccc},
row{1-2} = {bg=gray!25},
row{even[3-13]} = {bg=gray!5}
}
\toprule
\SetCell[r=2]{c}Benchmark & \SetCell[c=2]{c}{Temporal Reasoning} &   &\SetCell[r=2]{c}Task Format& \SetCell[c=3]{c}{Information}& & & &\SetCell[r=2]{c}Method\\
\cmidrule[lr]{2-3}\cmidrule[lr]{5-8}
  &  Understanding & Forecasting & & Time Series & KG &  Textual  & API  & \\\midrule
TempQuestions~\citep{jia_tempquestions_2018} & \checkmark &      &   KGQA      &             &\checkmark&& & KGQA Systems     \\
CRONQuestions~\citep{saxena_question_2021} & \checkmark &      &   KGQA      &             &\checkmark&&& Bert-based FT     \\
TempoQR~\citep{mavromatis_tempoqr_2021} & \checkmark &      &   KGQA      &             &\checkmark&& &Bert-based FT    \\
TORQUE~\citep{ning_torque_2020}             & \checkmark &       & QA  &            &&\checkmark &  &Bert-based FT   \\
TempReason~\citep{tan_towards_2023}             & \checkmark &     &    QA   &          &\checkmark & \checkmark& & {ICL + Task-specific FT}  \\
TRAM~\citep{wang_tram_2024}              & \checkmark &    &   MCQ     &   & &\checkmark & & {ICL + Bert-based FT}      \\
TCELongBench~\citep{zhang_analyzing_2024}            & \checkmark & \checkmark      &  Ordering QA; MCQ &      & &\checkmark& &ICL   \\
ForecastQA~\citep{jin_forecastqa_2021}&  &    \checkmark   &  MCQ &      & &\checkmark&  &   Bert-based FT  \\
Approach~\cite{halawi_approaching_2024}&  & \checkmark   &   QA     &    &&\checkmark &  & ICL     \\
IntervalQA~\citep{zou_forecasting_2022} &  & \checkmark   &   Numerical QA; MCQ     &    \checkmark  &&\checkmark &  & ICL     \\
TKGs~\citep{icews, leetaru_gdelt_nodate}&       &    \checkmark   &  Link Prediction           &   & \checkmark& & &  {Graph FT + ICL + Task-specific FT}   \\ 
\midrule
\dataset  &  &   \checkmark    &    Relation List  &     &\checkmark &\checkmark&    \checkmark  &LLM Agent       \\
\bottomrule
\end{tblr}%
}
\label{tab:compare}
\end{table*}

In recent years, various benchmarks have been developed to evaluate temporal reasoning capabilities in AI systems, focusing primarily on question answering and link prediction. 
\dataset distinguishes itself by assessing LLMs in the \textit{forecasting} task, and further employing a relational task format, incorporating diverse information sources of knowledge graphs, text, and code-based APIs, and utilizing an agent-based methodology that supports intermediate reasoning steps. We summarize our differences to these existing benchmarks in Table~\ref{tab:compare}, and provide further discussions from the following aspects and more in Appendix~\ref{append:related_work}.

\subsection{Temporal Reasoning Benchmarks}

Many benchmarks sensing the temporal reasoning ability of AI models have been constructed, but they have different focuses and settings with \dataset, particularly in terms of task, information, and method, as shown in Table~\ref{tab:compare}. One line of benchmarks focuses on the \textit{temporal understanding} ability of the model~\citep{jia_tempquestions_2018,saxena_question_2021,mavromatis_tempoqr_2021,ning_torque_2020,tan_towards_2023,wang_tram_2024}, such as understanding the temporal relations between available facts in knowledge graphs (KGs) or text, either a short piece of text or a document corpus. While the \textit{temporal forecasting} task largely differs from understanding, where the reasoning target is \textbf{unseen} in the database for the model, and as such, the model has to not only understand but to reason. For forecasting, there are two main task formulations among previous benchmarks: the QA task format for benchmarks with history information represented in textual format~\citep{zou_forecasting_2022, zhang_analyzing_2024, jin_forecastqa_2021, halawi_approaching_2024, schoenegger_wisdom_2024}, and graph link prediction task format for temporal knowledge graph(TKG)-based benchmarks~\citep{icews, leetaru_gdelt_nodate}.  However, the uniformat of information sources either lacks of support to clearly structural facts or contextual detail for the model to perform advanced reasoning, while in \dataset, we provide both information sources. Additionally, we provide carefully constructed API with various data classes and functions that access to various part of the data. With the flexibility provided by code generation, the model is exposed to a broader and more flexible range of information. More importantly, \dataset distinguishes itself by introducing an agentic environment specifically designed to evaluate LLM agents in the forecasting task. This represents a significant departure from previous work, which has not explored or even considered the potential of performing temporal forecasting tasks using LLM agents.

\subsection{Evaluation of Language Agents}

Previous research has investigated the performance of LLM agents in a variety of domains, including arithmetic reasoning focused on obtaining correct solutions~\cite{roy-roth-2015-solving,cobbe2021training,patel-etal-2021-nlp}, proficiency assessment in utilizing tools and reporting results~\cite{li2023api,xu2023tool,zhuang2023toolqa}, evaluation of web navigation skills to find specific websites~\cite{deng2023mindweb,zhou2023webarena,liu2023agentbench}, and planning travel itineraries under given constraints~\cite{xie_travelplanner_2024}.
However, these evaluations do not fully address the challenges posed by tasks involving complex international events with diverse information formats and temporal attributes. \dataset presents a unique task in this context, where the agent must navigate and reason over the structured events and textual news articles with temporal information. This setup requires the agent to effectively handle multilateral relationships and information spanning different time periods. Furthermore, \dataset assesses the LLM agent's ability to reason and predict information that not be directly available in the provided database. This adds an additional layer of complexity, as the agent must leverage its understanding of the available information to make informed predictions about future events or fill in missing details. By evaluating LLM agents in this challenging setting, \dataset provides valuable insights into their capacity to process and reason over complex, temporally-structured information and their ability to generate accurate predictions based on incomplete data.

\subsection{LLMs for Tool-Use}
Large Language Models (LLMs) have demonstrated remarkable language understanding~\cite{radford2018improving} and reasoning capabilities~\cite{DBLP:conf/nips/Wei0SBIXCLZ22}. However, they also possess inherent limitations, such as their inability to provide up-to-date responses based on external knowledge or to perform complex mathematical reasoning. In response to these challenges, recent advancements have seen the integration of LLMs with various external tools~\cite{mialon2023augmented}. Notable examples include TALM~\cite{parisi2022talm} and ToolFormer~\cite{schick2023toolformer}, which utilize in-context learning to enhance the model's ability to leverage different tools in tasks like question answering and mathematical reasoning. Chameleon~\cite{lu2023chameleon} employs an LLM as a natural language planner to deduce the optimal sequence of tools to be used, subsequently executing these tools to generate the final output. AVIS~\cite{DBLP:conf/nips/HuI0CSRSF23} employs dynamic tree search to synthesize the most effective tool-use sequence. ToolkenGPT~\cite{DBLP:conf/nips/HaoLWH23} integrates tool-use operators as special tokens and trains the model through sequence-to-sequence training. ToolLLM~\cite{DBLP:journals/corr/abs-2307-16789} introduces an instruction tuning dataset encompassing over 16,000 real-world APIs, significantly enhancing the model's capability to utilize these tools effectively.

\section{Conclusion and Limitation}
\label{conclusion}

In conclusion, we introduce \dataset, a benchmark constructed for evaluating LLM agents in temporal forecasting international event with tools (APIs) to access an extensive historical event and news database. The results reveals the complexity and difficulty for current LLM Agents in generating contextually and syntax-wise correct code and performing complex temporal reasoning over the multi-party and multi-timestamped data for an effective prediction, highlighting a substantial space for further effects in this direction.

Our work also has several \textbf{limitations}: 1) Only a few representative LLMs were tested, leaving broader evaluations with more open-sourced models;
2) The current API is basic, primarily providing functions for counting, listing, and statistical distribution. Future enhancements could include time series analysis and the ability for agents to add new functions for more comprehensive analysis.
3) The study was limited by cost of API usage, so we only conduct small number of experimental rounds and result in high variance. More extensive testing is suggested for future studies to achieve more stable results. Full discussion of limitation can be found in Appendix~\ref{sec:limitation}. In the future, we plan to incorporate APIs to support more knowledge sources such as time-series and multimodal information, and consider testing more open-source LLMs and agent architectures.

% \newpage
\section*{Acknowledgements}
Research was supported in part by NIH U24DK097771, U54HG012517, NSF 1829071, 2106859, 2119643, 2200274, 2202693, and 2312501, DARPA HR00112490370, and Optum Lab.

\bibliographystyle{unsrtnat}
{
\small
\bibliography{_reference}
}

\newpage

\newpage
% \nolinenumbers
\appendix
% Reset depth to add sections and subsections to ToC
\addtocontents{toc}{\protect\setcounter{tocdepth}{3}}

% Setting colorlinks=black just for the table of contents
\hypersetup{linkcolor=black}

\renewcommand{\contentsname}{Appendix}

\tableofcontents % Lists only the appendix sections and subsections

% Revert to original link colors after table of contents
\hypersetup{linkcolor=red}

\newpage

\section{Reproducibility Statement}

We provide detailed information and explanation of our experimental setup, dataset, and evaluation metrics, ensuring reproducibility. Specifically, we describe the following:
\begin{itemize}
    \item Experimental setup (Sec. \ref{sec:exp}): Detailed descriptions of the hardware and software configurations, including libraries and tools used.
    \item Dataset (Sec. \ref{sec:dataset} \& Appendix \ref{sec:construction}): Comprehensive information on the dataset used, including sources, preprocessing steps, and any techniques applied.
    \item Evaluation Metrics (Sec. \ref{sec:evalmetric}): Clear definitions for the metrics chosen to evaluate performance. 
\end{itemize}

Additionally, we commit to releasing the following resources for the replication of our results:
\begin{itemize}
    \item Codebase: The complete codebase, including scripts for data preprocessing, model serving, and evaluation. This is made available on a public repository (\url{https://github.com/yecchen/MIRAI}).
    \item Dataset: The processed dataset, along with detailed instructions on how to construct the dataset (\url{https://drive.google.com/file/d/1xmSEHZ_wqtBu1AwLpJ8wCDYmT-jRpfrN/view?usp=sharing}).
\end{itemize}

By providing these resources, we aim to ensure that our work is fully reproducible and can be independently verified by the research community.

\section{Datasheet for MIRAI}
\label{supp:datasheet}

\subsection{Motivation}
\begin{enumerate}
    \item \textbf{For what purpose was the dataset created?} \\
    The MIRAI dataset was created to evaluate different capabilities of Large Language Model (LLM) agents in forecasting international events.
    \item \textbf{Who created the dataset and on behalf of which entity?} \\
    The dataset was developed by a group of university researchers studying LLM reasoning. All rights are held by the individuals themselves, rather than by third-party stakeholders.
    \item \textbf{Who funded the creation of the dataset?} \\
    The academic institutions support the university reasearch lab that the authors affiliate in.
\end{enumerate}

\subsection{Distribution}
\begin{enumerate}
    \item \textbf{Will the dataset be distributed to third parties outside of the entity?} \\
    Yes, the dataset, evaluation codes and leaderboards are intended to be publicly available to foster future research and development.
    \item \textbf{How will the dataset be distributed?} \\
    The database and codebase are currently available on \href{https://drive.google.com/file/d/1xmSEHZ_wqtBu1AwLpJ8wCDYmT-jRpfrN/view?usp=sharing}{Google Drive} and \href{https://github.com/yecchen/MIRAI}{Github}.
    \item \textbf{Have any third parties imposed IP-based or other restrictions on the data associated with the instances?} \\
    No, there are no IP-based or other restrictions on the data: MIRAI is curated based on the GDELT\footnote{\href{https://www.gdeltproject.org/}{GDELT Project: https://www.gdeltproject.org/}} Event Database, which is an open platform for research and analysis of global society and all datasets released by the GDELT Project are available for unlimited and unrestricted use for any academic, commercial, or governmental use of any kind without fee; Data are allowed for any redistribution, rehost, republish, and mirror of the GDELT datasets in any form, with necessary citations\footnote{\href{https://www.gdeltproject.org/about.html\#termsofuse}{GDELT Term of Use: https://www.gdeltproject.org/about.html\#termsofuse}}.
    \item \textbf{Do any export controls or other regulatory restrictions apply to the dataset or to individual instances?} \\
    No, there are no export controls or other regulatory restrictions applied to this dataset.
    \item \textbf{When will the dataset be distributed?} \\
    The dataset has been made publicly available with the \href{https://mirai-llm.github.io/}{project website}, \href{https://drive.google.com/file/d/1xmSEHZ_wqtBu1AwLpJ8wCDYmT-jRpfrN/view?usp=sharing}{Google Drive}, and \href{https://github.com/yecchen/MIRAI}{Github}.
    \item \textbf{Will the dataset be distributed under a copyright or other IP license, and/or under applicable terms of use (ToU)?} \\
    The dataset will be distributed under the CC BY-NC 4.0 license, allowing for use and distribution for non-commercial purposes with appropriate attribution.
\end{enumerate}

\subsection{Maintenance}
\label{supp:maintenance}
\begin{enumerate}
    \item \textbf{Who will be supporting/hosting/maintaining the dataset?} \\
    The dataset maintenance will be supporting/hosting/maintaining by the authors.
    \item \textbf{How can the owner/curator/manager of the dataset be contacted?} \\
    The owner/curator/manager of the dataset can be contacted through the authors' emails.
    \item \textbf{Will the dataset be updated? (e.g., to correct labeling errors, add new instances, delete instances)?} \\
    Yes, the dataset will be updated whenever necessary to ensure accuracy, and announcements will be made accordingly.
    \item \textbf{If the dataset relates to people, are there applicable limits on the retention of the data associated with the instances (e.g., were the individuals in question told that their data would be retained for a fixed period of time and then deleted?)} \\
    NA. This dataset does not contain data related to individuals or personal identifiers.
    \item \textbf{Will older version of the dataset continue to be supported/hosted/maintained?}\\
    NA. Currently, there are no older versions of the dataset, as this is its first release. As the dataset evolves, the possibility of supporting and maintaining older versions will be considered based on community needs and resource availability.
    \item \textbf{If others want to extend/augment/build on/contribute to the dataset, is there a mechanism for them to do so?}\\
    Researchers and developers interested in extending, augmenting, or contributing to the dataset are encouraged to submit their changes through GitHub pull requests. For additional inquiries or detailed discussions, contacting the authors via email is recommended.
\end{enumerate}

\subsection{Composition}
\begin{enumerate}
    \item \textbf{What do the instances that comprise the dataset represent?} \\
    Each instance in the dataset represents a structured record of international events, including the date, involved countries, and type of event, along with associated news articles and metadata.
    \item \textbf{How many instances are there in total?} \\
    The database comprises 991,759 GDELT event records from January 2023 to November 2023, corresponding to 59,161 unique events and 296,630 news articles. The test set contains 705 relation prediction queries, and the test subset contains 100 queries.
    \item \textbf{Does the dataset contain all possible instances or is it a sample of instances from a larger set?} \\
    The dataset represents a curated sample from the entire GDELT database. It has been created through meticulously designed data cleaning and preprocessing steps aimed at enhancing the quality and reliability of the event data.
    \item \textbf{Is there a label or target associated with each instance?} \\
    Yes, each instance in the dataset is an event labeled with a relation type derived from the CAMEO\footnote{\href{https://parusanalytics.com/eventdata/data.dir/cameo.html}{Conflict and Mediation Event Observations (CAMEO): https://parusanalytics.com/eventdata/data.dir/cameo.html}} event taxonomy.
    \item \textbf{Is any information missing from individual instances?} \\
    No, all instances are complete with all available information.
    \item \textbf{Are there recommended data splits (e.g., training, development/validation, testing)?} \\
    The dataset includes records from January 2023 to November 2023. The recommended testing set is constructed based on records from November 2023. For testing, each instance utilizes all preceding records as its historical dataset to ensure accuracy and relevance in analysis.
    \item \textbf{Are there any errors, sources of noise, or redundancies in the dataset?} \\
    The dataset has undergone extensive cleaning and structuring to minimize errors and noise. However, residual noise from the original GDELT database may still be present.
    \item \textbf{Is the dataset self-contained, or does it link to or otherwise rely on external resources (e.g., websites, tweets, other datasets)?} \\
    While the dataset is primarily self-contained, it includes URLs to news articles, providing links to external resources for further context and verification.
    \item \textbf{Does the dataset contain data that might be considered confidential?} \\
    No, the dataset contains publicly available data and does not include confidential information.
    \item \textbf{Does the dataset contain data that, if viewed directly, might be offensive, insulting, threatening, or might otherwise cause anxiety?} \\
    The dataset may include descriptions of sensitive events, such as global conflicts, due to its focus on international relations. However, thorough checks such as flagging word ratios are employed during the data cleaning process to identify and potentially exclude inappropriate paragraphs or entire news articles and events. This ensures the minimization of distressing content while maintaining the integrity and relevance of the dataset for academic study.
\end{enumerate}

\subsection{Collection Process}
\begin{enumerate}
    \item \textbf{How was the data associated with each instance acquired?} \\
    Data for each instance was sourced from the GDELT project, which aggregates global event data and news articles from various worldwide media. Detailed information can be found in Section~\ref {sec:database-construction} and Appendix~\ref{sec:construction} of the paper.
    \item \textbf{What mechanisms or procedures were used to collect the data (e.g., hardware apparatus or sensor, manual human curation, software program, software API)? }\\
    Data collection was facilitated through software programs that aggregate event data and news articles from various sources. This automated collection is followed by data cleaning processes to enhance completeness and reliability. For a detailed description of the collection and cleaning methods, please refer to Section~\ref{sec:database-construction} and Appendix~\ref{sec:construction} of the paper. All dataset construction scripts are available in \href{https://github.com/yecchen/MIRAI}{Github} and also described in the README.
    \item \textbf{Who was involved in the data collection process? (e.g., students, crowdworkers, contractors) and how were they compensated (e.g., how much were crowdworkers paid)} \\
    The data collection and preprocessing were conducted by the authors, who are university researchers. This work was performed as part of their academic research activities.
    \item \textbf{Does the dataset relate to people?} \\
    No. The dataset does not contain personal data or directly relate to individual people.
    \item \textbf{Did you collect the data from the individuals in questions directly, or obtain it via third parties or other sources (e.g., websites)? } \\
    The dataset does not involve data collected from individuals directly. Instead, it is curated from the GDELT project, which aggregates information from various global news media sources.
\end{enumerate}

\subsection{Uses}
\begin{enumerate}
    \item \textbf{Has the dataset been used for any tasks already?} \\
    The dataset has not been used for any tasks other than the ones proposed and examined in this current paper, specifically for benchmarking the forecasting capabilities of LLM agents in predicting international relations.
    \item \textbf{What (other) tasks could the dataset be used for?} \\
    In addition to benchmarking LLM agents, the dataset could be valuable for research in geopolitics, the development of other event prediction algorithms, sentiment analysis of international events, and trend analysis in global political dynamics.
    \item \textbf{Is there anything about the composition of the dataset or the way it was collected and preprocessed/cleaned/labeled that might impact future uses?}\\
    The dataset was meticulously collected and preprocessed to ensure high quality and reliability. However, the reliance on the GDELT project as the primary data source and the specific cleaning process applied might influence its applicability to certain tasks. Users should consider these factors when applying the dataset to different research areas or methodologies.
    \item \textbf{Are there tasks for which the dataset should not be used?}\\
    The dataset should not be used for any tasks that violate the terms of use associated with the GDELT project.
\end{enumerate}

\section{Limitations} \label{sec:limitation}
We acknowledge a few limitations in our current work. Firstly, the forecasting events in the test set might have been exposed to the most recently published LLMs, such as the \texttt{gpt-4-turbo-2024-04-09}, which has training data up to December 2023. While the background knowledge of LLMs about international relations matters for accurate forecasting, this benchmark still challenges them in retrieving grounded facts from the database and performing reliable forecasting.
Secondly, many other base LLMs are not tested in the experiments. However, we chose some representative models for both open-source and closed-source LLMs, leaving more comprehensive tests for future work.
Thirdly, while the current API covers functions like count, listing, and basic distribution statistics of events and news articles, more functionality can be added to the function pool in the future. For example, to further encourage the agent to analyze based on temporal trends, functions related to time series could be added to the API. We could also allow agents to generate and add useful functions to the API during their reasoning process to support a more diverse toolset.
Fourthly, current experiments are limited by cost and time considerations. More rounds of experiments should be added for each different LLM and various settings. We observe some large variance in the experimental results, which may be caused by the current model temperature of 0.4 and the limited number of 5 rounds per experiment.

\section{Additional related work}
\label{append:related_work}

\subsection{Temporal Event Forecasting Methods}
LLMs have been tested for their temporal understanding through tasks such as temporal event ordering or storyline comprehension~\cite{tempbench, torque, going, situatedqa, tram}. In the context of temporal forecasting, LLMs have also been evaluated on recent text-based temporal forecasting benchmarks~\citep{zou_forecasting_2022, zhang_analyzing_2024}. However, these evaluations typically involve providing LLMs with retrieved context for in-context learning and then directly answering the forecast question. This approach limits the interaction between the LLM and the database and lacks important intermediate thinking and reasoning steps crucial for accurate forecasting. In contrast, \dataset establishes a comprehensive environment that enables an LLM to act as an agent, interacting with the database through multiple rounds of thinking, retrieving diverse information, and providing a complete reasoning path to the final forecast answer. This setup allows for a more thorough evaluation of the LLM's temporal reasoning capabilities and its ability to effectively retrieving and analyzing the available information.

Significant research has also been conducted in the field of structured event temporal forecasting. Various methods have been proposed, including aggregating temporal and relational information among entities~\cite{RENET, REGCN, EvoKG}, retrieving relevant historical events~\cite{cygnet, TimeTraveler, cluster}, and modeling the continuous time evolution of events~\cite{knowevolve, TANGO}. Efforts have also been made to incorporate textual event information into Temporal Knowledge Graphs (TKGs). Glean~\cite{Glean} and CMF~\cite{CMF} integrate textual embeddings into graph edges, while SeCoGD~\cite{SeCoGD} employs textual topic modeling to separate subgraphs. The MidEast-TE dataset and LoGo model~\cite{LoGo} utilize text clustering to construct complex events for forecasting with local and global contexts. However, these methods still perform forecast reasoning solely on graphs using graph-based techniques. Recent studies have explored the use of LLMs for temporal event forecasting by transforming the TKG formulation into text sequences and converting missing object prediction into next token prediction~\cite{PPT}. GPT-NeoX-ICL~\cite{GPT-NeoX-ICL} employs in-context learning of LLMs and constructs prompts as a list of historical events in quadruplet format. GENTKG~\cite{GENTKG} enhances the selection of historical event inputs using a temporal logical rule-based retrieval strategy, while LAMP~\cite{LMAP} applies LLMs to perform abductive reasoning to assist the retrieval process. However, these works only investigate LLMs with in-context learning or simple task-specific fine-tuning. In contrast, \dataset explores forecasting with an LLM agent that supports explicit information gathering and reasoning steps, enabling a hybrid approach that leverages both text and graph data.

\subsection{LLMs for Code Generation}

Early advancements in LLMs have greatly impacted the field of code generation. 
Notable early models such as GPT-3 \citep{brown2020language} and PALM \citep{chowdhery2023palm} have demonstrated the potential of LLMs to assist in code completion and generation. 
Subsequent models have built upon the foundation of pre-trained LLMs, further refining capabilities specific to code generation. This has led to the development of code-specific LLMs such as Codex \citep{chen2021evaluating}, Code T5 \citep{wang2021codet5}, CodeGen \citep{nijkamp2022codegen}, AlphaCode \citep{li2022competition}, StarCoder \citep{li2023starcoder}, WizardCoder \citep{luo2023wizardcoder}, and phi-1 \citep{gunasekar2023textbooks}. 
Moreover, code from open-source platforms such as GitHub has increasingly been incorporated into the pre-training data for recent LLMs \citep{touvron2023llama, achiam2023gpt}. This integration has led to improved performance of LLMs such as Code Llama \citep{roziere2023code}, Code-Qwen \citep{bai2023qwen}, and DeepSeek-Coder \citep{guo2024deepseek} on popular code-related tasks \citep{chen2021evaluating, austin2021program}.

\newpage
\section{Additional Details of the Database}

\subsection{Countries and ISO Codes}
We show the ISO 3166 Alpha-3 country codes and names in Table~\ref{tab:countrycodes} and Table~\ref{tab:countrycodes2}.
\vspace{0.2cm}

\begin{table*}[h]
    \centering
    \caption{List of Countries and their ISO-3166 Alpha-3 Codes (Part 1)}
    \resizebox{0.99\textwidth}{!}{
    \begin{tabular}{ll | ll | ll}
        \toprule
        \textbf{Code} & \textbf{Country} & \textbf{Code} & \textbf{Country} & \textbf{Code} & \textbf{Country} \\
        \midrule
        AND & Andorra & ARE & United Arab Emirates & AFG & Afghanistan \\
        ATG & Antigua and Barbuda & AIA & Anguilla & ALB & Albania \\
        ARM & Armenia & AGO & Angola & ATA & Antarctica \\
        ARG & Argentina & ASM & American Samoa & AUT & Austria \\
        AUS & Australia & ABW & Aruba & ALA & Åland \\
        AZE & Azerbaijan & BIH & Bosnia and Herzegovina & BRB & Barbados \\
        BGD & Bangladesh & BEL & Belgium & BFA & Burkina Faso \\
        BGR & Bulgaria & BHR & Bahrain & BDI & Burundi \\
        BEN & Benin & BLM & Saint Barthélemy & BMU & Bermuda \\
        BRN & Brunei & BOL & Bolivia & BES & Bonaire, Sint Eustatius, and Saba \\
        BRA & Brazil & BHS & Bahamas & BTN & Bhutan \\
        BVT & Bouvet Island & BWA & Botswana & BLR & Belarus \\
        BLZ & Belize & CAN & Canada & CCK & Cocos (Keeling) Islands \\
        COD & DR Congo & CAF & Central African Republic & COG & Congo Republic \\
        CHE & Switzerland & CIV & Ivory Coast & COK & Cook Islands \\
        CHL & Chile & CMR & Cameroon & CHN & China \\
        COL & Colombia & CRI & Costa Rica & CUB & Cuba \\
        CPV & Cabo Verde & CUW & Curaçao & CXR & Christmas Island \\
        CYP & Cyprus & CZE & Czechia & DEU & Germany \\
        DJI & Djibouti & DNK & Denmark & DMA & Dominica \\
        DOM & Dominican Republic & DZA & Algeria & ECU & Ecuador \\
        EST & Estonia & EGY & Egypt & ESH & Western Sahara \\
        ERI & Eritrea & ESP & Spain & ETH & Ethiopia \\
        FIN & Finland & FJI & Fiji & FLK & Falkland Islands \\
        FSM & Micronesia & FRO & Faroe Islands & FRA & France \\
        GAB & Gabon & GBR & United Kingdom & GRD & Grenada \\
        GEO & Georgia & GUF & French Guiana & GGY & Guernsey \\
        GHA & Ghana & GIB & Gibraltar & GRL & Greenland \\
        GMB & The Gambia & GIN & Guinea & GLP & Guadeloupe \\
        GNQ & Equatorial Guinea & GRC & Greece & SGS & South Georgia and South Sandwich Islands \\
        GTM & Guatemala & GUM & Guam & GNB & Guinea-Bissau \\
        GUY & Guyana & HKG & Hong Kong & HMD & Heard and McDonald Islands \\
        HND & Honduras & HRV & Croatia & HTI & Haiti \\
        HUN & Hungary & IDN & Indonesia & IRL & Ireland \\
        ISR & Israel & IMN & Isle of Man & IND & India \\
        IOT & British Indian Ocean Territory & IRQ & Iraq & IRN & Iran \\
        ISL & Iceland & ITA & Italy & JEY & Jersey \\
        JAM & Jamaica & JOR & Jordan & JPN & Japan \\
        KEN & Kenya & KGZ & Kyrgyzstan & KHM & Cambodia \\
        KIR & Kiribati & COM & Comoros & KNA & St Kitts and Nevis \\
        PRK & North Korea & KOR & South Korea & KWT & Kuwait \\
        CYM & Cayman Islands & KAZ & Kazakhstan & LAO & Laos \\
        LBN & Lebanon & LCA & Saint Lucia & LIE & Liechtenstein \\
        LKA & Sri Lanka & LBR & Liberia & LSO & Lesotho \\
        LTU & Lithuania & LUX & Luxembourg & LVA & Latvia \\
        LBY & Libya & MAR & Morocco & MCO & Monaco \\
        MDA & Moldova & MNE & Montenegro & MAF & Saint Martin \\
        MDG & Madagascar & MHL & Marshall Islands & MKD & North Macedonia \\
        MLI & Mali & MMR & Myanmar & MNG & Mongolia \\
        MAC & Macao & MNP & Northern Mariana Islands & MTQ & Martinique \\
        MRT & Mauritania & MSR & Montserrat & MLT & Malta \\
        MUS & Mauritius & MDV & Maldives & MWI & Malawi \\
        MEX & Mexico & MYS & Malaysia & MOZ & Mozambique \\
        NAM & Namibia & NCL & New Caledonia & NER & Niger \\
        NFK & Norfolk Island & NGA & Nigeria & NIC & Nicaragua \\
        NLD & The Netherlands & NOR & Norway & NPL & Nepal \\
        NRU & Nauru & NIU & Niue & NZL & New Zealand \\
        OMN & Oman & PAN & Panama & PER & Peru \\
        PYF & French Polynesia & PNG & Papua New Guinea & PHL & Philippines \\
        PAK & Pakistan & POL & Poland & SPM & Saint Pierre and Miquelon \\
        PCN & Pitcairn Islands & PRI & Puerto Rico & PSE & Palestine \\
        PRT & Portugal & PLW & Palau & PRY & Paraguay \\
        QAT & Qatar & REU & Réunion & ROU & Romania \\
        SRB & Serbia & RUS & Russia & RWA & Rwanda \\
        SAU & Saudi Arabia & SLB & Solomon Islands & SYC & Seychelles \\
        SDN & Sudan & SWE & Sweden & SGP & Singapore \\
        SHN & Saint Helena & SVN & Slovenia & SJM & Svalbard and Jan Mayen \\
        SVK & Slovakia & SLE & Sierra Leone & SMR & San Marino \\
        
        \bottomrule
    \end{tabular}
    }

    \label{tab:countrycodes}
\end{table*}

\begin{table*}[h]
    \centering
    \caption{List of Countries and their ISO-3166 Alpha-3 Codes (Part 2)}
    \resizebox{0.99\textwidth}{!}{
    \begin{tabular}{ll | ll | ll}
        \toprule
        \textbf{Code} & \textbf{Country} & \textbf{Code} & \textbf{Country} & \textbf{Code} & \textbf{Country} \\
        \midrule
        
        SEN & Senegal & SOM & Somalia & SUR & Suriname \\ 
        SSD & South Sudan & STP & São Tomé and Príncipe & SLV & El Salvador \\
        SXM & Sint Maarten & SYR & Syria & SWZ & Eswatini \\
        TCA & Turks and Caicos Islands & TCD & Chad & ATF & French Southern Territories \\
        TGO & Togo & THA & Thailand & TJK & Tajikistan \\
        TKL & Tokelau & TLS & Timor-Leste & TKM & Turkmenistan \\
        TUN & Tunisia & TON & Tonga & TUR & Türkiye \\
        TTO & Trinidad and Tobago & TUV & Tuvalu & TWN & Taiwan \\
        TZA & Tanzania & UKR & Ukraine & UGA & Uganda \\
        UMI & U.S. Outlying Islands & USA & United States & URY & Uruguay \\
        UZB & Uzbekistan & VAT & Vatican City & VCT & St Vincent and Grenadines \\
        VEN & Venezuela & VGB & British Virgin Islands & VIR & U.S. Virgin Islands \\
        VNM & Vietnam & VUT & Vanuatu & WLF & Wallis and Futuna \\
        WSM & Samoa & XKX & Kosovo & YEM & Yemen \\
        MYT & Mayotte & ZAF & South Africa & ZMB & Zambia \\
        ZWE & Zimbabwe \\
        \bottomrule
    \end{tabular}
    }
    \label{tab:countrycodes2}
\end{table*}
\vspace{0.2cm}

\subsection{Relations and CAMEO Codes}
We show the CAMEO relation codes and names, where the first-level relations are in two digits, and the second-level relations are in three digits with the first two digits be the same as its parent relation:
\vspace{0.2cm}
\begin{itemize}
    \item 01: Make public statement
    \begin{itemize}
        \item 010: Make statement, not specified
        \item 011: Decline comment
        \item 012: Make pessimistic comment
        \item 013: Make optimistic comment
        \item 014: Consider policy option
        \item 015: Acknowledge or claim responsibility
        \item 016: Reject accusation or deny responsibility
        \item 017: Engage in symbolic act
        \item 018: Make empathetic comment
        \item 019: Express accord
    \end{itemize}
    \item 02: Appeal
    \begin{itemize}
        \item  020: Make an appeal or request, not specified
        \item 021: Appeal for material cooperation
        \item 022: Appeal for diplomatic cooperation
        \item 023: Appeal for material aid
        \item 024: Appeal for political reform
        \item 025: Appeal to yield
        \item 026: Appeal to others to meet or negotiate
        \item 027: Appeal to others to settle dispute
        \item 028: Appeal to others to engage in or accept mediation
    \end{itemize}
    \item 03: Express intent to cooperate
    \begin{itemize}
        \item 030: Express intent to cooperate, not specified
        \item 031: Express intent to engage in material cooperation
        \item 032: Express intent to engage in diplomatic cooperation
        \item 033:  Express intent to provide material aid
        \item 034: Express intent to institute political reform
        \item 035: Express intent to yield
        \item 036: Express intent to meet or negotiate
        \item 037: Express intent to settle dispute
        \item 038: Express intent to accept mediation
        \item 039: Express intent to mediate
    \end{itemize}
    \item 04: Consult
    \begin{itemize}
        \item 040: Consult, not specified
        \item 041: Discuss by telephone
        \item 042: Make a visit
        \item 043: Host a visit
        \item 044: Meet at a third location
        \item 045: Engage in mediation
        \item 046: Engage in negotiation
    \end{itemize}
    \item 05: Engage in diplomatic cooperation
    \begin{itemize}
        \item 050:Engage in diplomatic cooperation, not specified
        \item 051: Praise or endorse
        \item 052: Defend verbally
        \item 053: Rally support on behalf of
        \item 054: Grant diplomatic recognition
        \item 055: Apologize
        \item 056: Forgive
        \item 057: Sign formal agreement
    \end{itemize}
    \item 06: Engage in material cooperation
    \begin{itemize}
        \item 060: Engage in material cooperation, not specified
        \item 061: Cooperate economically
        \item 062: Cooperate militarily
        \item 063: Engage in judicial cooperation
        \item 064: Share intelligence or information
    \end{itemize}
    \item 07: Provide aid
    \begin{itemize}
        \item 070: Provide aid, not specified
        \item 071: Provide economic aid
        \item 072: Provide military aid
        \item 073: Provide humanitarian aid
        \item 074: Provide military protection or peacekeeping
        \item 075: Grant asylum
    \end{itemize}
   \item 08: Yield
   \begin{itemize}
       \item 080: Yield, not specified
       \item 081: Ease administrative sanctions
       \item 082: Ease political dissent
       \item 083: Accede to requests or demands for political reform
       \item 084: Return or release
       \item 085: Ease economic sanction or boycott or embargo
       \item 086: Allow international involvement
       \item 087: De-escalate military engagement
   \end{itemize}
   \item 09: Investigate
       \begin{itemize}
           \item 090: Investigate, not specified
           \item 091: Investigate crime or corruption
           \item 092: Investigate human rights abuses
           \item 093: Investigate military action
           \item 094: Investigate war crimes
       \end{itemize}
  
    \item 10: Demand
     \begin{itemize}
         \item 100: Demand, not specified
         \item 101: Demand material cooperation
         \item 102: Demand for diplomatic cooperation
         \item 103: Demand material aid
         \item 104: Demand political reform
         \item 105: Demand that target yield
         \item 106: Demand meeting or negotiation
         \item 107: Demand settling of dispute
         \item 108: Demand mediation
     \end{itemize}
   
   \item 11: Disapprove
       \begin{itemize}
           \item 110: Disapprove, not specified
           \item 111: Criticize or denounce
           \item 112: Accuse
           \item 113: Rally opposition against
           \item 114: Complain officially
           \item 115: Bring lawsuit against
           \item 116: Find guilty or liable (legally)
       \end{itemize}
    \item 12: Reject
       \begin{itemize}
           \item 120: All rejections and refusals
           \item 121: Reject material cooperation
           \item 122: Reject request or demand for material aid
           \item 123: Reject request or demand for political reform
           \item 124: Refuse to yield
           \item 125: Reject proposal to meet or discuss or negotiate
           \item 126: Reject mediation
           \item 127: Reject plan or agreement to settle dispute
           \item 128: Defy norms or law
           \item 129: Veto
       \end{itemize}
   
   \item 13: Threaten
   \begin{itemize}
       \item 130: Threaten, not specified
       \item 131: Threaten non-force
       \item 132: Threaten with administrative sanctions
       \item 133: Threaten political dissent
       \item 134: Threaten to halt negotiations
       \item 135: Threaten to halt mediation
       \item 136: Threaten to halt international involvement
       \item 137: Threaten with repression
       \item 138: Threaten with military force
       \item 139: Give ultimatum
   \end{itemize}
    \item 14: Protest
    \begin{itemize}
        \item 140: Engage in political dissent, not specified
        \item 141: Demonstrate or rally
        \item 142: Conduct hunger strike
        \item 143: Conduct strike or boycott
        \item 144: Obstruct passage or block
        \item 145: Protest violently or riot
    \end{itemize}

   \item 15: Exhibit military posture
    \begin{itemize}
        \item 150: Exhibit military or police power, not specified
        \item 151: Increase police alert status
        \item 152: Increase military alert status
        \item 153: Mobilize or increase police power
        \item 154: Mobilize or increase armed forces
        \item 155: Mobilize or increase cyber-forces
    \end{itemize}

   \item 16: Reduce relations
   
    \begin{itemize}
        \item 160: Reduce relations, not specified
        \item 161: Reduce or break diplomatic relations
        \item 162: Reduce or stop material aid
        \item 163: Impose embargo or boycott or sanctions
        \item 164: Halt negotiations
        \item 165: Halt mediation
        \item 166: Expel or withdraw
    \end{itemize}

   \item 17: Coerce
   \begin{itemize}
       \item 170: Coerce
       \item 171: Seize or damage property
       \item 172: Impose administrative sanctions
       \item 173: Arrest or detain
       \item 174: Expel or deport individuals
       \item 175: Use repression
       \item 176: Attack cybernetically
   \end{itemize}{}
    \item 18: Assault
    \begin{itemize}
        \item 180: Use unconventional violence, not specified
        \item 181: Abduct or hijack or take hostage
        \item 182: Physically assault
        \item 183: Conduct suicide or car or other non-military bombing
        \item 184: Use as human shield
        \item 185: Attempt to assassinate
        \item 186: Assassinate
    \end{itemize}
  \item 19: Fight
  \begin{itemize}
      \item 190: Use conventional military force, not specified
      \item 191: Impose blockade or restrict movement
      \item 192: Occupy territory
      \item 193: Fight with small arms and light weapons
      \item 194: Fight with artillery and tanks
      \item 195: Employ aerial weapons
      \item 196: Violate ceasefire
  \end{itemize}
   
    \item 20: Engage in unconventional mass violence
    \begin{itemize}
        \item 200: Use massive unconventional force, not specified
        \item 201: Engage in mass expulsion
        \item 202: Engage in mass killings
        \item 203: Engage in ethnic cleansing
        \item 204: Use weapons of mass destruction
    \end{itemize}
\end{itemize}

\clearpage

\section{Details of Dataset Construction} \label{sec:construction}

We construct the database based on the GDELT\footnote{\url{https://www.gdeltproject.org/}} project, which systematically captures global news media, extracts information about countries and events mentioned, and updates this data every 15 minutes. The data stored in GDLET contains attributes about the event date, actor, action (relation), geography, and source news.  The original GDELT dataset, while extensive, contains significant noise due to its rapid update frequency and the diverse nature of its sources. We thus propose the folowing data-cleaning pre-processing steps.

\mypar{Events.} We focus on the GDELT data from Jan. 1, 2023, to Nov. 30, 2023. To mitigate potential information leakage, we ensured that the event date coincided exactly with its source news publish date. We standardized the \texttt{CountryCode} to the ISO-3166 and excluded any missing or outdated entries. As our analysis was restricted to international events, we then filte out all domestic events where the subject was identical to the object at the country level. The \texttt{EventCode} was standardized to the second level of the CAMEO ontology, providing a specific yet consistent level of detail. We opted not to include third-level relation information for two primary reasons: the absence of sublevel relations for all second-level categories, which introduces inconsistencies, and the higher likelihood of extraction errors in the finely detailed third-level data in the GDELT dataset.

\mypar{News articles.} To enhance the reliability of the information, we utilized the news sources of the events to assess credibility. Specifically, we calculated the daily mentions of each event—defined as the number of distinct news sources discussing the event—based on the 15-minute updates from GDELT. We only retained events with at least 50 daily mentions to diminish the influence of less reliable sources, such as personal blogs or advertisements, and to decrease the incidence of erroneously extracted events. We then download the news documents from the corresponding news URLs of each event. In the current database, we only keep textual information from the news, including new titles and news content. Following the OBELICS~\citep{DBLP:conf/nips/LaurenconSTBSLW23} protocol for web text cleaning, we applied rigorous filters at both the paragraph and document levels, targeting criteria such as a low number of words, high repetition ratios of characters and words, and an excessive presence of special characters and flagged words. These measures were crucial in significantly reducing the noise from web content.

\section{Term of Use for GDELT Dataset}\label{sec:term}
Based on \url{https://www.gdeltproject.org/about.html#termsofuse}, GDELT dataset ``\textit{is an open platform for research and analysis of global society and thus all datasets released by the GDELT Project are available for unlimited and unrestricted use for any academic, commercial, or governmental use of any kind without fee.}'', as long as ``\textit{any use or redistribution of the data must include a citation to the GDELT Project and a link to this website (\url{https://www.gdeltproject.org/}).}'', which we've cited in abstract.

\section{Broader Impact}\label{sec:impact}
\vspace{-0.2cm}
\paragraph{Positive impact.}
This paper aims to construct high-level and high-quality instruction to improve the scientific reasoning capability of LLMs, which helps LLMs to better give the answers to questions at the college level. 
Collecting diverse instructions, annotating self-reflective instructions, and filtering out low-quality instructions provide researchers insights to prepare training datasets. 

\paragraph{Negative impact.} A drawback of this work is that the scale of the training dataset and model is relatively small, and we can address this by bootstrapping a more large training dataset. We believe that the benefits of data generation manner outweigh the downside.

\section{Additional Details about Agent Setup}

\textbf{Final answer extraction.} Our agent has two stopping criteria: 1) The agent makes the final answer. 2) The reasoning process fails to reach an answer (Consecutive Invalid Actions, Consecutive Repetitive Actions, or Max Iterations Exceeded). Afterwards, we always perform an answer extraction step, which is performed by \texttt{gpt-3.5-turbo}. The prompt for answer extraction is shown in Appendix \ref{append:ansprompt}.

\clearpage

\section{Additional Experimental Results and Analysis}

\subsection{Analysis of the Final Status of Different LLM Agents}
\label{append:final_status}
\begin{table*}[!ht]
    \centering
    \caption{Average number of ReAct iterations and the number of test cases ending in different final status.}
    \resizebox{\textwidth}{!}{%
\begin{tblr}{colspec = {cccccccc},
row{1-2} = {bg=gray!25},
row{5-6,9-10} = {bg=gray!10}}
    \toprule
        \SetCell[r=2]{c}{Model} & \SetCell[r=2]{c}{Training Data \\ Cutoff Date} & \SetCell[r=2]{c}{Action Type} &\SetCell[r=2]{c}{Avg. Iterations} & \SetCell[c=4]{c}{Final Status} \\
        \cmidrule[lr]{5-8}
        & &  &  & \SetCell{c,m}{Final Answer} & \SetCell{c,m}{Consecutive \\ Invalid Actions} & \SetCell{c,m}{Consecutive \\ Repetitive Actions} & \SetCell{c,m}{Max Iterations \\ Exceeded}\\
        \midrule
         \SetCell[r=2]{c}{\texttt{Mistral-7B}}  & \SetCell[r=2]{c}{---} & Single Function & 4.23&	79	&19	&2&	0 \\
        & & Code Block& 2.84&	54	&45&	1	&0\\
        \SetCell[r=2]{c}{\texttt{gpt-3.5-turbo}}& \SetCell[r=2]{c}{2021-09} & Single Function & 2.54&	98&	1	&1	&0\\
        &&Code Block&3.43&	84	&15	&1	&0\\
        \SetCell[r=2]{c}{\texttt{gpt-4-turbo}}& \SetCell[r=2]{c}{2023-12} & Single Function & 8.76&	\textbf{99}	&0	&1&	0\\
        &&Code Block&4.68&	93	&5	&2&	0\\
        \SetCell[r=2]{c}{\texttt{gpt-4o}}& \SetCell[r=2]{c}{2023-10} & Single Function & 8.79	&98	&0&	0&	2\\
        &&Code Block&5.58	&95&	4&	1	&0\\
    \bottomrule
    \end{tblr}%
    }
    \label{tab:main_err}
\end{table*}

Following the experiment on different base LLMs in Sec. \ref{sec:exp_basemodel}, we further analyze the final status of the agent when it triggers the termination conditions in the ReAct process. The agent-environment interactions were terminated based on four different statuses:
\begin{itemize}[nosep]
    \item Final Answer: This status is triggered when the model explicitly indicates completion by declaring a "final answer," indicating that it has generated its final forecast.  
    \item Consecutive Invalid Actions: Interaction is halted if the model consecutively executes non-executable actions more than three times.
    \item Consecutive Repetitive Actions: A similar threshold of three consecutive repetitive actions prompts termination under this status.
    \item Maximum Iterations Exceeded: Termination occurs if the agent engages in more than 20 rounds of iterations without making a forecast answer.
\end{itemize}

We show the final status 
\textbf{1) The GPT series demonstrates enhanced capability in concluding interactions with a final answer:} The GPT series frequently concludes interactions with a final answer, indicating its superior ability to comprehend instructions, utilize tools correctly, and perform reasoning to generate the final forecast. In contrast, the smaller open-source model, Mistral-7b, has a round one-quarter of cases and one-half of the cases generating consecutive invalid answers when using single function and code block, respectively, suggesting limited abilities in adhering to data types, functions, Python syntax, and following detailed instructions.

\textbf{2) Generating code blocks proves more challenging than generating single functions as action:} Across all models, there is a higher occurrence of errors when generating code blocks compared to single functions. This pattern highlights the greater complexity and increased likelihood of errors associated with composing multiple lines of code over simple function calls.

\textbf{3) Advanced models engage in more extensive information gathering, resulting in better forecasting performance: }Advanced models, such as GPT-4 Turbo and GPT-4o, engage in significantly more information gathering, with average iterations much more or even double than GPT-3.5 Turbo and Mistral-7B. This intensive data collection from the database may contribute to their enhanced performance, as previously illustrated in Table~\ref{tab:main_res}. These models exhibit better planning capability, effectively identifying and searching for related historical information.

\begin{figure}
    \centering
    \includegraphics[width=0.6\textwidth]{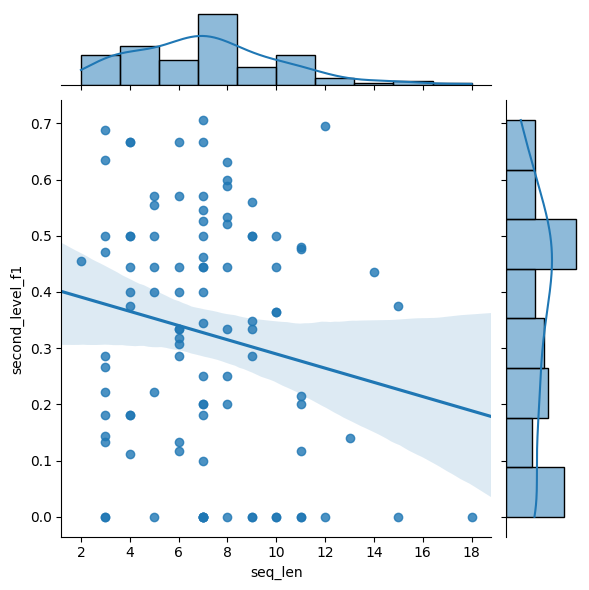}
    \caption{Correlation of F1 Accuracy to Action Sequence Length}
    \label{fig:seq}
\end{figure}

\subsection{Analysis on Function Types}
\label{append:function_type}

\clearpage

\section{Prompts}
\label{append:prompts}

\subsection{System Prompts}
The system prompt provides the necessary background information, task description, and guidelines for the LLM agent. In our case, we introduce the forecasting task, basic information of the database, the defined API, and the planning strategies in the system prompt.

\subsubsection{System Prompt for ReAct Agent with Action Type as ``\hlc[uclablue!20]{Single Function}''}
% \chenchen{correct ```, fix indentation, table length exceed page length}
\begin{tcolorbox}[title=System prompt for ReAct agent with action type as single function (part 1), colframe=custom3, colback=white]
You are an expert in forecasting future events based on historical data. The database contains news articles from January 1, 2023 to the current date \{current\_date\_nlp\} and the events extracted from these articles. The events are in the form of (date, subject country, relation, object country), where the countries are represented by ISO 3166-1 alpha-3 codes and the relations are represented by the CAMEO codes defined in the `Conflict and Mediation Event Observations' ontology. The relations are hierarchical: first-level relations are general parent relations represented by two-digit CAMEO codes, while second-level relations are more specific child relations represented by three-digit CAMEO codes. Child relations have the same first two digits as their parent relations. For example, `01' is a first-level relation, and `010' and `011' are some of its second-level relations. The relations in the database are represented in the second-level form.\\

Your task is to forecast the future relations between two entities in a given query. You have access to a defined Python API that allows you to query the database for historical events and statistics, and to get precise information about the ISO country codes and CAMEO relation codes.\\

The defined API is described as follows: 

{\textasciigrave}{\textasciigrave}{\textasciigrave}python\\
\{api\_description\}\\
{\textasciigrave}{\textasciigrave}{\textasciigrave}\\

You will use an iterative approach, interleaving 'Thought', 'Action', and 'Observation' steps to collect information and perform the forecast. You may perform up to \{max\_iterations\} iterations. The steps are as follows: \\

- `Thought': Analyze the current information and reason about the current situation, and predicts which API you want to use (try to use different APIs to collect diverse information) or make a decision that you want to make a final answer. 

- `Action': Use the API to gather more information or provide the final forecast.

\hspace*{6mm} - If using the API: \hlc[uclablue!20]{the action must be only one single line of exactly one function }
\hspace*{6mm} \hlc[uclablue!20]{call from the API with appropriate inputs, without additional code, explanations, or} 
\hspace*{6mm} \hlc[uclablue!20]{natural language descriptions.}
    
\hspace*{6mm} - If making the final forecast: the action must start immediately with `Final Answer:', 
\hspace*{6mm} and follow with the results in the expected JSON format. 
    
- `Observation': Return the output of the called function. \\

To make a reasonable forecast, you should collect both news and relational evidence to support your prediction. When you are fully confident that you accumulate enough information to make the final forecast, you should start the `Thought' with your reasoning using the news and structural information to make the prediction, and then start the `Action' step with `Final Answer:' followed by the answer in the expected JSON format. The answer should be a JSON dictionary where the keys are the forecasted two-digit first-level CAMEO codes and the values are lists of forecasted three-digit second-level CAMEO codes that are child relations of the key. For example, `Action: Final Answer: {{"01": ["010", "011", "012"], "02": ["020", "023"]}}'.

\end{tcolorbox}

\begin{tcolorbox}[title=System prompt for ReAct agent with action type as single function (part 2), colframe=custom3, colback=white]
The final answer will be evaluated based on the precision and recall of the forecasted first-level and second-level relations, so only include confident first-level and second-level CAMEO codes in your final forecast.\\

Try to use different APIs to collect diverse information (including multi-hop relations), such as the precise meaning of CAMEO codes, insights from news content, relational data, and statistical analyses to support your forecasts. Consider not only the frequency of the relations but also the temporal aspects of the data when making your forecast. 
\end{tcolorbox}

\subsubsection{System Prompt for ReAct Agent with Action Type as ``\hlc[blue!20]{Code Block}''}
% \chenchen{correct ```, fix indentation, table length exceed page length}
\begin{tcolorbox}[title=System prompt for ReAct agent with action type as code block (part 1), colframe=custom3, colback=white]
You are an expert in forecasting future events based on historical data. The database contains news articles from January 1, 2023 to the current date \{current\_date\_nlp\} and the events extracted from these articles. The events are in the form of (date, subject country, relation, object country), where the countries are represented by ISO 3166-1 alpha-3 codes and the relations are represented by the CAMEO codes defined in the `Conflict and Mediation Event Observations' ontology. The relations are hierarchical: first-level relations are general parent relations represented by two-digit CAMEO codes, while second-level relations are more specific child relations represented by three-digit CAMEO codes. Child relations have the same first two digits as their parent relations. For example, `01' is a first-level relation, and `010' and `011' are some of its second-level relations. The relations in the database are represented in the second-level form.\\

Your task is to forecast the future relations between two entities in a given query. You have access to a defined Python API that allows you to query the database for historical events and statistics, and to get precise information about the ISO country codes and CAMEO relation codes. You are also authorized to utilize additional safe, well-established Python libraries such as numpy, pandas, scikit-learn, and NetworkX to enhance your data analysis and forecasting accuracy.\\

The defined API is described as follows:\\
{\textasciigrave}{\textasciigrave}{\textasciigrave} python\\
\{api\_description\}\\
{\textasciigrave}{\textasciigrave}{\textasciigrave}\\

You will use an iterative approach, interleaving `Thought', `Action', and `Observation' steps to collect information and perform the forecast. You may perform up to {max\_iterations} iterations. The steps are as follows:\\

- `Thought': Analyze the current information and reason about the current situation, and predicts which API you want to use (try to use different APIs to collect diverse information) or make a decision that you want to make a final answer.\\
- `Action': Use the API to gather more information or provide the final forecast.\\
\hspace*{6mm} - If gathering more data:  \hlc[blue!20]{the action must be an executable Python code snippet that} 
\hspace*{6mm}  \hlc[blue!20]{starts with `{\textasciigrave}{\textasciigrave}{\textasciigrave}python' and ends with `{\textasciigrave}{\textasciigrave}{\textasciigrave}'. It can contain multiple lines of codes and} 
\hspace*{6mm}  \hlc[blue!20]{function calls using the defined API or Python libraries. You must use print() to output} 
\hspace*{6mm}  \hlc[blue!20]{the results, and only the printed output will be returned in the observation step.}\\
\hspace*{6mm} - If making the final forecast: the action must start immediately with `Final Answer:', 
\hspace*{6mm} and follow with the answer in the expected JSON format. This should not be enclosed 
\hspace*{6mm} within triple backticks.\\
- `Observation': Return the printed output of the executed code snippet.\\
\end{tcolorbox}

\begin{tcolorbox}[title=System prompt for ReAct agent with action type as code block (part 2), colframe=custom3, colback=white]
To make a reasonable forecast, you should collect both news and relational evidence to support your prediction. When you are fully confident that you accumulate enough information to make the final forecast, you should start the `Thought' with your reasoning using the news and structural information to make the prediction, and then start the `Action' step with `Final Answer:' followed by the answer in the expected JSON format. The answer should be a JSON dictionary where the keys are the forecasted two-digit first-level CAMEO codes and the values are lists of forecasted three-digit second-level CAMEO codes that are child relations of the key. For example, `Action: Final Answer: \{"01": ["010", "011", "012"], "02": ["020", "023"]\}'.\\

The final answer will be evaluated based on the precision and recall of the forecasted first-level and second-level relations, so only include confident first-level and second-level CAMEO codes in your final forecast.\\

Try to use different APIs and Python libraries to collect diverse information (including multi-hop relations), such as the precise meaning of CAMEO codes, insights from news content, relational data, and statistical analyses to support your forecasts. Consider not only the frequency of the relations but also the temporal aspects of the data when making your forecast.
\end{tcolorbox}

% \clearpage
\vspace{1cm}
\subsubsection{System Prompt for Direct Agent}
\begin{tcolorbox}[title=System prompt for direct agent, colframe=custom3, colback=white]
You are an expert in forecasting future events based on historical data. The events are in the form of (date, subject country, relation, object country), where the countries are represented by ISO 3166-1 alpha-3 codes and the relations are represented by the CAMEO codes defined in the `Conflict and Mediation Event Observations' ontology. The relations are hierarchical: first-level relations are general parent relations represented by two-digit CAMEO codes, while second-level relations are more specific child relations represented by three-digit CAMEO codes. Child relations have the same first two digits as their parent relations. For example, `01' is a first-level relation, and `010' and `011' are some of its second-level relations. The relations in the database are represented in the second-level form.\\

Your task is to forecast the future relations between two entities in a given query. The answer should be a JSON dictionary where the keys are the forecasted two-digit first-level CAMEO codes and the values are lists of forecasted three-digit second-level CAMEO codes that are child relations of the key. For example, `Final Answer: \{\{``01'': [``010'', ``011'', ``012''], ``02'': [``020'', ``023'']\}\}'. \\

The final answer will be evaluated based on the precision and recall of the forecasted first-level and second-level relations, so only include confident first-level and second-level CAMEO codes in your final forecast. 
\end{tcolorbox}

\clearpage

\subsubsection{System Prompt for CoT Agent}
\begin{tcolorbox}[title=System prompt for CoT, colframe=custom3, colback=white]
You are an expert in forecasting future events based on historical data. The events are in the form of (date, subject country, relation, object country), where the countries are represented by ISO 3166-1 alpha-3 codes and the relations are represented by the CAMEO codes defined in the `Conflict and Mediation Event Observations' ontology. The relations are hierarchical: first-level relations are general parent relations represented by two-digit CAMEO codes, while second-level relations are more specific child relations represented by three-digit CAMEO codes. Child relations have the same first two digits as their parent relations. For example, `01' is a first-level relation, and `010' and `011' are some of its second-level relations. The relations in the database are represented in the second-level form. \\

Your task is to forecast the future relations between two entities in a given query.\textbf{ To make a reasonable forecast, you should first think and reason based on your background knowledge. When you are confident that you have conducted enough analysis to make the final answer}, you should start answering by 'Therefore, the final answer is:` followed by the answer in the expected JSON format. The JSON format should be a JSON dictionary where the keys are the forecasted two-digit first-level CAMEO codes and the values are lists of forecasted three-digit second-level CAMEO codes that are child relations of the key. For example, `\{\{``01'': [``010'', ``011'', ``012''], ``02'': [``020'', ``023'']\}\}'. \\

The final answer will be evaluated based on the precision and recall of the forecasted first-level and second-level relations, so only include confident first-level and second-level CAMEO codes in your final forecast. 

\end{tcolorbox}

\vspace{1cm}
\subsection{Query Prompt}
\begin{tcolorbox}[title=Query prompt,  colframe=custom4, colback=white]
Please forecast the relations that \{actor1\_name\} will take towards \{actor2\_name\} on \{future\_date\_nlp\} based on historical information up to \{current\_date\_nlp\}. I.e. forecast the relation CAMEO codes in query event Event(date=\{future\_date\}, head\_entity=ISOCode(\{actor1\_code\}), relation=CAMEOCode(?), tail\_entity=ISOCode(\{actor2\_code\})).
\end{tcolorbox}
\newpage

\subsection{Answer Extraction Prompt}
\label{append:ansprompt}
\begin{tcolorbox}[title=Answer extraction prompt,  colframe=custom2, colback=white]
Please help me extract final answer for forecasting the future relations between two entities in a given query: forecast the relations that \{actor1\_name\} will take towards \{actor2\_name\} on \{future\_date\_nlp\} based on historical information up to {current\_date\_nlp}. I.e. forecast the relation CAMEO codes in query event Event(date=\{future\_date\}, head\_entity=ISOCode(\{actor1\_code\}), relation=CAMEOCode(?), tail\_entity=ISOCode(\{actor2\_code\})). \\

I have used interleaving `Thought', `Action', and `Observation' steps to collect information from the database and perform the forecast. The database contains news articles from January 1, 2023 to the current date {current\_date\_nlp} and the events extracted from these articles. The events are in the form of (date, subject country, relation, object country), where the countries are represented by ISO 3166-1 alpha-3 codes and the relations are represented by the CAMEO codes defined in the `Conflict and Mediation Event Observations' ontology. The relations are hierarchical: first-level relations are general parent relations represented by two-digit CAMEO codes, while second-level relations are more specific child relations represented by three-digit CAMEO codes. Child relations have the same first two digits as their parent relations. For example, `01' is a first-level relation, and `010' and `011' are some of its second-level relations. The relations in the database are represented in the second-level form.\\

The final forecast answer need to forecast both first-level and second-level CAMEO codes, and will be evaluated based on the precision and recall of both levels of relations. The final answer content should be a JSON dictionary where the keys are the forecasted two-digit first-level CAMEO codes and the values are lists of forecasted three-digit second-level CAMEO codes that are child relations of the key. For example, \{\{``01'': [``010'', ``011'', ``012''], ``02'': [``020'', ``023'']\}\}.\\

The latest information and forecast I have collected is as follows: 

\{info\}\\

If final forecast answer has been made in the collected information indicated by "Final Answer:", you must only reformat the final forecast answer in the expected JSON dictionary format inside XML tags. For example: <answer>\{\{``01'': [``010'', ``011'', ``012''], ``02'': [``020'', ``023'']\}\}</answer>.\\

Otherwise, if no final forecast is made, you must reason based on the information you have collected and generate a confident final forecast answer to the query, and then reformat your answer in the expected JSON dictionary format inside XML tags.

\end{tcolorbox}

\clearpage

\newpage
\section{Additional Forecasting Examples of LLM Agent}
\label{append:examples}

In this section, we show examples of how LLM agents perform reasoning for the forecasting query. The query-specific values are highlighted in yellow in the system prompt and query prompt, which follows the prompt templates shown in Appendix~\ref{append:prompts}. For simplicity, we use \texttt{\{api\_description\}} as a placeholder for the API specification in the prompt, which is replaced by the actual contents shown in Appendix~\ref{append:api} in experiments.

\subsection{GPT-4o-based Agent with ReAct and ``\hlc[blue!20]{Code Block}'' Action}

\subsubsection{Query Details}
\begin{itemize}
    \item \textbf{Query Quadruplet}: (2023-11-03, AUS, ?, CHN)
    \item \textbf{Temporal Distance}: 1; therefore, the current date is 2023-11-02
    \item \textbf{Agent Max Steps}: 20
\end{itemize}

\subsubsection{Query Prompt}
Please forecast the relations that \hlc[yellow!30]{Australia} will take towards \hlc[yellow!30]{China} on \hlc[yellow!30]{November 03, 2023} based on historical information up to \hlc[yellow!30]{November 02, 2023}. I.e. forecast the relation CAMEO codes in query event \hlc[yellow!30]{Event(date=2023-11-03, head\_entity=ISOCode(AUS), relation=CAMEOCode(?), tail\_entity=ISOCode(CHN))}.

\subsubsection{System Prompt}
You are an expert in forecasting future events based on historical data. The database contains news articles from January 1, 2023 to the current date \hlc[yellow!30]{November 02, 2023} and the events extracted from these articles. The events are in the form of (date, subject country, relation, object country), where the countries are represented by ISO 3166-1 alpha-3 codes and the relations are represented by the CAMEO codes defined in the "Conflict and Mediation Event Observations" ontology. The relations are hierarchical: first-level relations are general parent relations represented by two-digit CAMEO codes, while second-level relations are more specific child relations represented by three-digit CAMEO codes. Child relations have the same first two digits as their parent relations. For example, "01" is a first-level relation, and "010" and "011" are some of its second-level relations. The relations in the database are represented in the second-level form.
\\\\
Your task is to forecast the future relations between two entities in a given query. You have access to a defined Python API that allows you to query the database for historical events and statistics, and to get precise information about the ISO country codes and CAMEO relation codes. You are also authorized to utilize additional safe, well-established Python libraries such as numpy, pandas, scikit-learn, and NetworkX to enhance your data analysis and forecasting accuracy.
\\\\
The defined API is described as follows:\\
{\textasciigrave}{\textasciigrave}{\textasciigrave} python\\
\texttt{\{api\_description\}}\\
{\textasciigrave}{\textasciigrave}{\textasciigrave}
\\\\
You will use an iterative approach, interleaving "Thought", "Action", and "Observation" steps to collect information and perform the forecast. You may perform up to \hlc[yellow!30]{20} iterations. The steps are as follows:
\\\\
- "Thought": Analyze the current information and reason about the current situation, and predicts which API you want to use (try to use different APIs to collect diverse information) or make a decision that you want to make a final answer.\\
- "Action": Use the API to gather more information or provide the final forecast.\\
    \hspace*{6mm}- If gathering more data: \hlc[blue!20]{the action must be an executable Python code snippet that starts with {\textasciigrave}{\textasciigrave}{\textasciigrave}python and ends with {\textasciigrave}{\textasciigrave}{\textasciigrave}. It can contain multiple lines of codes and function calls using the defined API or Python libraries. You must use print() to output the results, and only the printed output will be returned in the observation step.}\\
    \hspace*{6mm}- If making the final forecast: the action must start immediately with "Final Answer:", and follow with the answer in the expected JSON format. This should not be enclosed within triple backticks.\\
- "Observation": Return the printed output of the executed code snippet.
\\\\
To make a reasonable forecast, you should collect both news and relational evidence to support your prediction. When you are fully confident that you accumulate enough information to make the final forecast, you should start the "Thought" with your reasoning using the news and structural information to make the prediction, and then start the "Action" step with "Final Answer:" followed by the answer in the expected JSON format. The answer should be a JSON dictionary where the keys are the forecasted two-digit first-level CAMEO codes and the values are lists of forecasted three-digit second-level CAMEO codes that are child relations of the key. For example, "Action: Final Answer: \{"01": ["010", "011", "012"], "02": ["020", "023"]\}".
\\\\
The final answer will be evaluated based on the precision and recall of the forecasted first-level and second-level relations, so only include confident first-level and second-level CAMEO codes in your final forecast.
\\\\
Try to use different APIs and Python libraries to collect diverse information (including multi-hop relations), such as the precise meaning of CAMEO codes, insights from news content, relational data, and statistical analyses to support your forecasts. Consider not only the frequency of the relations but also the temporal aspects of the data when making your forecast.

\subsubsection{Agent ReAct Process}
\begin{enumerate}
    \item \textbf{Thought}:
    To forecast the relations that Australia will take towards China on November 03, 2023, we need to gather historical data on the interactions between these two countries. This includes examining the frequency and types of relations observed in the past, as well as any recent trends or significant events that might influence future relations.

    We will proceed with the following steps:

    \begin{enumerate}
        \item \textbf{Retrieve Historical Events}: Gather events involving Australia (AUS) and China (CHN) to understand the context and types of relations that have occurred in the past.
        \item \textbf{Analyze Relation Distribution}: Analyze the distribution of different CAMEO codes in these events to identify the most common types of interactions.
        \item \textbf{Examine Recent Trends}: Focus on recent events to detect any emerging trends or shifts in relations.
        \item \textbf{Contextual Analysis}: Review news articles to gain insights into the context and underlying reasons for these relations.
    \end{enumerate}

    Let"s start by retrieving historical events between Australia and China.

    \textbf{Action}:
    \begin{lstlisting}[language=python]
from datetime import datetime

# Define the date range for the historical data collection
start_date = Date("2023-01-01")
end_date = Date("2023-11-02")
date_range = DateRange(start_date=start_date, end_date=end_date)

# Define the ISO codes for Australia and China
aus_code = ISOCode("AUS")
chn_code = ISOCode("CHN")

# Retrieve historical events involving Australia and China
events = get_events(date_range=date_range, head_entities=[aus_code], tail_entities=[chn_code])

# Print the retrieved events
for event in events:
    print(event)\end{lstlisting}
    \textbf{Observation}:
    \begin{lstlisting}[language=python]
Event(date=Date("2023-11-02"), head_entity=ISOCode("AUS"), relation=CAMEOCode("042"), tail_entity=ISOCode("CHN"))
Event(date=Date("2023-11-02"), head_entity=ISOCode("AUS"), relation=CAMEOCode("040"), tail_entity=ISOCode("CHN"))
Event(date=Date("2023-11-02"), head_entity=ISOCode("AUS"), relation=CAMEOCode("036"), tail_entity=ISOCode("CHN"))
Event(date=Date("2023-11-01"), head_entity=ISOCode("AUS"), relation=CAMEOCode("042"), tail_entity=ISOCode("CHN"))
Event(date=Date("2023-11-01"), head_entity=ISOCode("AUS"), relation=CAMEOCode("173"), tail_entity=ISOCode("CHN"))
Event(date=Date("2023-11-01"), head_entity=ISOCode("AUS"), relation=CAMEOCode("036"), tail_entity=ISOCode("CHN"))
Event(date=Date("2023-10-31"), head_entity=ISOCode("AUS"), relation=CAMEOCode("036"), tail_entity=ISOCode("CHN"))
Event(date=Date("2023-10-31"), head_entity=ISOCode("AUS"), relation=CAMEOCode("042"), tail_entity=ISOCode("CHN"))
Event(date=Date("2023-10-31"), head_entity=ISOCode("AUS"), relation=CAMEOCode("046"), tail_entity=ISOCode("CHN"))
Event(date=Date("2023-10-30"), head_entity=ISOCode("AUS"), relation=CAMEOCode("036"), tail_entity=ISOCode("CHN"))
Event(date=Date("2023-10-30"), head_entity=ISOCode("AUS"), relation=CAMEOCode("042"), tail_entity=ISOCode("CHN"))
Event(date=Date("2023-10-30"), head_entity=ISOCode("AUS"), relation=CAMEOCode("046"), tail_entity=ISOCode("CHN"))
Event(date=Date("2023-10-29"), head_entity=ISOCode("AUS"), relation=CAMEOCode("036"), tail_entity=ISOCode("CHN"))
Event(date=Date("2023-10-27"), head_entity=ISOCode("AUS"), relation=CAMEOCode("036"), tail_entity=ISOCode("CHN"))
Event(date=Date("2023-10-27"), head_entity=ISOCode("AUS"), relation=CAMEOCode("042"), tail_entity=ISOCode("CHN"))
Event(date=Date("2023-10-26"), head_entity=ISOCode("AUS"), relation=CAMEOCode("173"), tail_entity=ISOCode("CHN"))
Event(date=Date("2023-10-26"), head_entity=ISOCode("AUS"), relation=CAMEOCode("061"), tail_entity=ISOCode("CHN"))
Event(date=Date("2023-10-26"), head_entity=ISOCode("AUS"), relation=CAMEOCode("042"), tail_entity=ISOCode("CHN"))
Event(date=Date("2023-10-25"), head_entity=ISOCode("AUS"), relation=CAMEOCode("042"), tail_entity=ISOCode("CHN"))
Event(date=Date("2023-10-23"), head_entity=ISOCode("AUS"), relation=CAMEOCode("042"), tail_entity=ISOCode("CHN"))
Event(date=Date("2023-10-23"), head_entity=ISOCode("AUS"), relation=CAMEOCode("036"), tail_entity=ISOCode("CHN"))
Event(date=Date("2023-10-23"), head_entity=ISOCode("AUS"), relation=CAMEOCode("031"), tail_entity=ISOCode("CHN"))
Event(date=Date("2023-10-23"), head_entity=ISOCode("AUS"), relation=CAMEOCode("061"), tail_entity=ISOCode("CHN"))
Event(date=Date("2023-10-22"), head_entity=ISOCode("AUS"), relation=CAMEOCode("040"), tail_entity=ISOCode("CHN"))
Event(date=Date("2023-10-22"), head_entity=ISOCode("AUS"), relation=CAMEOCode("061"), tail_entity=ISOCode("CHN"))
Event(date=Date("2023-10-22"), head_entity=ISOCode("AUS"), relation=CAMEOCode("036"), tail_entity=ISOCode("CHN"))
Event(date=Date("2023-10-22"), head_entity=ISOCode("AUS"), relation=CAMEOCode("043"), tail_entity=ISOCode("CHN"))
Event(date=Date("2023-10-22"), head_entity=ISOCode("AUS"), relation=CAMEOCode("042"), tail_entity=ISOCode("CHN"))
Event(date=Date("2023-10-21"), head_entity=ISOCode("AUS"), relation=CAMEOCode("042"), tail_entity=ISOCode("CHN"))
Event(date=Date("2023-10-21"), head_entity=ISOCode("AUS"), relation=CAMEOCode("040"), tail_entity=ISOCode("CHN"))\end{lstlisting}
    \item \textbf{Thought}:
    The retrieved events indicate a variety of relations between Australia and China. To forecast the future relations, we need to analyze the distribution of these relations to identify the most common types and any recent trends.

    Next, we will analyze the distribution of the different CAMEO codes in these events to identify the most frequent interactions. This will help us understand the dominant types of relations and any emerging patterns.

    \textbf{Action}:
    \begin{lstlisting}[language=python]
from collections import Counter

# Extract the CAMEO codes from the retrieved events
cameo_codes = [event.relation.code for event in events]

# Count the frequency of each CAMEO code
cameo_code_counts = Counter(cameo_codes)

# Print the distribution of CAMEO codes
for code, count in cameo_code_counts.items():
    print(f"CAMEO Code: {code}, Count: {count}")\end{lstlisting}
    \textbf{Observation}:
    \begin{lstlisting}[language=python]
CAMEO Code: 042, Count: 10
CAMEO Code: 040, Count: 3
CAMEO Code: 036, Count: 8
CAMEO Code: 173, Count: 2
CAMEO Code: 046, Count: 2
CAMEO Code: 061, Count: 3
CAMEO Code: 031, Count: 1
CAMEO Code: 043, Count: 1\end{lstlisting}
    \item \textbf{Thought}:
    The distribution of CAMEO codes from the historical events between Australia and China reveals that certain types of interactions are more frequent than others. Specifically, the most common relations are:

    - \textbf{042}: 10 occurrences\\
    - \textbf{036}: 8 occurrences\\
    - \textbf{040}: 3 occurrences\\
    - \textbf{061}: 3 occurrences\\
    - \textbf{046}: 2 occurrences\\
    - \textbf{173}: 2 occurrences\\
    - \textbf{031}: 1 occurrence\\
    - \textbf{043}: 1 occurrence\\
    To make a more informed forecast, we should understand what these CAMEO codes represent. We will map these codes to their descriptions to gain insights into the nature of these interactions.

    \textbf{Action}:
    \begin{lstlisting}[language=python]
# Define the CAMEO codes to map
cameo_codes_to_map = ["042", "036", "040", "061", "046", "173", "031", "043"]

# Map the CAMEO codes to their descriptions
relations = [map_cameo_to_relation(CAMEOCode(code)) for code in cameo_codes_to_map]

# Print the descriptions of the CAMEO codes
for relation in relations:
    print(f"CAMEO Code: {relation.cameo_code.code}, Name: {relation.name}, Description: {relation.description}")\end{lstlisting}
    \textbf{Observation}:
    \begin{lstlisting}[language=python]
CAMEO Code: 042, Name: Make a visit, Description: Travel to another location for a meeting or other event.
CAMEO Code: 036, Name: Express intent to meet or negotiate, Description: Offer, promise, agree to, or otherwise indicate willingness or commitment to meet, visit, or engage in talks or negotiations.
CAMEO Code: 040, Name: Consult, not specified, Description: All consultations and meetings, not otherwise specified.
CAMEO Code: 061, Name: Cooperate economically, Description: Initiate, resume, improve, or expand economic exchange or cooperation.
CAMEO Code: 046, Name: Engage in negotiation, Description: Negotiate or bargain with others.
CAMEO Code: 173, Name: Arrest or detain, Description: Legal or extrajudicial arrests, detentions, or imprisonments.
CAMEO Code: 031, Name: Express intent to engage in material cooperation, Description: Offer, promise, agree to, or otherwise indicate willingness or commitment to engage in or expand material cooperative exchange, such as express intent to cooperate economically, militarily, on judicial matters, and on intelligence.
CAMEO Code: 043, Name: Host a visit, Description: Host or receive a visitor at residence, office or home country.\end{lstlisting}
    \item \textbf{Thought}:
    The mapped CAMEO codes and their descriptions provide valuable insights into the nature of interactions between Australia and China. Here are the key relations observed:

    - \textbf{042 (Make a visit)}: 10 occurrences – Indicates frequent travel for meetings or events.\\
    - \textbf{036 (Express intent to meet or negotiate)}: 8 occurrences – Shows a strong intent to engage in talks or negotiations.\\
    - \textbf{040 (Consult, not specified)}: 3 occurrences – General consultations and meetings.\\
    - \textbf{061 (Cooperate economically)}: 3 occurrences – Economic cooperation initiatives.\\
    - \textbf{046 (Engage in negotiation)}: 2 occurrences – Actual negotiations or bargaining.\\
    - \textbf{173 (Arrest or detain)}: 2 occurrences – Legal or extrajudicial arrests or detentions.\\
    - \textbf{031 (Express intent to engage in material cooperation)}: 1 occurrence – Intent to cooperate materially.\\
    - \textbf{043 (Host a visit)}: 1 occurrence – Hosting visitors.\\

    Given this data, we can see that the most frequent interactions involve visits, expressions of intent to meet or negotiate, and economic cooperation. These interactions suggest a pattern of diplomatic engagement and economic collaboration, with occasional legal actions.

    To further solidify our forecast, we should examine recent news articles to understand the context and any underlying reasons for these interactions. This will help us identify any emerging trends or significant events that might influence future relations.

    Next, we will retrieve recent news articles involving Australia and China to gain contextual insights.

    \textbf{Action}:
    \begin{lstlisting}[language=python]
# Define the date range for recent news articles
recent_start_date = Date("2023-10-01")
recent_end_date = Date("2023-11-02")
recent_date_range = DateRange(start_date=recent_start_date, end_date=recent_end_date)

# Retrieve recent news articles involving Australia and China
news_articles = get_news_articles(date_range=recent_date_range, head_entities=[aus_code], tail_entities=[chn_code])

# Print the retrieved news articles
for article in news_articles:
    print(article)\end{lstlisting}
    \textbf{Observation}:
    \begin{lstlisting}[language=python]
(Date(date="2023-11-02"), "No quick fix to remaining tariffs ahead of China visit")
(Date(date="2023-11-02"), "China"s aggression in Indo-Pacific leads to "concern and angst": Peter Dutton, Australia"s leader of opposition")
(Date(date="2023-11-02"), "Australia and China eye new ways to heal old wounds")
(Date(date="2023-11-02"), "Albanese must aim beyond bilateral outcomes on Beijing visit")
(Date(date="2023-11-02"), "Western Australian Premier Announces Visit to China")
(Date(date="2023-11-02"), "Australian PM"s China visit will help normalize relations, says ex-trade minister")
(Date(date="2023-11-02"), "Nicholas Moore on how to get in early on the next big growth story")
(Date(date="2023-11-02"), "PM"s China trip cements new era for ties")
(Date(date="2023-11-02"), "PM to take a hard line on trade with China")
(Date(date="2023-11-01"), "Australian Prime Minister Anthony Albanese to raise imprisoned democracy blogger during China visit")\end{lstlisting}
    \item \textbf{Thought}:
    The recent news articles provide valuable context for the relations between Australia and China. Key themes include:\\

    1. \textbf{Diplomatic Visits and Engagements}: Multiple articles highlight the upcoming visit of Australian Prime Minister Anthony Albanese to China. This aligns with the frequent occurrence of CAMEO code \textbf{042 (Make a visit)}.\\
    2. \textbf{Economic Cooperation and Trade}: Discussions around trade and economic cooperation are prominent, indicating ongoing efforts to improve economic ties. This supports the relevance of CAMEO codes \textbf{061 (Cooperate economically)} and \textbf{036 (Express intent to meet or negotiate)}.\\
    3. \textbf{Concerns and Tensions}: Articles mention concerns about China"s aggression in the Indo-Pacific and human rights issues, suggesting underlying tensions. This is reflected in the occasional occurrence of CAMEO code \textbf{173 (Arrest or detain)}.

    Given the historical data and recent news context, the most likely relations that Australia will take towards China on November 03, 2023, include diplomatic visits, economic cooperation, and negotiations, with a possibility of addressing underlying tensions.

    Final Answer:
    \begin{lstlisting}[language=python]
{
  "03": ["031", "036"],
  "04": ["040", "042", "043", "046"],
  "06": ["061"],
  "17": ["173"]
}\end{lstlisting}   

    \textbf{Action}:
    Final Answer:
    \begin{lstlisting}[language=python]
{
  "03": ["031", "036"],
  "04": ["040", "042", "043", "046"],
  "06": ["061"],
  "17": ["173"]
}\end{lstlisting}
\end{enumerate}

\subsubsection{Agent Final Status}
\begin{itemize}
    \item \textbf{End State}: Final Answer
    \item \textbf{Number of Steps Taken}: 5
    \item \textbf{Final Answer}:
    \begin{lstlisting}[language=python]
{
  "03": ["031", "036"],
  "04": ["040", "042", "043", "046"],
  "06": ["061"],
  "17": ["173"]
}\end{lstlisting}
We show the corresponding relation names of the predicted codes here for a better review:
        \begin{lstlisting}[language=python]
{
  "Express intent to cooperate": ["Express intent to engage in material cooperation", "Express intent to meet or negotiate"],
  "Consult": ["Consult, not specified", "Make a visit", "Host a visit", "Engage in negotiation"],
  "Engage in material cooperation": ["Cooperate economically"],
  "Coerce": ["Arrest or detain"]
}\end{lstlisting}
    \item \textbf{Ground Truth Answer}:
    \begin{lstlisting}[language=python]
{
  "03": ["036"],
  "04": ["042"],
  "17": ["172"]
}\end{lstlisting}
We show the corresponding relation names of the ground-truth codes here for a better review:
        \begin{lstlisting}[language=python]
{
  "Express intent to cooperate": ["Express intent to meet or negotiate"],
  "Consult": ["Make a visit"],
  "Coerce": ["Impose administrative sanctions"]
}\end{lstlisting}
\end{itemize}

\subsection{GPT-4o-based Agent with ReAct and ``\hlc[uclablue!20]{Single Function}'' Action}

\subsubsection{Query Details}
\begin{itemize}
    \item \textbf{Query Quadruplet}: (2023-11-03, AUS, ?, CHN)
    \item \textbf{Temporal Distance}: 1; therefore, the current date is 2023-11-02
    \item \textbf{Agent Max Steps}: 20
\end{itemize}

\subsubsection{Query Prompt}
Please forecast the relations that \hlc[yellow!30]{Australia} will take towards \hlc[yellow!30]{China} on \hlc[yellow!30]{November 03, 2023} based on historical information up to \hlc[yellow!30]{November 02, 2023}. I.e. forecast the relation CAMEO codes in query event \hlc[yellow!30]{Event(date=2023-11-03, head\_entity=ISOCode(AUS), relation=CAMEOCode(?), tail\_entity=ISOCode(CHN))}.

\subsubsection{System Prompt}
You are an expert in forecasting future events based on historical data. The database contains news articles from January 1, 2023 to the current date \hlc[yellow!30]{November 02, 2023} and the events extracted from these articles. The events are in the form of (date, subject country, relation, object country), where the countries are represented by ISO 3166-1 alpha-3 codes and the relations are represented by the CAMEO codes defined in the "Conflict and Mediation Event Observations" ontology. The relations are hierarchical: first-level relations are general parent relations represented by two-digit CAMEO codes, while second-level relations are more specific child relations represented by three-digit CAMEO codes. Child relations have the same first two digits as their parent relations. For example, "01" is a first-level relation, and "010" and "011" are some of its second-level relations. The relations in the database are represented in the second-level form.
\\\\
Your task is to forecast the future relations between two entities in a given query. You have access to a defined Python API that allows you to query the database for historical events and statistics, and to get precise information about the ISO country codes and CAMEO relation codes.
\\\\
The defined API is described as follows:\\
{\textasciigrave}{\textasciigrave}{\textasciigrave} python\\
\texttt{\{api\_description\}}\\
{\textasciigrave}{\textasciigrave}{\textasciigrave}
\\\\
You will use an iterative approach, interleaving "Thought", "Action", and "Observation" steps to collect information and perform the forecast. You may perform up to \hlc[yellow!30]{20} iterations. The steps are as follows:
\\\\
- "Thought": Analyze the current information and reason about the current situation, and predicts which API you want to use (try to use different APIs to collect diverse information) or make a decision that you want to make a final answer.\\
- "Action": Use the API to gather more information or provide the final forecast.\\
    \hspace*{6mm}- If using the API: \hlc[uclablue!20]{the action must be only one single line of exactly one function call from the API with appropriate inputs, without additional code, explanations, or natural language descriptions}.\\
    \hspace*{6mm}- If making the final forecast: the action must start immediately with "Final Answer:", and follow with the results in the expected JSON format.\\
- "Observation": Return the output of the called function.
\\\\
To make a reasonable forecast, you should collect both news and relational evidence to support your prediction. When you are fully confident that you accumulate enough information to make the final forecast, you should start the "Thought" with your reasoning using the news and structural information to make the prediction, and then start the "Action" step with "Final Answer:" followed by the answer in the expected JSON format. The answer should be a JSON dictionary where the keys are the forecasted two-digit first-level CAMEO codes and the values are lists of forecasted three-digit second-level CAMEO codes that are child relations of the key. For example, "Action: Final Answer: \{"01": ["010", "011", "012"], "02": ["020", "023"]\}".
\\\\
The final answer will be evaluated based on the precision and recall of the forecasted first-level and second-level relations, so only include confident first-level and second-level CAMEO codes in your final forecast.
\\\\
Try to use different APIs to collect diverse information (including multi-hop relations), such as the precise meaning of CAMEO codes, insights from news content, relational data, and statistical analyses to support your forecasts. Consider not only the frequency of the relations but also the temporal aspects of the data when making your forecast.

\subsubsection{Agent ReAct Process}
\begin{enumerate}
    \item \textbf{Thought}:
    To forecast the relations that Australia (AUS) will take towards China (CHN) on November 03, 2023, I will start by gathering historical data on the interactions between these two countries. This will include the types of relations, their frequencies, and the context in which they occurred. I will use the following steps:

    \begin{enumerate}
        \item Retrieve the distribution of second-level relations between Australia and China.
        \item Analyze the distribution to identify the most common types of interactions.
        \item Examine recent news articles to understand the context and any ongoing trends or events that might influence future relations.
    \end{enumerate}

    \textbf{Action}:
    \begin{lstlisting}[language=python]
get_relation_distribution(head_entities=[ISOCode("AUS")], tail_entities=[ISOCode("CHN")])\end{lstlisting}
    \textbf{Observation}:
    \begin{lstlisting}[language=python]
{CAMEOCode(code="042"): 32,
 CAMEOCode(code="036"): 30,
 CAMEOCode(code="040"): 17,
CAMEOCode(code="061"): 16,
CAMEOCode(code="046"): 12,
CAMEOCode(code="043"): 11,
CAMEOCode(code="010"): 6,
CAMEOCode(code="020"): 5,
CAMEOCode(code="173"): 5,
CAMEOCode(code="012"): 5,
CAMEOCode(code="051"): 4,
CAMEOCode(code="163"): 3,
CAMEOCode(code="128"): 2,
CAMEOCode(code="050"): 2,
CAMEOCode(code="084"): 2,
CAMEOCode(code="090"): 2,
CAMEOCode(code="174"): 1,
CAMEOCode(code="112"): 1,
CAMEOCode(code="130"): 1,
CAMEOCode(code="160"): 1,
CAMEOCode(code="172"): 1,
CAMEOCode(code="192"): 1,
CAMEOCode(code="190"): 1,
CAMEOCode(code="060"): 1,
CAMEOCode(code="030"): 1,
CAMEOCode(code="035"): 1,
CAMEOCode(code="138"): 1,
CAMEOCode(code="071"): 1,
CAMEOCode(code="031"): 1}\end{lstlisting}
    \item \textbf{Thought}:
    The distribution of second-level relations between Australia and China shows a variety of interactions, with some relations being more frequent than others. The most common interactions include:

    - "042" (Consult): 32 occurrences\\
    - "036" (Express intent to cooperate): 30 occurrences\\
    - "040" (Assure): 17 occurrences\\
    - "061" (Cooperate economically): 16 occurrences\\
    - "046" (Praise or endorse): 12 occurrences\\
    - "043" (Apologize): 11 occurrences\\

    These frequent interactions suggest a trend towards diplomatic and cooperative engagements. However, to make a more informed forecast, it is essential to understand the context and any recent developments that might influence future relations.

    Next, I will retrieve recent news articles involving Australia and China to gather insights on the current context and any ongoing trends or events.

    \textbf{Action}:
    \begin{lstlisting}[language=python]
get_news_articles(date_range=DateRange(start_date=Date("2023-10-01"), end_date=Date("2023-11-02")), head_entities=[ISOCode("AUS")], tail_entities=[ISOCode("CHN")])\end{lstlisting}
    \textbf{Observation}:
    \begin{lstlisting}[language=python]
[(Date(date="2023-11-02"), "No quick fix to remaining tariffs ahead of China visit"),
 (Date(date="2023-11-02"), "China"s aggression in Indo-Pacific leads to "concern and angst": Peter Dutton, Australia"s leader of opposition"),
 (Date(date="2023-11-02"), "Australia and China eye new ways to heal old wounds"),
 (Date(date="2023-11-02"), "Albanese must aim beyond bilateral outcomes on Beijing visit"),
 (Date(date="2023-11-02"), "Western Australian Premier Announces Visit to China"),
 (Date(date="2023-11-02"), "Australian PM"s China visit will help normalize relations, says ex-trade minister"),
 (Date(date="2023-11-02"), "Nicholas Moore on how to get in early on the next big growth story"),
 (Date(date="2023-11-02"), "PM"s China trip cements new era for ties"),
 (Date(date="2023-11-02"), "PM to take a hard line on trade with China"),
 (Date(date="2023-11-01"), "Australian Prime Minister Anthony Albanese to raise imprisoned democracy blogger during China visit")]\end{lstlisting}
    \item \textbf{Thought}:
    The recent news articles indicate a mix of diplomatic, economic, and security-related interactions between Australia and China. Key themes include:

    - Diplomatic engagements and efforts to normalize relations, as seen in articles about the Australian Prime Minister"s visit to China and efforts to heal old wounds.\\
    - Economic cooperation and trade discussions, including addressing tariffs and trade issues.\\
    - Security concerns and criticisms, particularly related to China"s actions in the Indo-Pacific region and human rights issues.\\

    Given these themes, it is likely that the future relations between Australia and China will involve a combination of diplomatic assurances, economic cooperation, and some critical statements or actions related to security and human rights.

    Next, I will retrieve detailed descriptions of the most frequent second-level relations to ensure accurate mapping and understanding of these interactions.

    \textbf{Action}:
    \begin{lstlisting}[language=python]
map_cameo_to_relation(CAMEOCode("042"))\end{lstlisting}
    \textbf{Observation}:
    \begin{lstlisting}[language=python]
Relation(cameo_code=CAMEOCode("042"), name="Make a visit", description="Travel to another location for a meeting or other event.")\end{lstlisting}
    \item \textbf{Thought}:
    The relation "042" (Make a visit) involves traveling to another location for a meeting or other event, which aligns with the recent news about the Australian Prime Minister"s visit to China. This suggests that diplomatic visits are a significant aspect of the interactions between Australia and China.

    To further refine the forecast, I will retrieve detailed descriptions for the other most frequent second-level relations: "036", "040", "061", "046", and "043".

    \textbf{Action}:
    \begin{lstlisting}[language=python]
map_cameo_to_relation(CAMEOCode("036"))\end{lstlisting}
    \textbf{Observation}:
    \begin{lstlisting}[language=python]
Relation(cameo_code=CAMEOCode("036"), name="Express intent to meet or negotiate", description="Offer, promise, agree to, or otherwise indicate willingness or commitment to meet, visit, or engage in talks or negotiations.")\end{lstlisting}
    \item \textbf{Thought}:
    The relation "036" (Express intent to meet or negotiate) involves indicating a willingness or commitment to meet, visit, or engage in talks or negotiations. This aligns with the ongoing diplomatic efforts and discussions mentioned in the recent news articles.

    Next, I will retrieve detailed descriptions for the relations "040", "061", "046", and "043" to further understand the context of these interactions.

    \textbf{Action}:
    \begin{lstlisting}[language=python]
get_parent_relation(CAMEOCode("040"))\end{lstlisting}
    \textbf{Observation}:
    \begin{lstlisting}[language=python]
Relation(cameo_code=CAMEOCode("04"), name="Consult", description="All consultations and meetings.")\end{lstlisting}
    \item \textbf{Thought}:
    The parent relation `04` (Consult) encompasses all consultations and meetings, which is consistent with the frequent diplomatic interactions between Australia and China. The specific relation "040" (Assure) likely involves providing assurances during these consultations.

    To further refine the forecast, I will retrieve detailed descriptions for the relations "061", "046", and "043".

    \textbf{Action}:""

    \textbf{Observation}: No feedback from the environment due to the null action.
    \item \textbf{Thought}:
    I will continue retrieving detailed descriptions for the relations "061", "046", and "043" to further understand the context of these interactions.

    \textbf{Action}:""

    \textbf{Observation}: No feedback from the environment due to the null action.
    \item \textbf{Thought}:
    It seems there was an issue with my previous actions. I will now proceed to retrieve the detailed descriptions for the remaining frequent second-level relations: "061", "046", and "043".
    \textbf{Action}:
    \begin{lstlisting}[language=python]
map_cameo_to_relation(CAMEOCode("061"))\end{lstlisting}
    \textbf{Observation}:
    \begin{lstlisting}[language=python]
Relation(cameo_code=CAMEOCode("061"), name="Cooperate economically", description="Initiate, resume, improve, or expand economic exchange or cooperation.")\end{lstlisting}
    \item \textbf{Thought}:
    The relation "061" (Cooperate economically) involves initiating, resuming, improving, or expanding economic exchange or cooperation. This aligns with the recent news articles discussing trade issues and economic cooperation between Australia and China.

    Next, I will retrieve detailed descriptions for the relations "046" and "043".

    \textbf{Action}:
    \begin{lstlisting}[language=python]
map_cameo_to_relation(CAMEOCode("046"))\end{lstlisting}
    \textbf{Observation}:
    \begin{lstlisting}[language=python]
Relation(cameo_code=CAMEOCode("046"), name="Engage in negotiation", description="Negotiate or bargain with others.")\end{lstlisting}

    \item \textbf{Thought}:
    The relation "046" (Engage in negotiation) involves negotiating or bargaining with others. This is consistent with the ongoing discussions and negotiations mentioned in the news articles, particularly regarding trade and diplomatic issues.

    Next, I will retrieve the detailed description for the relation "043".

    \textbf{Action}:""

    \textbf{Observation}: No feedback from the environment due to the null action.

    \item \textbf{Thought}:
    It seems there was an issue with my previous actions. I will now proceed to retrieve the detailed description for the relation "043" to complete the understanding of the most frequent interactions between Australia and China.

    \textbf{Action}:
    \begin{lstlisting}[language=python]
map_cameo_to_relation(CAMEOCode("043"))\end{lstlisting}
    \textbf{Observation}:
    \begin{lstlisting}[language=python]
Relation(cameo_code=CAMEOCode("043"), name="Host a visit", description="Host or receive a visitor at residence, office or home country.")\end{lstlisting}

    \item \textbf{Thought}:
    The relation "043" (Host a visit) involves hosting or receiving a visitor at a residence, office, or home country. This aligns with the diplomatic visits and engagements mentioned in the recent news articles.

    Based on the detailed descriptions and the recent news context, the most frequent interactions between Australia and China involve diplomatic visits, economic cooperation, negotiations, and assurances. These interactions suggest a trend towards diplomatic and cooperative engagements, with some critical statements or actions related to security and human rights.

    Given this information, I am confident in forecasting the following relations between Australia and China on November 03, 2023:

    - Diplomatic visits and engagements ("042", "043", "036")
    - Economic cooperation ("061")
    - Negotiations ("046")
    - Assurances ("040")

    \textbf{Action}:
    Final Answer: \{"04": ["040", "042", "043", "046"], "03": ["036"], "06": ["061"]\}
\end{enumerate}

\subsubsection{Agent Final Status}
\begin{itemize}
    \item \textbf{End State}: Final Answer
    \item \textbf{Number of Steps Taken}: 12
    \item \textbf{Final Answer}:
    \begin{lstlisting}[language=python]
{
  "03": ["036"],
  "04": ["040", "042", "043", "046"],
  "06": ["061"]
}\end{lstlisting}
We show the corresponding relation names of the predicted codes here for a better review:
        \begin{lstlisting}[language=python]
{
  "Express intent to cooperate": ["Express intent to meet or negotiate"],
  "Consult": ["Consult, not specified", "Make a visit", "Host a visit", "Engage in negotiation"],
  "Engage in material cooperation": ["Cooperate economically"]
}\end{lstlisting}
    \item \textbf{Ground Truth Answer}:
    \begin{lstlisting}[language=python]
{
  "03": ["036"],
  "04": ["042"],
  "17": ["172"]
}\end{lstlisting}
We show the corresponding relation names of the ground-truth codes here for a better review:
        \begin{lstlisting}[language=python]
{
  "Express intent to cooperate": ["Express intent to meet or negotiate"],
  "Consult": ["Make a visit"],
  "Coerce": ["Impose administrative sanctions"]
}\end{lstlisting}
\end{itemize}

\newpage
\section{Additional Information about API}
\label{append:api}

\subsection{Overview of API Data Classes and Functions}

\begin{table}[h]
\small{
    \centering
    \caption{API data classes and their attributes}
    \resizebox{0.99\textwidth}{!}{%
    \begin{tabular}{ll}
    \toprule
    \textbf{Class Name} & \textbf{Attributes and Types} \\
    \midrule
    \texttt{Date} & \lstinline|date: str  # `YYYY-MM-DD'| \\
    \midrule
    \texttt{DateRange} & \lstinline|start_date: Optional[Date]|, \lstinline|end_date: Optional[Date]| \\
    \midrule
    \texttt{ISOCode} & \lstinline|code: str  # 3-letter ISO code| \\
    \midrule
    \texttt{Country} & \lstinline|iso_code: ISOCode|, \lstinline|name: str| \\
    \midrule
    \texttt{CAMEOCode} & \lstinline|code: str  # CAMEO code| \\
    \midrule
    \texttt{Relation} & \lstinline|cameo_code: CAMEOCode|, \lstinline|name: str|, \lstinline|description: str| \\
    \midrule
    \texttt{Event} & \lstinline|date: Date|, \lstinline|head_entity: ISOCode|, \lstinline|relation: CAMEOCode|, \lstinline|tail_entity: ISOCode| \\
    \midrule
    \texttt{NewsArticle} & \lstinline|date: Date|, \lstinline|title: str|, \lstinline|content: str|, \lstinline|events: List[Event]| \\
    \bottomrule
    \end{tabular}
    }
    }
\end{table}

\begin{table}[h]
\small{
    \centering
    \caption{API functions categorized by functionality}
    \begin{tabular}{>{\raggedright\arraybackslash}p{0.95\linewidth}}
    \toprule
    \textbf{Functions related to Countries and Relations} \\
    \midrule
    \lstinline|map_country_name_to_iso(name: str) -> List[Country]| \\
    \lstinline|map_iso_to_country_name(iso_code: ISOCode) -> str| \\
    \lstinline|map_relation_description_to_cameo(description: str) -> List[Relation]| \\
    \lstinline|map_cameo_to_relation(cameo_code: CAMEOCode) -> Relation| \\
    \lstinline|get_parent_relation(cameo_code: CAMEOCode) -> Relation| \\
    \lstinline|get_child_relations(cameo_code: CAMEOCode) -> List[Relation]| \\
    \lstinline|get_sibling_relations(cameo_code: CAMEOCode) -> List[Relation]| \\
    \midrule
    \textbf{Functions related to Events} \\
    \midrule
    \lstinline|count_events(date_range: Optional[DateRange], head_entities: Optional[List[ISOCode]], tail_entities: Optional[List[ISOCode]], relations: Optional[List[CAMEOCode]]) -> int| \\
    \addlinespace
    \lstinline|get_events(date_range: Optional[DateRange], head_entities: Optional[List[ISOCode]], tail_entities: Optional[List[ISOCode]], relations: Optional[List[CAMEOCode]], text_description: Optional[str]) -> List[Event]| \\
    \addlinespace
    \lstinline|get_entity_distribution(date_range: Optional[DateRange], involved_relations: Optional[List[CAMEOCode]], interacted_entities: Optional[List[ISOCode]], entity_role: Optional[str]) -> Dict[ISOCode, int]| \\
    \addlinespace
    \lstinline|get_relation_distribution(date_range: Optional[DateRange], head_entities: Optional[List[ISOCode]], tail_entities: Optional[List[ISOCode]]) -> Dict[CAMEOCode, int]| \\
    \midrule
    \textbf{Functions related to News} \\
    \midrule
    \lstinline|count_news_articles(date_range: Optional[DateRange], head_entities: Optional[List[ISOCode]], tail_entities: Optional[List[ISOCode]], relations: Optional[List[CAMEOCode]], keywords: Optional[List[str]]) -> int| \\
    \addlinespace
    \lstinline|get_news_articles(date_range: Optional[DateRange], head_entities: Optional[List[ISOCode]], tail_entities: Optional[List[ISOCode]], relations: Optional[List[CAMEOCode]], keywords: Optional[List[str]], text_description: Optional[str]) -> List[Tuple[Date, str]]| \\
    \addlinespace
    \lstinline|browse_news_article(date: Date, title: str) -> str| \\
    \bottomrule
    \end{tabular}
    }
\end{table}

\clearpage

\subsection{API Specification}

In this section, we provide the API specification, which is part of the system prompt for the ReAct agent. It contains the definition of data classes and functions, with detailed Python docstring. For dataclass, the specification includes class definition and description, attributes and their description, and the example of initializing the class object. For functions, the specification includes the function definition and description, the parameter and description, the return and description, and the example of usage.

\begin{lstlisting}
@dataclass
class Date:
    """Represents a date."""
    date: str # Date in the format 'YYYY-MM-DD'
    # Example: Date("2022-01-01")

@dataclass
class DateRange:
    """Represents a range of dates (inclusive)."""
    start_date: Optional[Date] # If None, the earliest date is used
    end_date: Optional[Date] # If None, the current date is used
    # Example: DateRange(start_date=Date("2022-01-01"), end_date=Date("2022-01-31"))

@dataclass
class ISOCode:
    """Represents an ISO alpha-3 country code."""
    code: str # 3-letter ISO code
    # Example: ISOCode("USA")

@dataclass
class Country:
    """Represents a country entity."""
    iso_code: ISOCode
    name: str
    # Example: Country(iso_code=ISOCode("USA"), name="United States")

@dataclass
class CAMEOCode:
    """Represents a CAMEO verb code."""
    code: str # 2-digit CAMEO code for first level relations, 3-digit CAMEO code for second level relations
    # Example: CAMEOCode("01"), CAMEOCode("010")

@dataclass
class Relation:
    """Represents a relation."""
    cameo_code: CAMEOCode
    name: str
    description: str # A brief description of what event the relation represents
    # Example: Relation(cameo_code=CAMEOCode("010"), name="Make statement, not specified", description="All public statements expressed verbally or in action, not otherwise specified."

@dataclass
class Event:
    """Represents an event characterized by date, head entity, relation, and tail entity."""
    date: Date
    head_entity: ISOCode
    relation: CAMEOCode
    tail_entity: ISOCode
    # Example: Event(date=Date("2022-01-01"), head_entity=ISOCode("USA"), relation=CAMEOCode("010"), tail_entity=ISOCode("CAN"))

@dataclass
class NewsArticle:
    """Represents a news article, including metadata and content."""
    date: Date
    title: str
    content: str # Full text content of the news article
    events: List[Event] # List of events mentioned in the article
    # Example: NewsArticle(date=Date("2022-01-01"), title="Trade agreement between USA and China", content="On January 1, 2022, a trade agreement was signed between the USA and China...", events=[Event(date=Date("2022-01-01"), head_entity=ISOCode("USA"), relation=CAMEOCode("010"), tail_entity=ISOCode("CHN")])

    def __str__(self):
        return f"{self.date}:\n{self.title}\n{self.content}"

def map_country_name_to_iso(name: str) -> List[Country]:
    """
    Maps a country name to the most likely corresponding Country objects with ISO codes.

    Parameters:
        name (str): The country name to map.

    Returns:
        List[Country]: A list of 5 most likely Country objects sorted by relevance.

    Example:
        >>> map_country_name_to_iso("Korea")
        [Country(iso_code=ISOCode("KOR"), name="Republic of Korea"), Country(iso_code=ISOCode("PRK"), name="Democratic People's Republic of Korea")]
    """
    pass

def map_iso_to_country_name(iso_code: ISOCode) -> str:
    """
    Maps an ISO code to a country name.

    Parameters:
        iso_code (ISOCode): The ISO code to map.

    Returns:
        str: The corresponding country name.

    Example:
        >>> map_iso_to_country_name(ISOCode("CHN"))
        "China"
    """
    pass

def map_relation_description_to_cameo(description: str) -> List[Relation]:
    """
    Maps a relation description to the most likely Relation objects.

    Parameters:
        description (str): The relation description to map.

    Returns:
        List[Relation]: A list of 5 most likely Relations sorted by relevance.

    Example:
        >>> map_relation_description_to_cameo("Fight with guns")
        [Reltion(cameo_code=CAMEOCode("19"), name="Fight", description="All uses of conventional force and acts of war typically by organized armed groups."), Relation(cameo_code=CAMEOCode("193"), name="Fight with small arms and light weapons", description="Attack using small arms and light weapons such as rifles, machine-guns, and mortar shells."), Relation(cameo_code=CAMEOCode("190"), name="Use conventional military force, not specified", description="All uses of conventional force and acts of war typically by organized armed groups, not otherwise specified.")]
    """
    pass

def map_cameo_to_relation(cameo_code: CAMEOCode) -> Relation:
    """
    Maps a CAMEO code to a relation, including its name and description.

    Parameters:
        cameo_code (CAMEOCode): The CAMEO code to map.

    Returns:
        Relation: The corresponding relation.

    Example:
        >>> map_cameo_to_relation(CAMEOCode("190"))
        Relation(cameo_code=CAMEOCode("190"), name="Use conventional military force, not specified", description="All uses of conventional force and acts of war typically by organized armed groups, not otherwise specified.")
    """
    pass

def get_child_relations(cameo_code: CAMEOCode) -> List[Relation]:
    """
    Retrieves child relations of a given relation identified by CAMEO code.

    Parameters:
        cameo_code (CAMEOCode): The CAMEO code of the relation whose children are sought. Only first level relations are accepted.

    Returns:
        List[Relation]: A list of second level child relations.

    Example:
        >>> get_child_relations(CAMEOCode("19"))
        [Relation(caemo_code=CAMEOCode("190"), name="Use conventional military force, not specified", description="All uses of conventional force and acts of war typically by organized armed groups, not otherwise specified."), Relation(cameo_code=CAMEOCode("191"), name="Impose blockade or restrict movement", description="Prevent entry into and/or exit from a territory using armed forces."), ...]
    """
    pass

def get_sibling_relations(cameo_code: CAMEOCode) -> List[Relation]:
    """
    Retrieves sibling relations of a given relation identified by CAMEO code.

    Parameters:
        cameo_code (CAMEOCode): The CAMEO code of the relation whose siblings are sought. Both first and second level relations are accepted.

    Returns:
        List[Relation]: A list of sibling relations at the same level.

    Example:
        >>> get_sibling_relations(CAMEOCode("193"))
        [Relation(caemo_code=CAMEOCode("190"), name="Use conventional military force, not specified", description="All uses of conventional force and acts of war typically by organized armed groups, not otherwise specified."), Relation(cameo_code=CAMEOCode("191"), name="Impose blockade or restrict movement", description="Prevent entry into and/or exit from a territory using armed forces."), ...]
    """
    pass

def count_events(date_range: Optional[DateRange] = None, head_entities: Optional[List[ISOCode]] = None, tail_entities: Optional[List[ISOCode]] = None, relations: Optional[List[CAMEOCode]] = None) -> int:
    """
    Counts the number of events in the knowledge graph based on specified conditions.

    Parameters:
        date_range (Optional[DateRange]): Range of dates to filter the events. If None, all dates are included.
        head_entities (Optional[List[ISOCode]]): List of head entity ISO codes to be included. If None, all countries are included.
        tail_entities (Optional[List[ISOCode]]): List of tail entity ISO codes to be included. If None, all countries are included.
        relations (Optional[List[CAMEOCode]]): List of relation CAMEO codes to be included. If first level relations are listed, all second level relations under them are included. If None, all relations are included.

    Returns:
        int: Count of unique events matching the conditions.

    Example:
        >>> count_events(date_range=DateRange(start_date=Date("2022-01-01"), end_date=Date("2022-01-31")), head_entities=[ISOCode("USA"), ISOCode("CHN")], tail_entities=None, relations=[CAMEOCode("010")])
        4
    """
    pass

def get_events(date_range: Optional[DateRange] = None, head_entities: Optional[List[ISOCode]] = None, tail_entities: Optional[List[ISOCode]] = None, relations: Optional[List[CAMEOCode]] = None, text_description: Optional[str] = None) -> List[Event]:
    """
    Retrieves events from the knowledge graph based on specified conditions.
    Inherits common filter parameters from count_events. See count_events for more details on these parameters.

    Additional Parameters:
        text_description (Optional[str]): Textual description to match with the source news articles of events. If None, the returned events are sorted by date in descending order; otherwise, sorted by relevance of the source news article to the description.

    Returns:
        List[Event]: A list of maximum 30 events matching the specified conditions.

    Example:
        >>> get_events(date_range=DateRange(start_date=Date("2022-01-01"), end_date=Date("2022-01-31")), head_entities=[ISOCode("USA"), ISOCode("CHN")], tail_entities=None, relations=[CAMEOCode("010")], text_description="economic trade")
        [Event(date=Date("2022-01-15"), head_entity=ISOCode("USA"), relation=CAMEOCode("010"), tail_entity=ISOCode("CAN"), Event(date=Date("2022-01-10"), head_entity=ISOCode("CHN"), relation=CAMEOCode("010"), tail_entity=ISOCode("USA"), ...]
    """
    pass

def get_entity_distribution(date_range: Optional[DateRange] = None, involved_relations: Optional[List[CAMEOCode]] = None, interacted_entities: Optional[List[ISOCode]] = None, entity_role: Optional[str] = None) -> Dict[ISOCode, int]:
    """
    Gets the distribution of entities in the knowledge graph under specified conditions.

    Parameters:
        date_range (Optional[DateRange]): Range of dates to filter the events. If None, all dates are included.
        involved_relations (Optional[List[CAMEOCode]]): List of relations that the returned entities must be involved in any of. If first level relations are listed, all second level relations under them are included. If None, all relations are included.
        interacted_entities (Optional[List[ISOCode]]): List of entities that the returned entities must have interacted with any of. If None, all entities are included.
        entity_role (Optional[EntityRole]): Specifies the role of the returned entity in the events. Options are 'head', 'tail', or 'both'. If 'both' or None, the returned entity can be either head or tail.

    Returns:
        Dict[ISOCode, int]: A dictionary mapping returned entities' ISO codes to the number of events with the specified conditions in which they are involved, sorted by counts in descending order.

    Example:
        >>> get_entity_distribution(date_range=DateRange(start_date=Date("2022-01-01"), end_date=Date("2022-01-31")), involved_relations=[CAMEOCode("010")], interacted_entities=[ISOCode("USA"), ISOCode("CHN")], entity_role="tail")
        {ISOCode("USA"): 3, ISOCode("CHN"): 1}
    """
    pass

def get_relation_distribution(date_range: Optional[DateRange] = None, head_entities: Optional[List[ISOCode]] = None, tail_entities: Optional[List[ISOCode]] = None) -> Dict[CAMEOCode, int]:
    """
    Gets the distribution of second level relations in the knowledge graph under specified conditions.

    Parameters:
        date_range (Optional[DateRange]): Range of dates to filter the events. If None, all dates are included.
        head_entities (Optional[List[ISOCode]]): List of head entities that the events must involve any of. If None, all head entities are included.
        tail_entities (Optional[List[ISOCode]]): List of tail entities that the events must involve any of. If None, all tail entities are included.

    Returns:
        Dict[CAMEOCode, int]: A dictionary mapping second level relations' CAMEO codes to the number of events with the specified conditions in which they are involved, sorted by counts in descending order.

    Example:
        >>> get_relation_distribution(date_range=DateRange(start_date=Date("2022-01-01"), end_date=Date("2022-01-31")), head_entities=[ISOCode("USA"), ISOCode("CHN")], tail_entities=None)
        {CAMEOCode("010"): 3, CAMEOCode("011"): 1}
    """
    pass

def count_news_articles(date_range: Optional[DateRange] = None, head_entities: Optional[List[ISOCode]] = None, tail_entities: Optional[List[ISOCode]] = None, relations: Optional[List[CAMEOCode]] = None, keywords: Optional[List[str]] = None) -> int:
    """
    Counts the number of news articles based on specified conditions.

    Parameters:
        date_range (Optional[DateRange]): Range of dates to filter the news articles. If None, all dates are included.
        head_entities (Optional[List[ISOCode]]): At least one of the entities must be mentioned in the articles and be the head entity in the events. If None, all entities are included.
        tail_entities (Optional[List[ISOCode]]): At least one of the entities must be mentioned in the articles and be the tail entity in the events. If None, all entities are included.
        relations (Optional[List[CAMEOCode]]): At least one of the relations must be mentioned in the articles. If first level relations are listed, all second level relations under them are included. If None, all relations are included.
        keywords (Optional[List[str]]): At least one of the keywords must be present in the articles. If None, all articles are included.

    Returns:
        int: The count of news articles matching the conditions.

    Example:
        >>> count_news_articles(date_range=DateRange(start_date=Date("2022-01-01"), end_date=Date("2022-01-31")), head_entities=[ISOCode("USA"), ISOCode("CHN")], tail_entities=[ISOCode("USA"), ISOCode("CHN")], relations=[CAMEOCode("010")], keywords=["trade"])
        2
    """
    pass

def get_news_articles(date_range: Optional[DateRange] = None, head_entities: Optional[List[ISOCode]] = None, tail_entities: Optional[List[ISOCode]] = None, relations: Optional[List[CAMEOCode]] = None, keywords: Optional[List[str]] = None, text_description: Optional[str] = None) -> List[Tuple[Date, str]]:
    """
    Retrieves news articles based on specified conditions.
    Inherits common filter parameters from count_news_articles. See count_news_articles for more details on these parameters.

    Additional Parameters:
        text_description (Optional[str]): Textual description to match with the news articles. If None, the returned articles are sorted by date in descending order; otherwise, sorted by relevance to the description.

    Returns:
        List[Tuple[Date, str]]: A list of maximum 15 news articles matching the specified conditions, each represented by a tuple of date and title.

    Example:
        >>> get_news_articles(date_range=DateRange(start_date=Date("2022-01-01"), end_date=Date("2022-01-31")), head_entities=[ISOCode("USA"), ISOCode("CHN")], tail_entities=[ISOCode("USA"), ISOCode("CHN")], relations=[CAMEOCode("010")], keywords=["trade"], text_description="Economic trade is encouraged between USA and China.")
        [(NewsArticle.date=Date("2022-01-15"), NewsArticle.title="China and USA sign trade deal"), (NewsArticle.date=Date("2022-01-10"), NewsArticle.title="Trade agreement between USA and China")]
    """
    pass

def browse_news_article(date: Date, title: str) -> str:
    """
    Retrieves the full text of a news article by its title.

    Parameters:
        date (Date): The date of the news article to retrieve.
        title (str): The title of the news article to retrieve.

    Returns:
        str: The date, the title and full contents of the news article.

    Example:
        >>> browse_news_article(Date("2022-01-10"), "Trade agreement between USA and China")
        2022-01-10:
        Trade agreement between USA and China
        On January 10, 2022, a trade agreement was signed between the USA and China to promote economic cooperation...
    """
    pass
\end{lstlisting}

\clearpage

\subsection{API Implementation}
In this section, we provide the full implementation of the API, which is part of the environment and is accessible to the Python executor.

\begin{lstlisting}
@dataclass
class Date:
    """Represents a date."""
    date: str # Date in the format 'YYYY-MM-DD'
    # Example: Date("2022-01-01")

    def __init__(self, date: str):
        # check type
        if not isinstance(date, str):
            raise ValueError(f"Attribute 'date' of class Date must be a string in the format 'YYYY-MM-DD', but received: {date} in type {type(date)}")

        # check if date is in the correct format by trying to convert it to a date object
        try:
            datetime.datetime.strptime(date, '%Y-%m-%d')
        except ValueError:
            raise ValueError(f"Date must be in the format 'YYYY-MM-DD', but received: {date}")
        if date < DEFAULT_START_DATE:
            raise ValueError(f"Date must be on or after {DEFAULT_START_DATE}, but received: {date}")
        if DEFAULT_END_DATE and date > DEFAULT_END_DATE:
            raise ValueError(f"Date must be on or before the current date {DEFAULT_END_DATE}, but received: {date}")

        self.date = date

    def __str__(self):
        return f"Date('{self.date}')"

    def __hash__(self):
        return hash(self.date)

    def __eq__(self, other):
        if isinstance(other, Date):
            return self.date == other.date
        return False

@dataclass
class DateRange:
    """Represents a range of dates (inclusive)."""
    start_date: Optional[Date] # If None, the earliest date is used
    end_date: Optional[Date] # If None, the current date is used
    # Example: DateRange(start_date=Date("2022-01-01"), end_date=Date("2022-01-31"))

    def __init__(self, start_date: Optional[Date] = None, end_date: Optional[Date] = None):
        # check type
        if start_date and not isinstance(start_date, Date):
            raise ValueError(f"Attribute 'start_date' of class DateRange must be a Date object, but received type {type(start_date)}")
        if end_date and not isinstance(end_date, Date):
            raise ValueError(f"Attribute 'end_date' of class DateRange must be a Date object, but received type {type(end_date)}")

        self.start_date = start_date if start_date else Date(DEFAULT_START_DATE)
        self.end_date = end_date if end_date else Date(DEFAULT_END_DATE)
        if start_date and end_date and start_date.date > end_date.date:
            raise ValueError("Start date must be before or equal to end date, but received: start_date={}, end_date={}".format(start_date.date, end_date.date))

    def __str__(self):
        return f"DateRange(start_date={self.start_date}, end_date={self.end_date})"

    def __hash__(self):
        return hash((self.start_date, self.end_date))

    def __eq__(self, other):
        if isinstance(other, DateRange):
            return self.start_date == other.start_date and self.end_date == other.end_date
        return False

@dataclass
class ISOCode:
    """Represents an ISO alpha-3 country code."""
    code: str # 3-letter ISO code
    # Example: ISOCode("USA")

    def __init__(self, code: str):
        # check type
        if not isinstance(code, str):
            raise ValueError(f"Attribute 'code' of class ISOCode must be a string, but received type {type(code)}")

        if len(code) != 3:
            raise ValueError(f"ISO code must be a 3-letter string, but received: {code}")
        if code not in dict_iso2alternames:
            raise ValueError(f"ISO code must be a valid ISO alpha-3 country code, but received: {code}")
        self.code = code

    def __str__(self):
        return f"ISOCode('{self.code}')"

    def __hash__(self):
        return hash(self.code)

    def __eq__(self, other):
        if isinstance(other, ISOCode):
            return self.code == other.code
        return False

@dataclass
class Country:
    """Represents a country entity."""
    iso_code: ISOCode
    name: str
    # Example: Country(iso_code=ISOCode("USA"), name="United States")

    def __init__(self, iso_code: ISOCode, name: str):
        # check type
        if not isinstance(iso_code, ISOCode):
            raise ValueError(f"Attribute 'iso_code' of class Country must be an ISOCode object, but received type {type(iso_code)}")
        if not isinstance(name, str):
            raise ValueError(f"Attribute 'name' of class Country must be a string, but received type {type(name)}")

        if dict_iso2alternames[iso_code.code][0] != name:
            raise ValueError(f"Country name must match the name corresponding to the ISO code, but received: {name} for ISO code: {iso_code.code}")
        self.iso_code = iso_code
        self.name = name

    def __str__(self):
        return f"Country(iso_code={self.iso_code}, name='{self.name}')"

    def __hash__(self):
        return hash((self.iso_code, self.name))

    def __eq__(self, other):
        if isinstance(other, Country):
            return self.iso_code == other.iso_code and self.name == other.name
        return False

@dataclass
class CAMEOCode:
    """Represents a CAMEO verb code."""
    code: str # 2-digit CAMEO code for first level relations, 3-digit CAMEO code for second level relations
    # Example: CAMEOCode("01"), CAMEOCode("010")

    def __init__(self, code: str):
        # check type
        if not isinstance(code, str):
            raise ValueError(f"Attribute 'code' of class CAMEOCode must be a string, but received type {type(code)}")

        if len(code) not in [2, 3]:
            raise ValueError(f"CAMEO code must be a valid 2 or 3-digit string defined in the 'Conflict and Mediation Event Observations' Codebook, but received: {code}")
        if code not in dict_code2relation:
            raise ValueError(f"CAMEO code must be a valid CAMEO code defined in the 'Conflict and Mediation Event Observations' Codebook, but received: {code}")
        self.code = code

    def __str__(self):
        return f"CAMEOCode('{self.code}')"

    def __hash__(self):
        return hash(self.code)

    def __eq__(self, other):
        if isinstance(other, CAMEOCode):
            return self.code == other.code
        return False

@dataclass
class Relation:
    """Represents a relation."""
    cameo_code: CAMEOCode
    name: str
    description: str # A brief description of what event the relation represents
    # Example: Relation(cameo_code=CAMEOCode("010"), name="Make statement, not specified", description="All public statements expressed verbally or in action, not otherwise specified."

    def __init__(self, cameo_code: CAMEOCode, name: str, description: str):
        # check type
        if not isinstance(cameo_code, CAMEOCode):
            raise ValueError(f"Attribute 'cameo_code' of class Relation must be a CAMEOCode object, but received type {type(cameo_code)}")
        if not isinstance(name, str):
            raise ValueError(f"Attribute 'name' of class Relation must be a string, but received type {type(name)}")
        if not isinstance(description, str):
            raise ValueError(f"Attribute 'description' of class Relation must be a string, but received type {type(description)}")

        if dict_code2relation[cameo_code.code]['Name'] != name:
            raise ValueError(f"Relation name must match the name corresponding to the CAMEO code, but received: {name} for CAMEO code {cameo_code.code}")
        if dict_code2relation[cameo_code.code]['Description'] != description:
            raise ValueError(f"Relation description must match the description corresponding to the CAMEO code, but received: {description} for CAMEO code {cameo_code.code}")
        self.cameo_code = cameo_code
        self.name = name
        self.description = description

    def __str__(self):
        return f"Relation(cameo_code={self.cameo_code}, name='{self.name}', description='{self.description}')"

    def __hash__(self):
        return hash((self.cameo_code, self.name, self.description))

    def __eq__(self, other):
        if isinstance(other, Relation):
            return self.cameo_code == other.cameo_code and self.name == other.name and self.description == other.description
        return False

@dataclass
class Event:
    """Represents an event characterized by date, head entity, relation, and tail entity."""
    date: Date
    head_entity: ISOCode
    relation: CAMEOCode
    tail_entity: ISOCode
    # Example: Event(date=Date("2022-01-01"), head_entity=ISOCode("USA"), relation=CAMEOCode("010"), tail_entity=ISOCode("CAN"))

    def __init__(self, date: Date, head_entity: ISOCode, relation: CAMEOCode, tail_entity: ISOCode):
        # check type
        if not isinstance(date, Date):
            raise ValueError(f"Attribute 'date' of class Event must be a Date object, but received type {type(date)}")
        if not isinstance(head_entity, ISOCode):
            raise ValueError(f"Attribute 'head_entity' of class Event must be an ISOCode object, but received type {type(head_entity)}")
        if not isinstance(relation, CAMEOCode):
            raise ValueError(f"Attribute 'relation' of class Event must be a CAMEOCode object, but received type {type(relation)}")
        if not isinstance(tail_entity, ISOCode):
            raise ValueError(f"Attribute 'tail_entity' of class Event must be an ISOCode object, but received type {type(tail_entity)}")

        self.date = date
        self.head_entity = head_entity
        self.relation = relation
        self.tail_entity = tail_entity

    def __str__(self):
        return f"Event(date={self.date}, head_entity={self.head_entity}, relation={self.relation}, tail_entity={self.tail_entity})"

    def __hash__(self):
        return hash((self.date, self.head_entity, self.relation, self.tail_entity))

    def __eq__(self, other):
        if isinstance(other, Event):
            return self.date == other.date and self.head_entity == other.head_entity and self.relation == other.relation and self.tail_entity == other.tail_entity
        return False

@dataclass
class NewsArticle:
    """Represents a news article, including metadata and content."""
    date: Date
    title: str
    content: str # Full text content of the news article
    events: List[Event] # List of events mentioned in the article
    # Example: NewsArticle(date=Date("2022-01-01"), title="Trade agreement between USA and China", content="On January 1, 2022, a trade agreement was signed between the USA and China...", events=[Event(date=Date("2022-01-01"), head_entity=ISOCode("USA"), relation=CAMEOCode("010"), tail_entity=ISOCode("CHN")])

    def __init__(self, date: Date, title: str, content: str, events: List[Event]):
        # check type
        if not isinstance(date, Date):
            raise ValueError(f"Attribute 'date' of class NewsArticle must be a Date object, but received type {type(date)}")
        if not isinstance(title, str):
            raise ValueError(f"Attribute 'title' of class NewsArticle must be a string, but received type {type(title)}")
        if not isinstance(content, str):
            raise ValueError(f"Attribute 'content' of class NewsArticle must be a string, but received type {type(content)}")
        if not isinstance(events, list):
            raise ValueError(f"Attribute 'events' of class NewsArticle must be a list, but received type {type(events)}")

        self.date = date
        self.title = title
        self.content = content
        self.events = events

    def __str__(self):
        return f"{self.date}:\n{self.title}\n{self.content}"

    def __hash__(self):
        return hash((self.date, self.title, self.content))

    def __eq__(self, other):
        if isinstance(other, NewsArticle):
            return self.date == other.date and self.title == other.title and self.content == other.content
        return False

def map_country_name_to_iso(name: str) -> List[Country]:
    """
    Maps a country name to the most likely corresponding Country objects with ISO codes.

    Parameters:
        name (str): The country name to map.

    Returns:
        List[Country]: A list of 5 most likely Country objects sorted by relevance.

    Example:
        >>> map_country_name_to_iso("Korea")
        [Country(iso_code=ISOCode("KOR"), name="Republic of Korea"), Country(iso_code=ISOCode("PRK"), name="Democratic People's Republic of Korea")]
    """
    # check type
    if not isinstance(name, str):
        raise ValueError(f"Input 'name' must be a string, but received type {type(name)}")

    if name in dict_countryname2iso:
        return [Country(iso_code=ISOCode(dict_countryname2iso[name]), name=name)]
    else:
        # get top 5 ISO codes with the highest cosine similarity
        name_embeddings = get_embedding(name)
        similarities = cosine_similarity(country_embeddings, name_embeddings)
        top_indices = np.argsort(similarities)[::-1][:5]
        countries = []
        for idx in top_indices:
            iso_code = list(dict_iso2alternames.keys())[idx]
            curr_name = dict_iso2alternames[iso_code][0]
            countries.append(Country(iso_code=ISOCode(iso_code), name=curr_name))
        return countries

def map_iso_to_country_name(iso_code: ISOCode) -> str:
    """
    Maps an ISO code to a country name.

    Parameters:
        iso_code (ISOCode): The ISO code to map.

    Returns:
        str: The corresponding country name.

    Example:
        >>> map_iso_to_country_name(ISOCode("CHN"))
        "China"
    """
    # check type
    if not isinstance(iso_code, ISOCode):
        raise ValueError(f"Input 'iso_code' must be an ISOCode object, but received type {type(iso_code)}")

    return dict_iso2alternames[iso_code.code][0]

def map_relation_description_to_cameo(description: str) -> List[Relation]:
    """
    Maps a relation description to the most likely Relation objects.

    Parameters:
        description (str): The relation description to map.

    Returns:
        List[Relation]: A list of 5 most likely Relations sorted by relevance.

    Example:
        >>> map_relation_description_to_cameo("Fight with guns")
        [Reltion(cameo_code=CAMEOCode("19"), name="Fight", description="All uses of conventional force and acts of war typically by organized armed groups."), Relation(cameo_code=CAMEOCode("193"), name="Fight with small arms and light weapons", description="Attack using small arms and light weapons such as rifles, machine-guns, and mortar shells."), Relation(cameo_code=CAMEOCode("190"), name="Use conventional military force, not specified", description="All uses of conventional force and acts of war typically by organized armed groups, not otherwise specified.")]
    """
    # check type
    if not isinstance(description, str):
        raise ValueError(f"Input 'description' must be a string, but received type {type(description)}")

    if description in dict_relation2code:
        code = dict_relation2code[description]
        return [Relation(cameo_code=CAMEOCode(code), name=dict_code2relation[code]['Name'], description=dict_code2relation[code]['Description'])]
    else:
        # get top 5 CAMEO codes with the highest cosine similarity
        description_embedding = get_embedding(description)
        similarities = cosine_similarity(relation_embeddings, description_embedding)
        top_indices = np.argsort(similarities)[::-1][:5]
        relations = []
        for idx in top_indices:
            code = list(dict_code2relation.keys())[idx]
            relations.append(Relation(cameo_code=CAMEOCode(code), name=dict_code2relation[code]['Name'], description=dict_code2relation[code]['Description']))
        return relations

def map_cameo_to_relation(cameo_code: CAMEOCode) -> Relation:
    """
    Maps a CAMEO code to a relation, including its name and description.

    Parameters:
        cameo_code (CAMEOCode): The CAMEO code to map.

    Returns:
        Relation: The corresponding relation.

    Example:
        >>> map_cameo_to_relation(CAMEOCode("190"))
        Relation(cameo_code=CAMEOCode("190"), name="Use conventional military force, not specified", description="All uses of conventional force and acts of war typically by organized armed groups, not otherwise specified.")
    """
    # check type
    if not isinstance(cameo_code, CAMEOCode):
        raise ValueError(f"Input 'cameo_code' must be a CAMEOCode object, but received type {type(cameo_code)}")

    info = dict_code2relation[cameo_code.code]
    return Relation(cameo_code=cameo_code, name=info['Name'], description=info['Description'])

def get_parent_relation(cameo_code: CAMEOCode) -> Relation:
    """
    Retrieves the parent relation of a given relation identified by CAMEO code.

    Parameters:
        cameo_code (CAMEOCode): The CAMEO code of the relation whose parent is sought. Only second level relations are accepted.

    Returns:
        Relation: The first level parent relation.

    Example:
        >>> get_parent_relation(CAMEOCode("193"))
        Relation(cameo_code=CAMEOCode("19"), name="Fight", description="All uses of conventional force and acts of war typically by organized armed groups.")
    """
    # check type
    if not isinstance(cameo_code, CAMEOCode):
        raise ValueError(f"Input 'cameo_code' must be a CAMEOCode object, but received type {type(cameo_code)}")

    if len(cameo_code.code) != 3:
        raise ValueError("Only second level relations are accepted, but received: {}".format(cameo_code.code))
    parent_code =cameo_code.code[:2]
    return map_cameo_to_relation(CAMEOCode(parent_code))

def get_child_relations(cameo_code: CAMEOCode) -> List[Relation]:
    """
    Retrieves child relations of a given relation identified by CAMEO code.

    Parameters:
        cameo_code (CAMEOCode): The CAMEO code of the relation whose children are sought. Only first level relations are accepted.

    Returns:
        List[Relation]: A list of second level child relations.

    Example:
        >>> get_child_relations(CAMEOCode("19"))
        [Relation(caemo_code=CAMEOCode("190"), name="Use conventional military force, not specified", description="All uses of conventional force and acts of war typically by organized armed groups, not otherwise specified."), Relation(cameo_code=CAMEOCode("191"), name="Impose blockade or restrict movement", description="Prevent entry into and/or exit from a territory using armed forces."), ...]
    """
    # check type
    if not isinstance(cameo_code, CAMEOCode):
        raise ValueError(f"Input 'cameo_code' must be a CAMEOCode object, but received type {type(cameo_code)}")

    if len(cameo_code.code) != 2:
        raise ValueError("Only first level relations are accepted, but received: {}".format(cameo_code.code))
    children = []
    for code, info in dict_code2relation.items():
        if code[:2] == cameo_code.code and len(code) == 3:
            children.append(Relation(cameo_code=CAMEOCode(code), name=info['Name'], description=info['Description']))
    return children

def get_sibling_relations(cameo_code: CAMEOCode) -> List[Relation]:
    """
    Retrieves sibling relations of a given relation identified by CAMEO code.

    Parameters:
        cameo_code (CAMEOCode): The CAMEO code of the relation whose siblings are sought. Both first and second level relations are accepted.

    Returns:
        List[Relation]: A list of sibling relations at the same level.

    Example:
        >>> get_sibling_relations(CAMEOCode("193"))
        [Relation(caemo_code=CAMEOCode("190"), name="Use conventional military force, not specified", description="All uses of conventional force and acts of war typically by organized armed groups, not otherwise specified."), Relation(cameo_code=CAMEOCode("191"), name="Impose blockade or restrict movement", description="Prevent entry into and/or exit from a territory using armed forces."), ...]
    """
    # check type
    if not isinstance(cameo_code, CAMEOCode):
        raise ValueError(f"Input 'cameo_code' must be a CAMEOCode object, but received type {type(cameo_code)}")

    if len(cameo_code.code) == 3:
        return get_child_relations(get_parent_relation(cameo_code).cameo_code)
    elif len(cameo_code.code) == 2:
        # get '01' to '20' relations
        first_level_codes = [str(i).zfill(2) for i in range(1, 21)]
        relations = []
        for code in first_level_codes:
            relations.append(map_cameo_to_relation(CAMEOCode(code)))
        return relations

def count_events(date_range: Optional[DateRange] = None, head_entities: Optional[List[ISOCode]] = None, tail_entities: Optional[List[ISOCode]] = None, relations: Optional[List[CAMEOCode]] = None) -> int:
    """
    Counts the number of events in the knowledge graph based on specified conditions.

    Parameters:
        date_range (Optional[DateRange]): Range of dates to filter the events. If None, all dates are included.
        head_entities (Optional[List[ISOCode]]): List of head entity ISO codes to be included. If None, all countries are included.
        tail_entities (Optional[List[ISOCode]]): List of tail entity ISO codes to be included. If None, all countries are included.
        relations (Optional[List[CAMEOCode]]): List of relation CAMEO codes to be included. If first level relations are listed, all second level relations under them are included. If None, all relations are included.

    Returns:
        int: Count of unique events matching the conditions.

    Example:
        >>> count_events(date_range=DateRange(start_date=Date("2022-01-01"), end_date=Date("2022-01-31")), head_entities=[ISOCode("USA"), ISOCode("CHN")], tail_entities=None, relations=[CAMEOCode("010")])
        4
    """
    # check type
    if date_range and not isinstance(date_range, DateRange):
        raise ValueError(f"Input 'date_range' must be a DateRange object, but received type {type(date_range)}")
    if head_entities and not isinstance(head_entities, list):
        raise ValueError(f"Input 'head_entities' must be a list, but received type {type(head_entities)}")
    if head_entities and not all(isinstance(iso, ISOCode) for iso in head_entities):
        raise ValueError(f"Elements in 'head_entities' must be ISOCode objects")
    if tail_entities and not isinstance(tail_entities, list):
        raise ValueError(f"Input 'tail_entities' must be a list, but received type {type(tail_entities)}")
    if tail_entities and not all(isinstance(iso, ISOCode) for iso in tail_entities):
        raise ValueError(f"Elements in 'tail_entities' must be ISOCode objects")
    if relations and not isinstance(relations, list):
        raise ValueError(f"Input 'relations' must be a list, but received type {type(relations)}")
    if relations and not all(isinstance(code, CAMEOCode) for code in relations):
        raise ValueError(f"Elements in 'relations' must be CAMEOCode objects")

    # process data_kg by filtering based on the specified conditions
    curr_data = data_kg.copy()
    curr_data.drop_duplicates(subset=['QuadEventCode'], inplace=True)
    curr_data = curr_data[curr_data['DateStr'] <= DEFAULT_END_DATE]
    if date_range:
        curr_data = curr_data[(curr_data['DateStr'] >= date_range.start_date.date) & (curr_data['DateStr'] <= date_range.end_date.date)]
    if head_entities:
        curr_data = curr_data[curr_data['Actor1CountryCode'].isin([iso.code for iso in head_entities])]
    if tail_entities:
        curr_data = curr_data[curr_data['Actor2CountryCode'].isin([iso.code for iso in tail_entities])]
    if relations:
        # if first level relations are listed, include all second level relations under them
        for code in relations:
            if len(code.code) == 2:
                relations.extend([CAMEOCode(c) for c in dict_code2relation if c[:2] == code.code and len(c) == 3])
        curr_data = curr_data[curr_data['EventBaseCode'].isin([code.code for code in relations])]
    return len(curr_data)

def get_events(date_range: Optional[DateRange] = None, head_entities: Optional[List[ISOCode]] = None, tail_entities: Optional[List[ISOCode]] = None, relations: Optional[List[CAMEOCode]] = None, text_description: Optional[str] = None) -> List[Event]:
    """
    Retrieves events from the knowledge graph based on specified conditions.
    Inherits common filter parameters from count_events. See count_events for more details on these parameters.

    Additional Parameters:
        text_description (Optional[str]): Textual description to match with the source news articles of events. If None, the returned events are sorted by date in descending order; otherwise, sorted by relevance of the source news article to the description.

    Returns:
        List[Event]: A list of maximum 30 events matching the specified conditions.

    Example:
        >>> get_events(date_range=DateRange(start_date=Date("2022-01-01"), end_date=Date("2022-01-31")), head_entities=[ISOCode("USA"), ISOCode("CHN")], tail_entities=None, relations=[CAMEOCode("010")], text_description="economic trade")
        [Event(date=Date("2022-01-15"), head_entity=ISOCode("USA"), relation=CAMEOCode("010"), tail_entity=ISOCode("CAN"), Event(date=Date("2022-01-10"), head_entity=ISOCode("CHN"), relation=CAMEOCode("010"), tail_entity=ISOCode("USA"), ...]
    """
    # check type
    if date_range and not isinstance(date_range, DateRange):
        raise ValueError(f"Input 'date_range' must be a DateRange object, but received type {type(date_range)}")
    if head_entities and not isinstance(head_entities, list):
        raise ValueError(f"Input 'head_entities' must be a list, but received type {type(head_entities)}")
    if head_entities and not all(isinstance(iso, ISOCode) for iso in head_entities):
        raise ValueError(f"Elements in 'head_entities' must be ISOCode objects")
    if tail_entities and not isinstance(tail_entities, list):
        raise ValueError(f"Input 'tail_entities' must be a list, but received type {type(tail_entities)}")
    if tail_entities and not all(isinstance(iso, ISOCode) for iso in tail_entities):
        raise ValueError(f"Elements in 'tail_entities' must be ISOCode objects")
    if relations and not isinstance(relations, list):
        raise ValueError(f"Input 'relations' must be a list, but received type {type(relations)}")
    if relations and not all(isinstance(code, CAMEOCode) for code in relations):
        raise ValueError(f"Elements in 'relations' must be CAMEOCode objects")
    if text_description and not isinstance(text_description, str):
        raise ValueError(f"Input 'text_description' must be a string, but received type {type(text_description)}")

    # process data_kg by filtering based on the specified conditions
    curr_data = data_kg.copy()
    curr_data = curr_data[curr_data['DateStr'] <= DEFAULT_END_DATE]
    if date_range:
        curr_data = curr_data[(curr_data['DateStr'] >= date_range.start_date.date) & (curr_data['DateStr'] <= date_range.end_date.date)]
    if head_entities:
        curr_data = curr_data[curr_data['Actor1CountryCode'].isin([iso.code for iso in head_entities])]
    if tail_entities:
        curr_data = curr_data[curr_data['Actor2CountryCode'].isin([iso.code for iso in tail_entities])]
    if relations:
        # if first level relations are listed, include all second level relations under them
        for code in relations:
            if len(code.code) == 2:
                relations.extend([CAMEOCode(c) for c in dict_code2relation if c[:2] == code.code and len(c) == 3])
        curr_data = curr_data[curr_data['EventBaseCode'].isin([code.code for code in relations])]
    if not text_description:
        # get max 30 events from the filtered data
        events = []
        curr_data.drop_duplicates(subset=['QuadEventCode'], inplace=True)
        # sorted by date in descending order
        curr_data.sort_values(by='DateStr', ascending=False, inplace=True)
        count = 0
        for _, row in curr_data.iterrows():
            if count >= 30:
                break
            events.append(Event(date=Date(row['DateStr']), head_entity=ISOCode(row['Actor1CountryCode']), relation=CAMEOCode(row['EventBaseCode']), tail_entity=ISOCode(row['Actor2CountryCode'])))
        return events[:30]
    else:
        # concat the Docids list of current data to get the news articles
        docids_list = [eval(docids) for docids in curr_data['Docids'].unique().tolist()]
        docids = list(set([item for sublist in docids_list for item in sublist]))
        docids = [str(docid) for docid in docids]
        news_articles = data_news[data_news['Docid'].isin(docids)]
        # get the max 30 docids with the highest BM25 score to the text_description
        corpus = news_articles['Title'] + ' ' + news_articles['Text']
        tokenized_corpus = [doc.split(" ") for doc in corpus]
        bm25 = BM25Okapi(tokenized_corpus)
        tokenized_query = text_description.split(" ")
        doc_scores = bm25.get_scores(tokenized_query)
        top_indices = np.argsort(doc_scores)[::-1][:30]
        news_articles = news_articles.iloc[top_indices]
        docids = news_articles['Docid'].tolist()
        # get max 30 events from the filtered data
        events = set()
        for docid in docids:
            if len(events) >= 30:
                break
            doc_curr_data = curr_data[curr_data['Docid'] == docid]
            # reverse the order of the events to get the latest events first
            doc_curr_data = doc_curr_data.sort_values(by='DateStr', ascending=False)
            for _, row in doc_curr_data.iterrows():
                events.add(Event(date=Date(row['DateStr']), head_entity=ISOCode(row['Actor1CountryCode']), relation=CAMEOCode(row['EventBaseCode']), tail_entity=ISOCode(row['Actor2CountryCode'])))
        return list(events)

def get_entity_distribution(date_range: Optional[DateRange] = None, involved_relations: Optional[List[CAMEOCode]] = None, interacted_entities: Optional[List[ISOCode]] = None, entity_role: Optional[str] = None) -> Dict[ISOCode, int]:
    """
    Gets the distribution of entities in the knowledge graph under specified conditions.

    Parameters:
        date_range (Optional[DateRange]): Range of dates to filter the events. If None, all dates are included.
        involved_relations (Optional[List[CAMEOCode]]): List of relations that the returned entities must be involved in any of. If first level relations are listed, all second level relations under them are included. If None, all relations are included.
        interacted_entities (Optional[List[ISOCode]]): List of entities that the returned entities must have interacted with any of. If None, all entities are included.
        entity_role (Optional[EntityRole]): Specifies the role of the returned entity in the events. Options are 'head', 'tail', or 'both'. If 'both' or None, the returned entity can be either head or tail.

    Returns:
        Dict[ISOCode, int]: A dictionary mapping returned entities' ISO codes to the number of events with the specified conditions in which they are involved, sorted by counts in descending order.

    Example:
        >>> get_entity_distribution(date_range=DateRange(start_date=Date("2022-01-01"), end_date=Date("2022-01-31")), involved_relations=[CAMEOCode("010")], interacted_entities=[ISOCode("USA"), ISOCode("CHN")], entity_role="tail")
        {ISOCode("USA"): 3, ISOCode("CHN"): 1}
    """
    # check type
    if date_range and not isinstance(date_range, DateRange):
        raise ValueError(f"Input 'date_range' must be a DateRange object, but received type {type(date_range)}")
    if involved_relations and not isinstance(involved_relations, list):
        raise ValueError(f"Input 'involved_relations' must be a list, but received type {type(involved_relations)}")
    if involved_relations and not all(isinstance(code, CAMEOCode) for code in involved_relations):
        raise ValueError(f"Elements in 'involved_relations' must be CAMEOCode objects")
    if interacted_entities and not isinstance(interacted_entities, list):
        raise ValueError(f"Input 'interacted_entities' must be a list, but received type {type(interacted_entities)}")
    if interacted_entities and not all(isinstance(iso, ISOCode) for iso in interacted_entities):
        raise ValueError(f"Elements in 'interacted_entities' must be ISOCode objects")
    if entity_role and entity_role not in ['head', 'tail', 'both']:
        raise ValueError(f"Input 'entity_role' must be a string 'head', 'tail', or 'both', but received: {entity_role}")

    # process data_kg by filtering based on the specified conditions
    curr_data = data_kg.copy()
    curr_data.drop_duplicates(subset=['QuadEventCode'], inplace=True)
    curr_data = curr_data[curr_data['DateStr'] <= DEFAULT_END_DATE]
    if date_range:
        curr_data = curr_data[(curr_data['DateStr'] >= date_range.start_date.date) & (curr_data['DateStr'] <= date_range.end_date.date)]
    if involved_relations:
        # if first level relations are listed, include all second level relations under them
        for code in involved_relations:
            if len(code.code) == 2:
                involved_relations.extend([CAMEOCode(c) for c in dict_code2relation if c[:2] == code.code and len(c) == 3])
        curr_data = curr_data[curr_data['EventBaseCode'].isin([code.code for code in involved_relations])]
    if interacted_entities:
        if entity_role=='head':
            curr_data = curr_data[curr_data['Actor2CountryCode'].isin([iso.code for iso in interacted_entities])]
        elif entity_role=='tail':
            curr_data = curr_data[curr_data['Actor1CountryCode'].isin([iso.code for iso in interacted_entities])]
        else:
            curr_data = curr_data[(curr_data['Actor1CountryCode'].isin([iso.code for iso in interacted_entities])) | (curr_data['Actor2CountryCode'].isin([iso.code for iso in interacted_entities]))]
    # count the number of events for each entity
    entity_counts = curr_data['Actor1CountryCode']._append(curr_data['Actor2CountryCode']).value_counts()
    entity_counts = entity_counts.to_dict()
    # sort the dictionary by values in descending order
    entity_counts = dict(sorted(entity_counts.items(), key=lambda item: item[1], reverse=True))
    entity_counts = {ISOCode(key): value for key, value in entity_counts.items()}
    return entity_counts

def get_relation_distribution(date_range: Optional[DateRange] = None, head_entities: Optional[List[ISOCode]] = None, tail_entities: Optional[List[ISOCode]] = None) -> Dict[CAMEOCode, int]:
    """
    Gets the distribution of second level relations in the knowledge graph under specified conditions.

    Parameters:
        date_range (Optional[DateRange]): Range of dates to filter the events. If None, all dates are included.
        head_entities (Optional[List[ISOCode]]): List of head entities that the events must involve any of. If None, all head entities are included.
        tail_entities (Optional[List[ISOCode]]): List of tail entities that the events must involve any of. If None, all tail entities are included.

    Returns:
        Dict[CAMEOCode, int]: A dictionary mapping second level relations' CAMEO codes to the number of events with the specified conditions in which they are involved, sorted by counts in descending order.

    Example:
        >>> get_relation_distribution(date_range=DateRange(start_date=Date("2022-01-01"), end_date=Date("2022-01-31")), head_entities=[ISOCode("USA"), ISOCode("CHN")], tail_entities=None)
        {CAMEOCode("010"): 3, CAMEOCode("011"): 1}
    """
    # check type
    if date_range and not isinstance(date_range, DateRange):
        raise ValueError(f"Input 'date_range' must be a DateRange object, but received type {type(date_range)}")
    if head_entities and not isinstance(head_entities, list):
        raise ValueError(f"Input 'head_entities' must be a list, but received type {type(head_entities)}")
    if head_entities and not all(isinstance(iso, ISOCode) for iso in head_entities):
        raise ValueError(f"Elements in 'head_entities' must be ISOCode objects")
    if tail_entities and not isinstance(tail_entities, list):
        raise ValueError(f"Input 'tail_entities' must be a list, but received type {type(tail_entities)}")
    if tail_entities and not all(isinstance(iso, ISOCode) for iso in tail_entities):
        raise ValueError(f"Elements in 'tail_entities' must be ISOCode objects")

    # process data_kg by filtering based on the specified conditions
    curr_data = data_kg.copy()
    curr_data.drop_duplicates(subset=['QuadEventCode'], inplace=True)
    curr_data = curr_data[curr_data['DateStr'] <= DEFAULT_END_DATE]
    if date_range:
        curr_data = curr_data[(curr_data['DateStr'] >= date_range.start_date.date) & (curr_data['DateStr'] <= date_range.end_date.date)]
    if head_entities:
        curr_data = curr_data[curr_data['Actor1CountryCode'].isin([iso.code for iso in head_entities])]
    if tail_entities:
        curr_data = curr_data[curr_data['Actor2CountryCode'].isin([iso.code for iso in tail_entities])]
    # count the number of events for each relation
    relation_counts = curr_data['EventBaseCode'].value_counts()
    relation_counts = relation_counts.to_dict()
    # sort the dictionary by values in descending order
    relation_counts = dict(sorted(relation_counts.items(), key=lambda item: item[1], reverse=True))
    relation_counts = {CAMEOCode(key): value for key, value in relation_counts.items()}
    return relation_counts

def count_news_articles(date_range: Optional[DateRange] = None, head_entities: Optional[List[ISOCode]] = None, tail_entities: Optional[List[ISOCode]] = None, relations: Optional[List[CAMEOCode]] = None, keywords: Optional[List[str]] = None) -> int:
    """
    Counts the number of news articles based on specified conditions.

    Parameters:
        date_range (Optional[DateRange]): Range of dates to filter the news articles. If None, all dates are included.
        head_entities (Optional[List[ISOCode]]): At least one of the entities must be mentioned in the articles and be the head entity in the events. If None, all entities are included.
        tail_entities (Optional[List[ISOCode]]): At least one of the entities must be mentioned in the articles and be the tail entity in the events. If None, all entities are included.
        relations (Optional[List[CAMEOCode]]): At least one of the relations must be mentioned in the articles. If first level relations are listed, all second level relations under them are included. If None, all relations are included.
        keywords (Optional[List[str]]): At least one of the keywords must be present in the articles. If None, all articles are included.

    Returns:
        int: The count of news articles matching the conditions.

    Example:
        >>> count_news_articles(date_range=DateRange(start_date=Date("2022-01-01"), end_date=Date("2022-01-31")), head_entities=[ISOCode("USA"), ISOCode("CHN")], tail_entities=[ISOCode("USA"), ISOCode("CHN")], relations=[CAMEOCode("010")], keywords=["trade"])
        2
    """
    # check type
    if date_range and not isinstance(date_range, DateRange):
        raise ValueError(f"Input 'date_range' must be a DateRange object, but received type {type(date_range)}")
    if head_entities and not isinstance(head_entities, list):
        raise ValueError(f"Input 'head_entities' must be a list, but received type {type(head_entities)}")
    if head_entities and not all(isinstance(iso, ISOCode) for iso in head_entities):
        raise ValueError(f"Elements in 'head_entities' must be ISOCode objects")
    if tail_entities and not isinstance(tail_entities, list):
        raise ValueError(f"Input 'tail_entities' must be a list, but received type {type(tail_entities)}")
    if tail_entities and not all(isinstance(iso, ISOCode) for iso in tail_entities):
        raise ValueError(f"Elements in 'tail_entities' must be ISOCode objects")
    if relations and not isinstance(relations, list):
        raise ValueError(f"Input 'relations' must be a list, but received type {type(relations)}")
    if relations and not all(isinstance(code, CAMEOCode) for code in relations):
        raise ValueError(f"Elements in 'relations' must be CAMEOCode objects")
    if keywords and not isinstance(keywords, list):
        raise ValueError(f"Input 'keywords' must be a list, but received type {type(keywords)}")
    if keywords and not all(isinstance(keyword, str) for keyword in keywords):
        raise ValueError(f"Elements in 'keywords' must be strings")

    # process data_kg by filtering based on the specified conditions
    curr_data = data_kg.copy()
    curr_data.drop_duplicates(subset=['QuadEventCode'], inplace=True)
    curr_data = curr_data[curr_data['DateStr'] <= DEFAULT_END_DATE]
    if date_range:
        curr_data = curr_data[(curr_data['DateStr'] >= date_range.start_date.date) & (curr_data['DateStr'] <= date_range.end_date.date)]
    if head_entities:
        curr_data = curr_data[curr_data['Actor1CountryCode'].isin([iso.code for iso in head_entities])]
    if tail_entities:
        curr_data = curr_data[curr_data['Actor2CountryCode'].isin([iso.code for iso in tail_entities])]
    if relations:
        # if first level relations are listed, include all second level relations under them
        for code in relations:
            if len(code.code) == 2:
                relations.extend([CAMEOCode(c) for c in dict_code2relation if c[:2] == code.code and len(c) == 3])
        curr_data = curr_data[curr_data['EventBaseCode'].isin([code.code for code in relations])]
    # concat the Docids list of current data to get the news articles
    docids_list = [eval(docids) for docids in curr_data['Docids'].unique().tolist()]
    docids = list(set([item for sublist in docids_list for item in sublist]))
    docids = [str(docid) for docid in docids]
    news_articles = data_news[data_news['Docid'].isin(docids)]
    if keywords:
        # filter the news articles that contain at least one of the keywords in the title or text string
        news_articles = news_articles[news_articles['Title'].str.contains('|'.join(keywords), case=False) | news_articles['Text'].str.contains('|'.join(keywords), case=False)]
    return len(news_articles)

def get_news_articles(date_range: Optional[DateRange] = None, head_entities: Optional[List[ISOCode]] = None, tail_entities: Optional[List[ISOCode]] = None, relations: Optional[List[CAMEOCode]] = None, keywords: Optional[List[str]] = None, text_description: Optional[str] = None) -> List[Tuple[Date, str]]:
    """
    Retrieves news articles based on specified conditions.
    Inherits common filter parameters from count_news_articles. See count_news_articles for more details on these parameters.

    Additional Parameters:
        text_description (Optional[str]): Textual description to match with the news articles. If None, the returned articles are sorted by date in descending order; otherwise, sorted by relevance to the description.

    Returns:
        List[Tuple[Date, str]]: A list of maximum 15 news articles matching the specified conditions, each represented by a tuple of date and title.

    Example:
        >>> get_news_articles(date_range=DateRange(start_date=Date("2022-01-01"), end_date=Date("2022-01-31")), head_entities=[ISOCode("USA"), ISOCode("CHN")], tail_entities=[ISOCode("USA"), ISOCode("CHN")], relations=[CAMEOCode("010")], keywords=["trade"], text_description="Economic trade is encouraged between USA and China.")
        [(NewsArticle.date=Date("2022-01-15"), NewsArticle.title="China and USA sign trade deal"), (NewsArticle.date=Date("2022-01-10"), NewsArticle.title="Trade agreement between USA and China")]
    """
    # check type
    if date_range and not isinstance(date_range, DateRange):
        raise ValueError(f"Input 'date_range' must be a DateRange object, but received type {type(date_range)}")
    if head_entities and not isinstance(head_entities, list):
        raise ValueError(f"Input 'head_entities' must be a list, but received type {type(head_entities)}")
    if head_entities and not all(isinstance(iso, ISOCode) for iso in head_entities):
        raise ValueError(f"Elements in 'head_entities' must be ISOCode objects")
    if tail_entities and not isinstance(tail_entities, list):
        raise ValueError(f"Input 'tail_entities' must be a list, but received type {type(tail_entities)}")
    if tail_entities and not all(isinstance(iso, ISOCode) for iso in tail_entities):
        raise ValueError(f"Elements in 'tail_entities' must be ISOCode objects")
    if relations and not isinstance(relations, list):
        raise ValueError(f"Input 'relations' must be a list, but received type {type(relations)}")
    if relations and not all(isinstance(code, CAMEOCode) for code in relations):
        raise ValueError(f"Elements in 'relations' must be CAMEOCode objects")
    if keywords and not isinstance(keywords, list):
        raise ValueError(f"Input 'keywords' must be a list, but received type {type(keywords)}")
    if keywords and not all(isinstance(keyword, str) for keyword in keywords):
        raise ValueError(f"Elements in 'keywords' must be strings")
    if text_description and not isinstance(text_description, str):
        raise ValueError(f"Input 'text_description' must be a string, but received type {type(text_description)}")

    # process data_kg by filtering based on the specified conditions
    curr_data = data_kg.copy()
    curr_data.drop_duplicates(subset=['QuadEventCode'], inplace=True)
    curr_data = curr_data[curr_data['DateStr'] <= DEFAULT_END_DATE]
    if date_range:
        curr_data = curr_data[
            (curr_data['DateStr'] >= date_range.start_date.date) & (curr_data['DateStr'] <= date_range.end_date.date)]
    if head_entities:
        curr_data = curr_data[curr_data['Actor1CountryCode'].isin([iso.code for iso in head_entities])]
    if tail_entities:
        curr_data = curr_data[curr_data['Actor2CountryCode'].isin([iso.code for iso in tail_entities])]
    if relations:
        # if first level relations are listed, include all second level relations under them
        for code in relations:
            if len(code.code) == 2:
                relations.extend([CAMEOCode(c) for c in dict_code2relation if c[:2] == code.code and len(c) == 3])
        curr_data = curr_data[curr_data['EventBaseCode'].isin([code.code for code in relations])]
    docids_list = [eval(docids) for docids in curr_data['Docids'].unique().tolist()]
    docids = list(set([item for sublist in docids_list for item in sublist]))
    docids = [str(docid) for docid in docids]
    news_articles = data_news[data_news['Docid'].isin(docids)]
    if keywords:
        # filter the news articles that contain at least one of the keywords in the title or text string
        news_articles = news_articles[news_articles['Title'].str.contains('|'.join(keywords), case=False) | news_articles['Text'].str.contains('|'.join(keywords), case=False)]
    if not text_description:
        # get max 15 news articles from the filtered data
        # sorted by date in descending order
        news_articles.sort_values(by='Date', ascending=False, inplace=True)
        news_articles = news_articles[['Date', 'Title']].head(15)
        return [(Date(row['Date']), row['Title']) for _, row in news_articles.iterrows()]
    else:
        # get the max 15 news articles with the highest BM25 score to the text_description
        corpus = news_articles['Title'] + ' ' + news_articles['Text']
        tokenized_corpus = [doc.split(" ") for doc in corpus]
        bm25 = BM25Okapi(tokenized_corpus)
        tokenized_query = text_description.split(" ")
        doc_scores = bm25.get_scores(tokenized_query)
        top_indices = np.argsort(doc_scores)[::-1][:15]
        news_articles = news_articles.iloc[top_indices]
        return [(Date(row['Date']), row['Title']) for _, row in news_articles.iterrows()]

def browse_news_article(date: Date, title: str) -> str:
    """
    Retrieves the full text of a news article by its title.

    Parameters:
        date (Date): The date of the news article to retrieve.
        title (str): The title of the news article to retrieve.

    Returns:
        str: The date, the title and full contents of the news article.

    Example:
        >>> browse_news_article(Date("2022-01-10"), "Trade agreement between USA and China")
        2022-01-10:
        Trade agreement between USA and China
        On January 10, 2022, a trade agreement was signed between the USA and China to promote economic cooperation...
    """
    # check type
    if not isinstance(date, Date):
        raise ValueError(f"Input 'date' must be a Date object, but received type {type(date)}")
    if not isinstance(title, str):
        raise ValueError(f"Input 'title' must be a string, but received type {type(title)}")

    # process data_news to find the news article with the specified date and title
    curr_data = data_news[(data_news['Date'] == date.date) & (data_news['Title'] == title)]
    if len(curr_data) == 0:
        raise ValueError(f"No news article found with the specified date {date.date} and title {title}")
    return f"{date}:\n{title}\n{curr_data['Text'].values[0]}"
\end{lstlisting}
\clearpage

\end{document}